\documentclass[journal]{IEEEtran}

\usepackage{graphicx}
\usepackage{amsmath, amsfonts,epsfig, multirow, floatflt}
\usepackage{amssymb}
\usepackage{subfigure}
\usepackage{cite}
\usepackage{algorithm,algorithmic}
\usepackage{hyperref}

\usepackage{enumitem}
\usepackage{array}
\usepackage{color}
\usepackage[table]{xcolor}
\usepackage{colortbl}
\newcolumntype{C}[1]{>{\centering\arraybackslash}p{#1}}

\makeatletter
\def\ps@headings{%
	\def\@oddhead{\mbox{}\scriptsize\rightmark \hfil \thepage}%
	\def\@evenhead{\scriptsize\thepage \hfil \leftmark\mbox{}}%
	\def\@oddfoot{}%
	\def\@evenfoot{}}
\makeatother 

\graphicspath{{figure/}}
\begin{document}

\title{Deep Reinforcement Learning for Autonomous Internet of Things: Model, Applications and Challenges}

\author{Lei~Lei {\it Senior Member, IEEE}, Yue~Tan,  Kan~Zheng {\it Senior Member, IEEE}, Shiwen~Liu, Kuan~Zhang, Xuemin (Sherman) Shen {\it Fellow, IEEE}
	
%\thanks{Manuscript received June 18, 2019; revised October 10, 2019; accepted April 13, 2020 (\emph{Corresponding author: Lei Lei})}

%\thanks{L. Lei is with the College of Engineering and Physical Sciences, University of Guelph, ON N1G 2W1, Canada (e-mail: leil@uoguelph.ca).}
%
%\thanks{Y. Tan, K. Zheng, and S. Liu are with the Intelligent Computing and Communication (IC$^2$) Lab, Key Laboratory of Universal Wireless Communications, Ministry of Education, Beijing University of Posts and Telecommunications (BUPT), Beijing 100876, China.}
%
%\thanks{K. Zhang is with is with the Department of Electrical \& Computer Engineering, University of Nebraska-Lincoln, PKI 206D, Scott Campus, 68182-0572, United States.}
%
%\thanks{X. Shen is with the Department of Electrical and Computer Engineering, University of Waterloo, Waterloo, ON N2L 3G1, Canada.}
%
%\thanks{This work was supported by the National Natural Science Foundation of China (NSFC) under Grant 61671089.}
}

\maketitle
	
\begin{abstract}
The Internet of Things (IoT) extends the Internet connectivity into billions of IoT devices around the world, where the IoT devices collect and share information to reflect status of the physical world. The Autonomous Control System (ACS), on the other hand, performs control functions on the physical systems without external intervention over an extended period of time. The integration of IoT and ACS results in a new concept - autonomous IoT (AIoT). The sensors collect information on the system status, based on which the intelligent agents in the IoT devices as well as the Edge/Fog/Cloud servers make control decisions for the actuators to react. In order to achieve autonomy, a promising method is for the intelligent agents to leverage the techniques in the field of artificial intelligence, especially reinforcement learning (RL) and deep reinforcement learning (DRL) for decision making. In this paper, we first provide a tutorial of DRL, and then propose a general model for the applications of RL/DRL in AIoT. Next, a comprehensive survey of the state-of-art research on DRL for AIoT is presented, where the existing works are classified and summarized under the umbrella of the proposed general DRL model. Finally, the challenges and open issues for future research are identified.

\end{abstract}

\begin{IEEEkeywords}
Autonomous Internet of Things; Deep Reinforcement Learning
\end{IEEEkeywords}

\section{Introduction}

\subsection{Autonomous Internet of Things}
The Internet of Things (IoT) connects a huge number of IoT devices to the Internet, where the IoT devices generate massive amount of sensory data to reflect status of the physical world. These data could be processed and analyzed by leveraging machine learning (ML) techniques, with the objective of making informed decisions to control the reactions of IoT devices to the physical world. In other words, IoT devices become autonomous with ambient intelligence by integrating IoT, ML and autonomous control. For example, smart thermostats can learn to autonomously control central heating systems based on the presence of users and their routine. IoT and autonomous control system (ACS) \cite{antsaklis1991} are originally independent concepts, and the realization of one does not necessarily require the other. The concept of autonomous IoT (AIoT) was proposed as the next wave of IoT that can explore its future potential \cite{aIoT}.\par
\begin{figure}[h]
	\centering
	\includegraphics[width=0.45\textwidth]{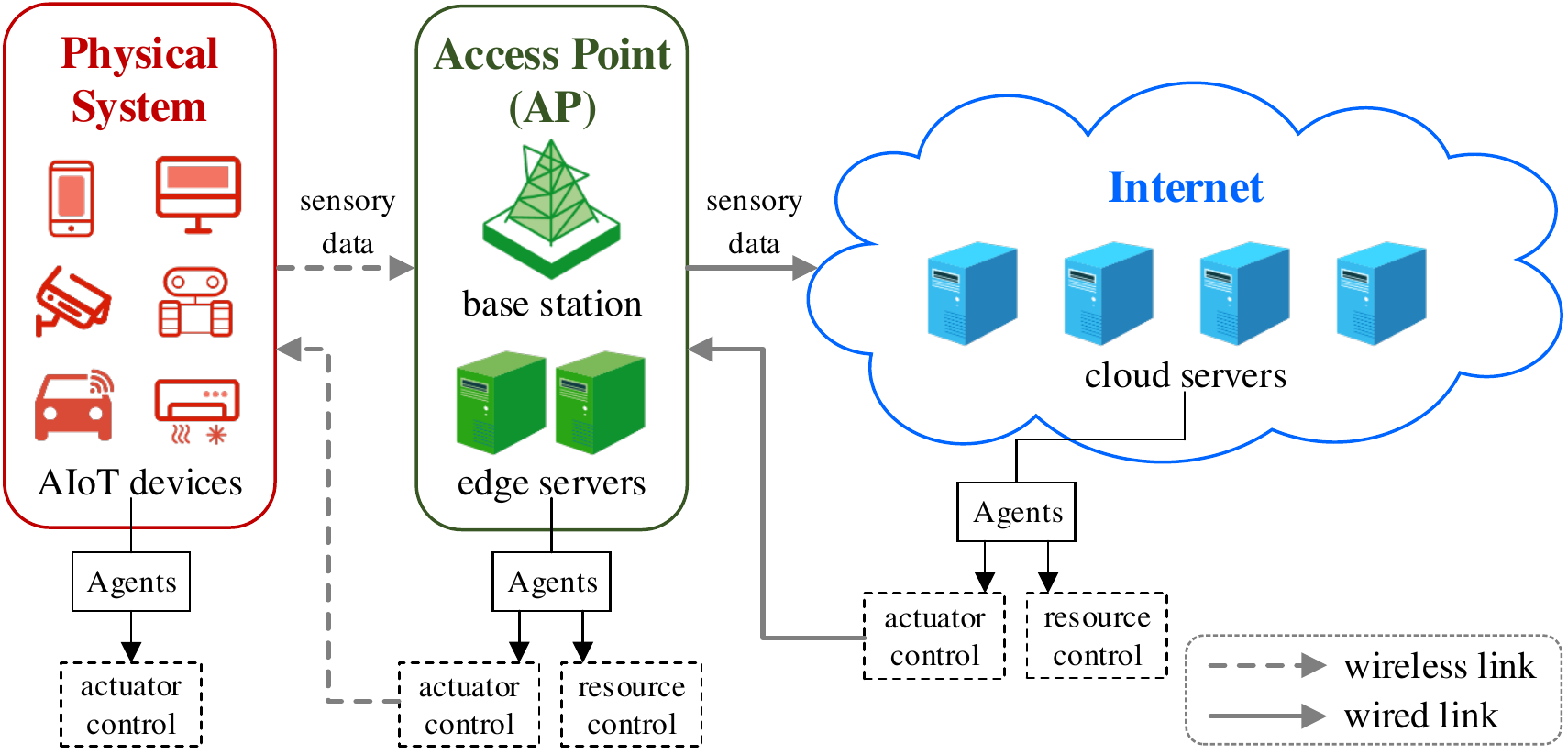}	
	\caption{Autonomous IoT system.}
	\label{intro_iot}
\end{figure}
The AIoT systems provide a dynamic and interactive environment with a number of AIoT devices, which sense the environment and make control decisions to react. As shown in Fig. \ref{intro_iot}, an AIoT system typically includes a physical system where the AIoT devices with sensors and actuators are deployed. The IoT devices are usually connected by wireless networks to an access point (AP) such as a mobile base station (BS), which acts as a gateway to the Internet where the cloud servers are deployed. Moreover, the edge/fog servers with limited data processing and storage capabilities as compared to the cloud servers may be deployed at the APs \cite{Mohammadi2018}. After the IoT devices acquire the sensory data that represent full or partial status of the physical system, they need to process the data and make control decisions for the actuators to react. The data processing tasks can be executed locally at the IoT devices, or remotely at the edge/fog/cloud servers. \par

%As the data processing tasks are offloaded from IoT devices to edge/fog node and further on to cloud servers, the computation delay will decrease with increasing computation capabilities. However, as the sensory data and control decisions need to be transmitted between the IoT devices, edge/fog node and even cloud servers, additional communication delays are incurred.\par

\subsection{Deep Reinforcement Learning}
Reinforcement learning (RL) introduces ambient intelligence into the AIoT systems by providing a class of solution methods to the closed-loop problem of processing the sensory data to generate control decisions to react. Specifically, the agents interact with the environment to learn optimal policies that map status or states to actions \cite{sutton2018reinforcement}. The learning agent must be able to sense the current state of the environment to some extent (e.g., sensing room temperature) and take the corresponding action (e.g., turn thermostat on or off) to affect the new state and the immediate reward so that a long-term reward over extended time period is maximized (e.g., keeping room temperature at a target value). Different from most forms of ML, e.g., supervised learning, the learner is not told which actions to take but must discover which actions yield the most long-term reward by trying them out. \par

While RL has been successfully applied to a variety of domains, it confronts a main challenge when tackling problems with real-world complexity, i.e., the agents must efficiently represent the state of the environment from high-dimensional sensory data, and use these information to learn optimal policies. Therefore, deep reinforcement learning (DRL), in which RL is assisted with deep learning (DL), has been developed to overcome the challenge \cite{mnih2015human}. One of the most famous applications of DRL is AlphaGo, the first computer program which can beat a human professional on a full-sized $19\times19$ board. \par

\begin{table*} [!h]
	\centering
	\renewcommand{\arraystretch}{1.5}
	\caption{Summary of Existing Survey/Overview Works in the Area of IoT and ML/DL/RL/DRL.}
	\label{table_survey}
	{
		\begin{tabular}{|c|c|c|c|c|c|}
			\hline
			\multicolumn{2}{|c|}{} & ML & DL & RL & DRL \\
			\hline
			\multicolumn{2}{|c|}{General-purpose IoT System} &\cite{Sezer2017,Samie2019,Mahdavinejad2018,Cui2018} & \cite{Mohammadi2018} & & \\
			\hline
			Specific IoT Application Areas & Intelligent Transportation Systems& \cite{zantalis2019review} & & &\\
			\cline{2-6}
			& smart city & & \cite{Chen2019} & & \\
			\cline{2-6}
			& smart building & \cite{Qolomany2019} & & &\\
			\cline{2-6}
			& smart grid & \cite{Hossain2019} & \multicolumn{3}{|c|}{\cite{zhang2018review}} \\
			\hline
			Wireless Communications and Networks & IoT specific & \cite{Sharma2019,Alsheikh2014}  & & & \\
			\cline{2-6}
			& General-purpose & \cite{Chen2019b} &\cite{Zappone2019,Mao2018} & & \cite{Luong2019}\\
			\hline
			\multicolumn{2}{|c|}{Cloud/Fog/Edge Computing} & \cite{Abdulkareem2019,Rodrigues2019} & \cite{Chen2019a} & & \cite{Zhu2018} \\
			\hline
		\end{tabular}
	}
\end{table*}

\subsection{Application of DRL in AIoT Systems}
It turns out that the formulation of RL/DRL models for the real-world AIoT systems is not as straightforward as it may appear to be. There are two types of entities in an RL/DRL model as discussed above - environment and agent. Firstly, the \textbf{environment} in RL/DRL can be restricted to reflect only the physical system, or be extended to include the wireless networks, the edge/fog servers and cloud servers as well. This is because that the network and computation performance, such as communication/computation delay, power consumption and network reliability, will have important impacts on the control performance of the physical system. Therefore, the control actions in RL/DRL can be divided into two levels: (physical system) actuator control and (communications/computation) resources control, as shown in Fig. \ref{intro_iot}. The two levels of control can be separated or jointly learned and optimized. Secondly, the \textbf{agent} in RL is a logical concept that makes decisions on action selection. In the AIoT systems, the agent with ambient intelligence can reside in the IoT devices, the edge/fog servers, and/or the cloud servers as shown in Fig. \ref{intro_iot}. The time sensitiveness of the IoT application is an important factor to determine the location of the agents. For example in autonomous driving, images from an autonomous vehicle's camera needs to be processed in real-time to avoid an accident. In this case, the agent should reside locally in the vehicle to make fast decisions, instead of transmitting the sensory data to the cloud and return the predictions back to the vehicle. However, there are many scenarios that it is not easy to determine the optimal locations for the agents, which may involve solving an RL problem in itself. Moreover, when there are multiple agents distributed in the IoT devices, the cooperation of the agents is also an important and challenging issue. \par

%When the agents are in edge node or cloud server, the sensory data need to be transmitted to the agents. Due to the transmission capability of the network, the sensory data might be delayed or lost or at a coarse granularity, so that the RL algorithm need to solve a partially observed problem.

%The control decisions not only include how to control the physical system, but also how to control the communications and computing resources to better support making right decisions to control the physical system. The two levels of control can be separated or jointly optimized. When separated optimized, the reward for the lower-level control is the network performance and/or computation performance such as delay, power consumption or reliability. When jointly optimized, the reward for the lower-level control is also the physical system performance. 
\subsection{Related Overview/Survey Articles}
Although AIoT is a relatively new concept, related research works already exist in IoT and ACS, respectively. In this paper, we will review the state-of-art research, and identify the model and challenges for the application of DRL in AIoT.\par 

The existing overview/survey articles related to this paper are summarized and classified in Table \ref{table_survey}. There are several recent survey articles discussing on the applications of ML/DL in general-purpose IoT systems for data analysis \cite{Sezer2017,Mahdavinejad2018,Mohammadi2018,Samie2019,Cui2018}. In addition, the overview of ML/DL/RL/DRL applications in some specific physical autonomous systems or IoT application areas are provided in \cite{zantalis2019review,Chen2019,Hossain2019,Qolomany2019}. As wireless communications and networks are essential parts of IoT systems, the survey and overview on ML/DL/RL/DRL applications in IoT specific \cite{Sharma2019,Alsheikh2014} or general-purpose wireless networks \cite{Luong2019,Mao2018,Zappone2019} are also listed in Table \ref{table_survey}. Finally, there are also a few overview articles on applying ML/DL/RL/DRL techniques to cloud/edge/fog computing systems, which are important subsystems in the IoT ecosystem. \par

%The work presented in \cite{Sezer2017} surveys the related studies in which ML and big data analytics are applied to achieve ``context-awareness", i.e., understand the environment, situation, or status using data from sensors. The authors in \cite{Mahdavinejad2018} provide a taxonomy of ML algorithms and discusses the challenges of their applications to analyse IoT big data by considering smart cities as the main use case. In \cite{Mohammadi2018}, an overview on using DL to facilitate the analytics and learning in the IoT domain is provided. The work in \cite{Samie2019} discusses the suitability of ML techniques for different processing layers of IoT systems from cloud to fog down to IoT devices. \par

%In \cite{zantalis2019review}, a review of ML techniques and IoT applications in Intelligent Transportation Systems (ITS) 

%Security\cite{Restuccia2018,Zolanvari2019} 

%A review of machine learning applications in smart grids is presented in \cite{zhang2018review}. Different from \cite{zhang2018review}, we focus only on the applications of RL and DRL on the energy management problem with DRES. 

\subsection{Contributions}
This paper focuses on a specific type of ML, i.e., DRL, and its application on a promising type of IoT system, i.e., AIoT. To the best of our knowledge, there are currently no survey/overview articles focusing specifically on the application of DRL in IoT system as shown in Table \ref{table_survey}. Moreover, the concept of AIoT as the future IoT system is relatively new and not adequately addressed in existing literature. The main contributions of this paper lie in the following aspects:

\begin{itemize}
	\item A comprehensive tutorial of DRL is provided. We first explain the relationship between DRL and its two fundamental building blocks, i.e., RL and DL. Then, a tutorial and review on the basic DRL algorithms is given, where the DRL algorithms are classified into two broad categories, i.e., value-based and policy gradient. Different from existing surveys on DRL \cite{Arulkumaran2017,Francois-Lavet2018}, we explain the various DRL algorithms from a unified perspective - the input, output, and loss functions of neural networks (NNs) which are used to approximate the different functions in RL algorithms. Moreover, we discuss the pros and cons of each category of DRL algorithms. Finally, we introduce two types of advanced DRL models and related DRL algorithms that are extremely important for AIoT systems, i.e., partially observable Markov decision process (POMDP)-based DRL and multi-agent (MA) DRL. 
	\item We propose a general DRL model for AIoT systems, where the environment is divided into perception layer, network layer, and application layer according to the IoT architecture. The RL elements including state, action, and reward for each layer as well as the integration of three layers are defined. The relationship between the logical layer and physical locations of an agent in the DRL model is discussed. The general DRL model not only creates a taxonomy to summarize and classify existing works, but also provide a framework to formulate DRL models for future works.    
	\item The emerging research contributions on the applications of DRL in the AIoT systems are reviewed under the umbrella of the proposed general DRL model. First, the general procedure to tackle research problems in this area is introduced. Then, we review and compare the different research works according to (1) whether a basic DRL model or an advanced DRL model such as MA or POMDP is considered; (2) the elements of DRL model as well as the adopted DRL algorithms; (3) the physical locations of the agents and whether centralized or distributed implementation is considered. Finally, we compare between the proposed general DRL model in AIoT and the DRL models in existing literature to derive useful insights for future works.
	\item As a new and emerging research field, there are many challenges and open issues in applying DRL to provide autonomous control in AIoT systems. Four main challenges are identified and discussed, such as incomplete perception problem and delayed control problem. These discussions provide useful information for those readers who seek promising future research directions.
\end{itemize} 

The remainder of the paper is organized as follows. In Section II, we review the RL/DRL methodologies. Section III introduces a general model for RL/DRL in AIoT with a detailed discussion on the key elements. In Section IV, the existing works are surveyed and compared. The challenges and open issues are identified and highlighted in Section V. Finally, the conclusion is given in Section VI.\par

\section{Overview of Deep Reinforcement Learning}

DRL has two fundamental building blocks - RL and DL, the basic concepts of which are introduced in Appendix A and B, respectively. In RL, a large amount of memory is usually required to store the value functions and Q-functions. In most of the real-world problems, the state sets are large, sometimes infinite, which makes it impossible to store the value functions or Q-functions in the form of tables. Therefore, the trial-and-error interaction with the environment is hard to be learned due to the formidable computation complexity and storage capacity requirements. This is where DL comes into the picture - some functions of RL such as value/Q-functions or policy functions are approximated with a smaller set of parameters by the application of DL. The combination of RL and DL results in the more powerful DRL.\par

\begin{figure}[!t]
	\centering
	\subfigure[General value-based methods for DRL.]{
		\begin{minipage}[b]{0.35\textwidth}
			\includegraphics[width=1\textwidth]{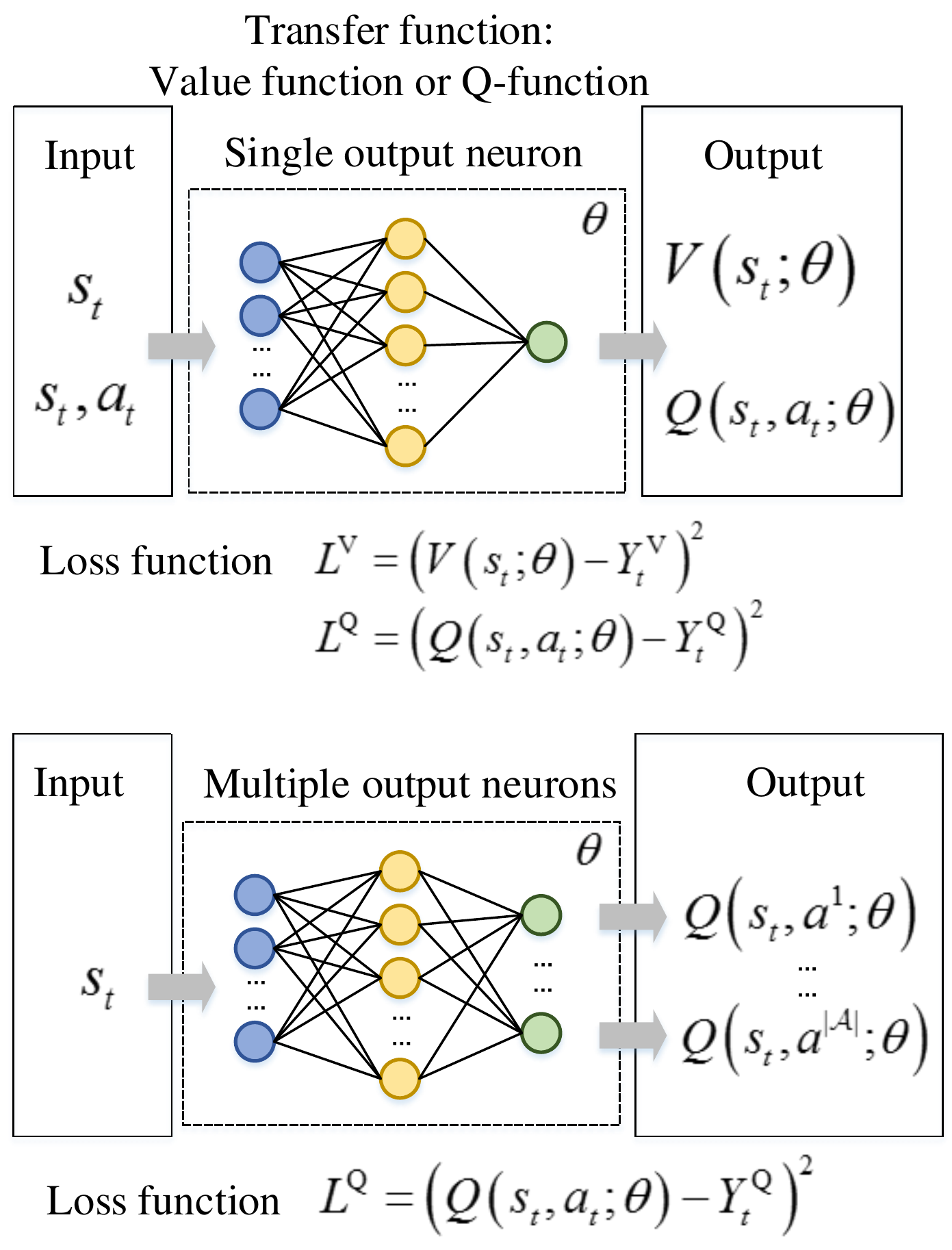}
		\end{minipage}
		\label{generalvalue}
	}
%	\subfigure[General value-based methods for DRL - Multiple output neurons]{
%		\begin{minipage}[b]{0.4\textwidth}
%			\includegraphics[width=1\textwidth]{nn2.pdf}
%		\end{minipage}
%		\label{nn2}
%	} 
	\subfigure[General policy gradient methods for DRL.]{
	\begin{minipage}[b]{0.35\textwidth}
		\includegraphics[width=1\textwidth]{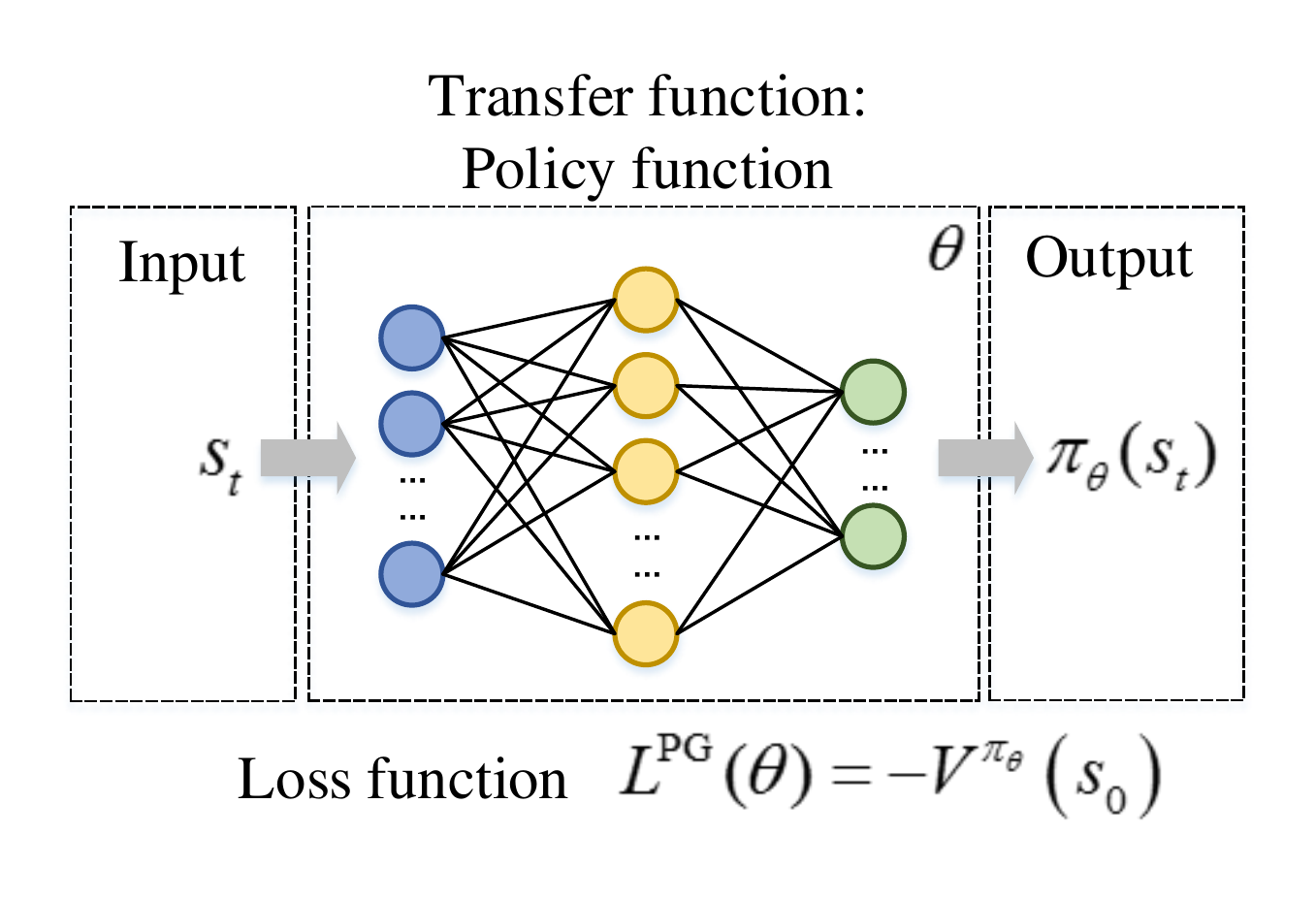}
	\end{minipage}
	\label{generalpg}
} 
	\caption{General methods for DRL.}
	\label{nn}
\end{figure}

In this section, we first classify the basic DRL algorithms into two broad categories, i.e., value-based and policy gradient methods, according to whether value/Q-functions or policy functions are approximated by NN as shown in Fig.\ref{nn}. The policy gradient methods are further discussed from three aspects:
\begin{itemize}
	\item Based on the different \emph{natures of the approximated policy functions}, we introduce stochastic policy gradient (SPG) versus deterministic policy gradient (DPG) methods;
	\item Based on the different \emph{ways of policy evaluation}, we introduce Monte Carlo policy gradient versus actor-critic methods;
	\item Based on the different \emph{learning or parameter update techniques}, we introduce simple policy gradient versus natural policy gradient (NPG) methods.
\end{itemize}

%\begin{figure}[!t]
%	\centering
%	\includegraphics[width=0.4\textwidth]{generalpg.pdf}
%	\caption{General policy gradient methods for DRL}
%	\label{generalpg}
%\end{figure}

\begin{figure*}[!t]
	\centering
	\includegraphics[width=0.7\textwidth]{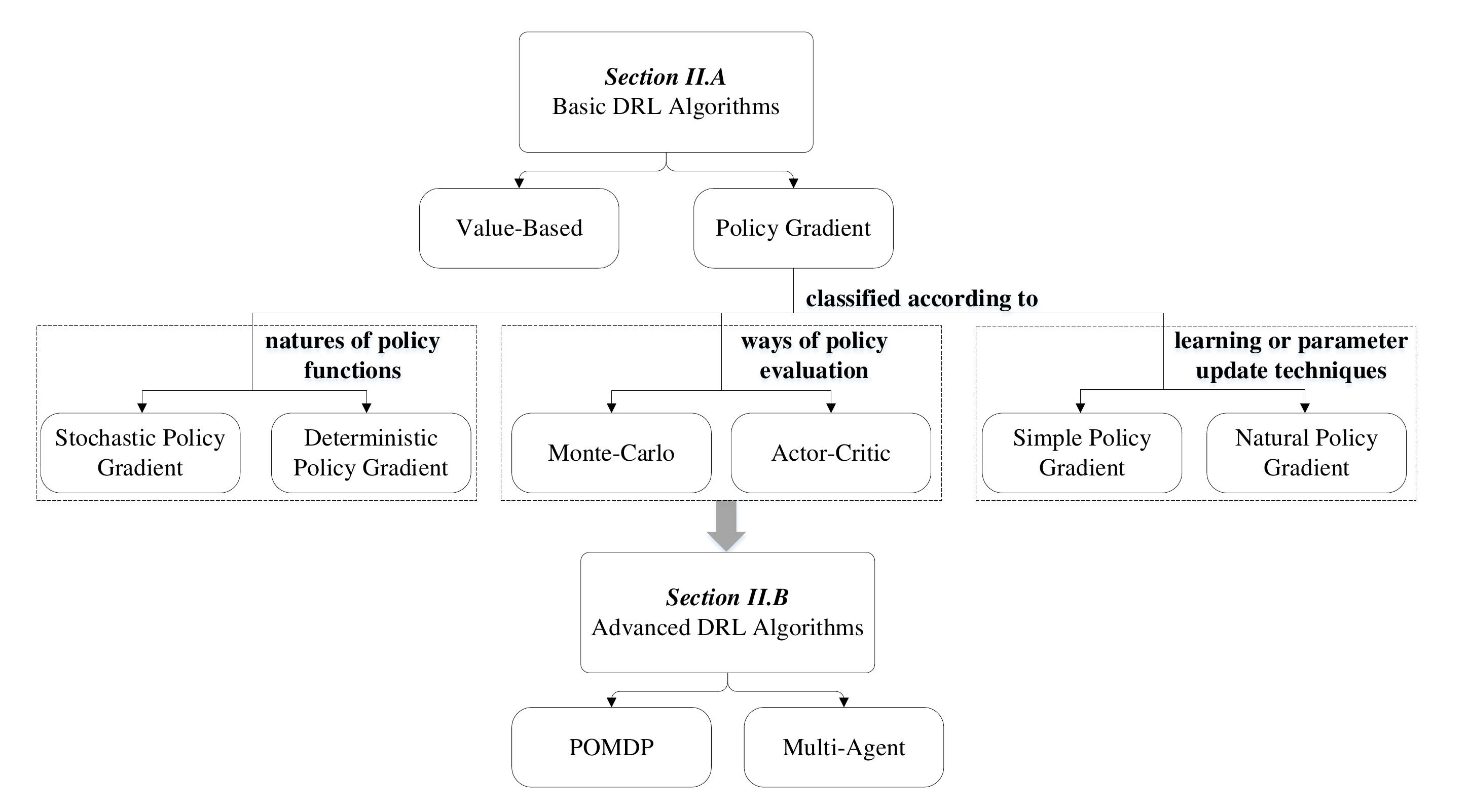}
	\caption{Organization of section II.}
	\label{drl_structure}
\end{figure*}

Then, we introduce two types of advanced DRL algorithms, i.e., POMDP-based DRL and MA-based DRL, that are envisioned to be extremely useful in addressing the open issues in AIoT. The organization of Section II is illustrated in Fig.\ref{drl_structure}.\par

%\begin{figure}
%	\centering
%	%\label{fig:subfig:b} %% label for second subfigure
%	\includegraphics[width=2.6in]{value_policy_ac.pdf}
%	\caption{Classification of common methods for DRL}
%	%\label{fig:subfig} %% label for entire figure
%\end{figure}

\subsection{Basic DRL Algorithms}
\subsubsection{Value-Based Methods}
In value-based methods for DRL as illustrated in Fig. \ref{generalvalue}, the states $s_{t}\in\mathcal{S}$ or state-action pairs $(s_{t},a_{t})\in\mathcal{S}\times\mathcal{A}$ are used as inputs to NNs, while Q-functions $Q^{\pi}(s_{t},a_{t})$ or value functions $V^{\pi}(s_{t})$ are approximated by parameters $\theta$ of NNs. An NN returns the approximated Q-functions or value functions for the input states or state-action pairs. There can be a single output neuron or multiple output neurons as shown in Fig. \ref{generalvalue}. For the former case, the output can be either $V^{\pi}(s_{t})$ or $Q^{\pi}(s_{t},a_{t})$ corresponding to the input $s_{t}$ or $(s_{t},a_{t})$. For the latter case, the outputs are the Q-functions for state $s_{t}$ combined with every action, i.e., $Q^{\pi}(s_{t},a^{1}),\cdots,Q^{\pi}(s_{t},a^{|\mathcal{A}|})$.  \par

To derive the loss functions, $Y_{t}^{\mathrm{Q}}$ and $Y_{t}^{\mathrm{V}}$ are defined as the target values of Q-functions and value functions, respectively. The regression loss  

\begin{equation}
L^{\mathrm{Q}}=\left(Q\left(s_{t}, a_{t}; \theta \right)-Y_{t}^{\mathrm{Q}}\right)^{2},
\label{equ5}
\end{equation}

\noindent or

\begin{equation}
L^{\mathrm{V}}=\left(V\left(s_{t}; \theta \right)-Y_{t}^{\mathrm{V}}\right)^{2},
\label{equ6}
\end{equation}

\noindent can be used to evaluate how well the NN approximate Q-functions or value functions in value-based methods.\par

%\begin{figure}
%	\centering 
%	\includegraphics[width=2.6in]{nn.pdf}
%	\caption{NN fitted Q functions or value functions}
%	\label{nn} %% label for entire figure 
%	\end{figure}
%
%\begin{figure}
%	\centering
%	%\label{fig:subfig:b} %% label for second subfigure
%	\includegraphics[width=2.6in]{dqn.pdf}
%	\caption{Illustration of DQN}
%	\label{dqn} %% label for entire figure
%\end{figure}
%
%\begin{figure}
%	\centering
%	%\label{fig:subfig:b} %% label for second subfigure
%	\includegraphics[width=2.6in]{ddqn.pdf}
%	\caption{Illustration of DDQN}
%	\label{ddqn} %% label for entire figure
%\end{figure}

\begin{figure}[!t]
	\centering
	\subfigure[DQN ]{
	\begin{minipage}[b]{0.45\textwidth}
		\includegraphics[width=1\textwidth]{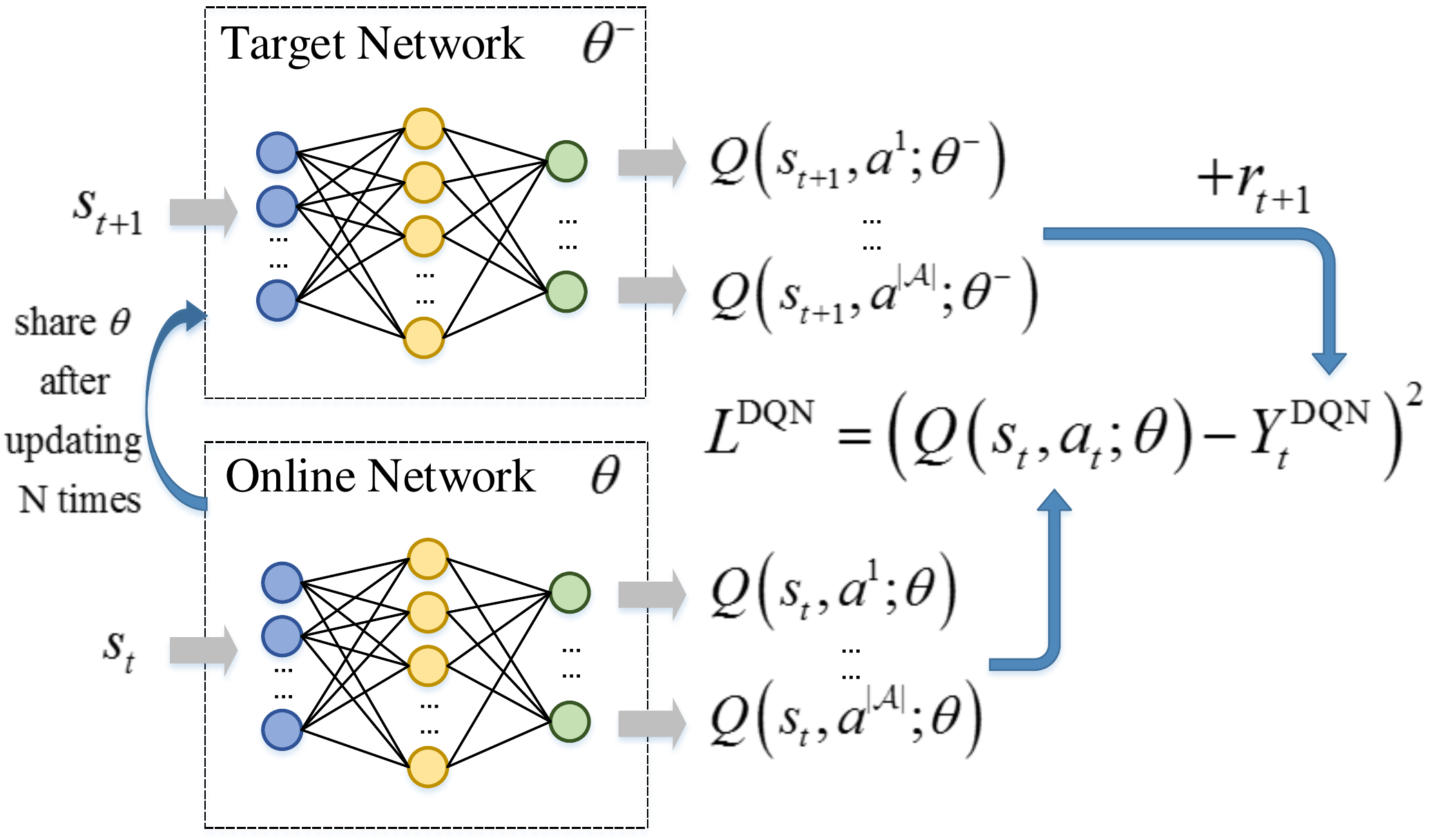}
	\end{minipage}
	\label{dqn}}
	\subfigure[DDQN ]{
	\begin{minipage}[b]{0.45\textwidth}
		\includegraphics[width=1\textwidth]{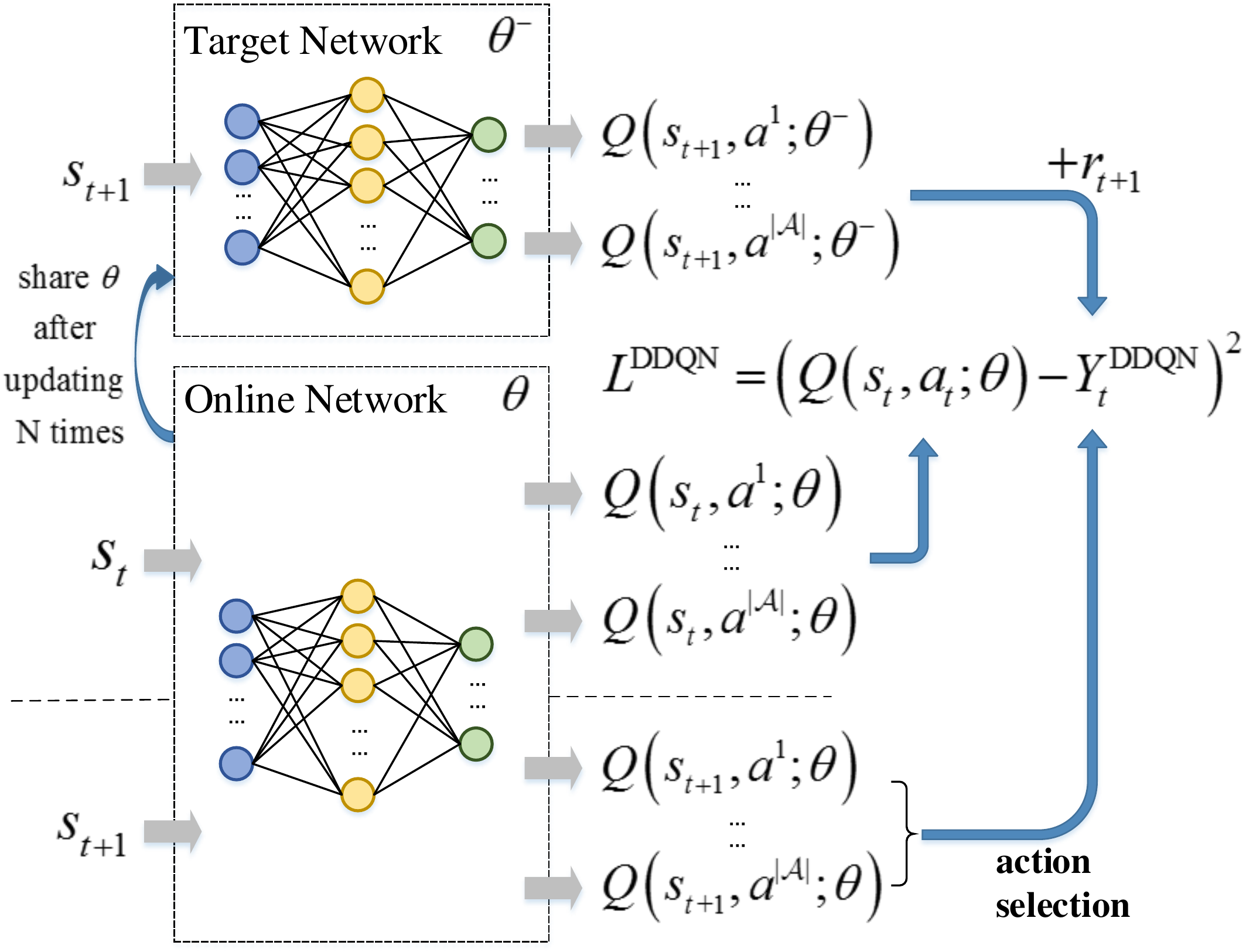}
	\end{minipage}
	\label{ddqn}}
	\caption{DQN and DDQN with target networks}
\label{dqns}

\end{figure}

\paragraph{Deep Q-Networks}
Based on the idea of NN fitted Q-functions, the Deep Q-networks (DQN) algorithm is introduced by Mnih \textit{et al.} in 2015 to obtain strong ability in ATARI games \cite{mnih2015human}. The illustration of DQN is shown in Fig. \ref{dqn}. The NN in DQN takes a state as input, and returns approximated Q-functions for every action under the input state. \par

In DQN, the algorithm first randomly initialize the parameters of networks as $\theta_{0}$. The target Q-function $Y_{t}^{\text{DQN}}$ is given by \eqref{equ7} according to Bellman equation as

\begin{equation}
Y_{t}^{\text{DQN}}=r_{t+1}+\gamma \max _{a_{t+1}} Q\left(s_{t+1}, a_{t+1}; \theta \right),
\label{equ7}
\end{equation}

\noindent where the subscripts $t$ or $t+1$ refer to the values of corresponding variables at the $t^{th}$ or $(t+1)^{th}$ iteration.

%\begin{figure*}[!t]
%	\centering
%	\subfigure[General value-based methods]{
%		\begin{minipage}[b]{0.48\textwidth}
%			\includegraphics[width=1\textwidth]{nn.pdf}
%		\end{minipage}
%		\label{nn}
%	}
%	\subfigure[DQN]{
%		\begin{minipage}[b]{0.48\textwidth}
%			\includegraphics[width=1\textwidth]{dqn.pdf}
%		\end{minipage}
%		\label{dqn}
%	}
%	
%	\subfigure[DDQN]{
%		\begin{minipage}[b]{0.48\textwidth}
%			\includegraphics[width=1\textwidth]{ddqn.pdf}
%		\end{minipage}
%		\label{ddqn}
%	} 
%	\caption{Value-based methods for DRL}
%\end{figure*}

%\begin{figure}[!t]
%		\centering
%	\includegraphics[width=0.45\textwidth]{dqn.pdf}
%	\caption{DQN}
%	\label{dqn}
%\end{figure}
%
%\begin{figure}[!t]
%	\centering
%	\includegraphics[width=0.45\textwidth]{ddqn.pdf}
%	\caption{DDQN}
%	\label{ddqn}
%\end{figure}

The parameters in DQN are updated by minimizing the loss function $L^{\text{DQN}}$, which can be derived from \eqref{equ5} by replacing $Y_{t}^{\text{Q}}$ with $Y_{t}^{\text{DQN}}$. \par

%	
%	\begin{equation}
%	L^{\text{DQN}}=\left(Q\left(s_{t}, a_{t}; \theta_{t}\right)-Y_{t}^{\text{DQN}}\right)^{2}.
%	\label{equ8}
%	\end{equation}
%	\begin{equation}
%	L^{\text{DDQN}}=\left(Q\left(s_{t}, a_{t}; \theta_{t}\right)-Y_{t}^{\text{DDQN}}\right)^{2}.
%	\label{equ8}
%	\end{equation}

By applying stochastic gradient descent, the parameters are updated as 
\begin{equation}
\theta \leftarrow \theta+\alpha\left(Y_{t}^{\text{DQN}}-Q\left(s_{t}, a_{t} ; \theta \right)\right) \nabla_{\theta} Q\left(s_{t}, a_{t} ; \theta\right),
\label{equ8}
\end{equation}
\noindent where $\alpha$ is the learning rate.\par

In order to deal with the limitations of DRL, two important techniques, freezing target networks and experience replay, are applied in DQN. To make the training process more stable and controllable, the target networks, whose parameters $\theta_{t}^{-}$ are kept fixed in a time period, are used to evaluate the Q-function of the next state, i.e., instead of \eqref{equ7}, we have
\begin{equation}
Y_{t}^{\text{DQN}}=r_{t+1}+\gamma \max _{a_{t+1}} Q\left(s_{t+1}, a_{t+1}; \theta^{-}\right).
\label{equ9}
\end{equation}
\noindent The parameters of online network $\theta_{t}$ are updated after each iteration. After a certain number of iterations, the online network shares its parameters to the target network. This reduces the risk of divergence and prevents the instabilities resulted from the too quick propagation. \par

To perform experience replay, the experience of the agent at each time step is stored in a data set. Then, the updates are made on this data set, which removes correlations in the observation sequence and smooths over changes in the data distribution. This technique allows the updates to cover a wide range state-action space and provides more possibility to make larger updates of the parameters. \par

\paragraph{Double DQN}
In DQN, the Q-function evaluated by target networks is used both to select and evaluate an action, which makes it more likely to overestimate the Q-function of an action. The estimating error will become larger if there are more actions. To overcome this problem, Hasselt \textit{et al.} proposed a Double DQN (DDQN) method in 2016, where two sets of parameters are used to derive the target value $Y_{t}^{\text{ DDQN }}$ as shown in Fig. \ref{ddqn} \cite{van2016deep}. Compared with \eqref{equ7}, the target Q-value in DDQN can be rewritten as
\begin{equation}
\label{equ10}
 Y_{t}^{\text { DDQN }} =  r_{t+1}+\gamma Q(s_{t+1},\arg\max_{a_{t+1}} Q(s_{t+1}, a_{t+1} ;\theta)  ; \theta^{-}),
\end{equation}
\noindent where the selection of the action is due to the parameters $\theta$ in online network and the evaluation of the current action is due to the parameters $\theta^{-}$ in target network. This means there will be less overestimation of the Q-Learning values and more stability to improve the performance of the DRL methods \cite{franccois2018introduction}. The loss function $L^{\text{DDQN}}$ can be derived from \eqref{equ5} by replacing $Y_{t}^{\text{Q}}$ with $Y_{t}^{\text{DDQN}}$ and the parameters can be updated accordingly. DDQN algorithm gets the benefit of double Q-Learning and keeps the rest of DQN algorithm.\par \par

Apart from DQN and DDQN, there are also other value-based methods, some of which are developed based on DQN and DDQN with some further improvement, such as DDQN with Proportional Prioritization \cite{schaul2015prioritized}, and DDQN with duel architecture \cite{wang2015dueling}.\par

\newtheorem{remark}{Remark}
\begin{remark}[\textbf{Pros and cons of value-based DRL methods}]	
Although DQN and its improved versions have been widely adopted in existing literature as discussed in Section IV - mainly due to their relative simplicity and good performance, there are some limitations with value-based DRL methods. First, it cannot solve RL problems with large or continuous action space. Second, it cannot solve RL problems where the optimal policy is stochastic requiring specific probabilities. Since value-based method can only learn deterministic policies, the majority of the algorithms are off-policy, such as DQN.
\end{remark}

\subsubsection{Policy Gradient Methods}

According to a policy $\pi$, action $a$ is selected when the environment is in state $s$. In policy gradient methods, NNs can be applied to directly approximate a policy as a function of state, i.e., $\pi_{\theta}(s)$. As shown in Fig. \ref{generalpg}, the states are used as inputs to the NNs, while policy $\pi$ is approximated by parameters $\theta$ of NNs as $\pi_\theta$. \par
%{\color{blue}An NN returns an approximated policy with each neuron in the output layer corresponding to the probability of selecting an action under the input state.}\par

%\begin{figure}[!t]
%	\centering
%	%	\label{policygradient}
%	\subfigure[SPG]{
%		\begin{minipage}[b]{0.4\textwidth}
%			\includegraphics[width=1\textwidth]{mcpg.pdf}
%		\end{minipage}		
%		\label{spg}
%	}
%	\subfigure[DPG]{
%		\begin{minipage}[b]{0.4\textwidth}
%			\includegraphics[width=1\textwidth]{dpg.pdf}
%		\end{minipage}		
%		\label{dpg}
%	}	
%	\caption{General policy gradient methods for DRL}
%	\label{policygradient}
%\end{figure}

%In contrast to DQN, the policy gradient methods for DRL is a direct mapping from state $s$ to action $a$, which leads to better convergence properties and higher efficiency in high-dimensional or continuous action spaces \cite{franccois2018introduction}.\par

To evaluate the performance of the current policy, the objective function is defined as
\begin{equation}
{J}(\theta)=V^{\pi_\theta}(s_{0})=\mathrm{E}_{\tau_{s_{0}}\sim\pi_{\theta}}\left[ G(\tau_{s_{0}})\right], \forall s_{0}\in\mathcal{S},
\label{equ11}
\end{equation}
\noindent where $V^{\pi_\theta}(s_{0})$ is the value function of policy $\pi_\theta$ as shown in (\ref{equ1}), and $\tau_{s_{0}}$ refers to the sampling trajectory with an initial state $s_{0}$. If we can find the parameters $\theta$ for policy $\pi_{\theta}$ so that the objective function ${J}(\theta)$ is maximized, we can solve the problem. The basic idea of policy gradient methods are to adjust the parameters in the direction of greater expected reward \cite{amari1998natural}. For this purpose, we can set the loss function of NN to be
\begin{equation}
L^{\mathrm{PG}}(\theta)=-{J}(\theta)=-V^{\pi_\theta}(s_{0}).
\label{equ17}
\end{equation}

In order to update the parameters, we need to express the gradient of ${J}(\theta)$ with respect to parameter $\theta$ as an expectation of stochastic estimates based on \eqref{equ11}. As mentioned in Section II-A, the policy in RL can be classified into two categories, i.e., the stochastic policy and the deterministic policy. Hence, the SPG method and DPG method are correspondingly discussed below.\par

\paragraph{Stochastic Policy Gradients vs. Deterministic Policy Gradient}
By applying DRL, a stochastic policy is approximated as $\pi_{\theta}=\pi(a_{t}|s_{t};\theta)$, which gives the probability of a specific action $a$ is taken in a specific state $s$, when the agent follows the policy parameterized by $\theta$. The policy parameters are usually the weights and bias of a NN \cite{franccois2018introduction}. For a DRL model with discrete state/action spaces, Softmax function is a typical probability density function. In the cases of continuous state/action spaces, Gaussian distribution is generally used to characterize the policy. An NN is applied to approximate the mean, and a set of parameters specifies the standard deviation of the Gaussian distribution \cite{kakade2002natural}\cite{schulman2015trust}.\par
According to the policy gradient theorem, we have
\begin{equation}
\nabla\mathrm{E}_{\tau_{s_{0}}\sim\pi_{\theta}} \!\! \left[ G(\tau_{s_{0}})\right] \!\! = \!\! \mathrm{E}_{\tau_{s_{t}}\sim\pi_{\theta}} \! \left[G(\tau_{s_{t}}) \nabla_{\theta} \! \log\pi(a_{t}|s_{t}; \! \theta)\right] \! .
\label{equ12}
\end{equation}
By applying stochastic gradient descent, the parameters are updated as 
\begin{equation}
\theta \leftarrow \theta + \alpha G(\tau_{s_{t}}) \nabla_{\theta} \log\pi(a_{t}|s_{t};\theta),
\label{equ13}
\end{equation}
\noindent where $\alpha$ is the learning rate. In this way, $\theta$ is adjusted to enlarge the probability of trajectory $G(\tau_{s_{t}})$ with higher total reward. \par
From the perspective of NN, we give the loss function of SPG algorithm as
\begin{equation}
{L}^{\mathrm{SPG}}(\theta)=-G(\tau_{s_{t}})\log\pi(a_{t}|s_{t};\theta).
\label{equ14}
\end{equation} \par

Different from SPG where the policy is modeled as a probability distribution over actions, DPG models the policy as a deterministic decision, i.e., $\pi_{\theta}=\pi(s_{t};\theta)$. According to the objective function given in \eqref{equ11} and the DPG theorem, we have
\begin{equation}
\nabla_{\theta}{J}(\theta)= \mathrm{E}_{s \sim \rho^{\pi_{\theta}}}[\nabla_{\theta} \pi(s_t;\theta) \nabla_{a} Q^{\pi_{\theta}}(s_t, a_t;\phi)|_{a=\pi(s_t;\theta)}],
\end{equation}
\noindent where  the policy improvement is decomposed into the gradient of the Q-function with respect to actions, and the gradient of the policy with respect to the policy parameters. $\rho^{\pi_{\theta}}$ is the state distribution following policy $\pi_{\theta}$. Thus, the parameters are updated as
\begin{equation}
\theta \leftarrow \theta \!+\! \alpha \left[\nabla_{\theta} \pi(s_t;\theta) \nabla_{a} Q^{\pi_{\theta}} \!\!\! \left.(s_t, a_t;\phi)\right|_{a=\pi(s_t;\theta)} \!\right].
\label{equ18}
\end{equation}
A differentiable function $Q^{w}(s_t,a_t;\phi)$ can be used as an approximator of ${Q^{{\pi_\theta }}}(s_t,a_t;\phi)$, and then the gradient ${\nabla_a}{Q^{{\pi_\theta }}}(s_t,a_t;\phi)$ can be replaced by $\nabla_{a} Q^{w}(s_t,a_t;\phi)$. The approximator is compatible with the deterministic policy, and ${\nabla _a}{Q^w}{\left. {({s_t},{a_t};\phi )} \right|_{a = \pi ({s_t};\theta )}}$ is achieved as $ {\nabla _\theta }\pi {({s_t};\theta )^ \top }w$ \cite{silver2014deterministic}.\par
From the perspective of NN, the loss function of DPG algorithm is set as
%\begin{equation}
%{L}^{\mathrm{DPG}}(\theta) = -\pi_{\theta}(s) \nabla_{a} Q^{\pi_{\theta}} \!\!\! \left.(s, a)\right|_{a=\pi_{\theta}(s)}.
%\end{equation}

\begin{equation}
\label{equ19}
{L}^{\mathrm{DPG}}(\theta) = -\pi(s_t;\theta) \nabla_{a} Q^{\pi_{\theta}} \!\!\! \left.(s_t, a_t;\phi)\right|_{a=\pi(s_t;\theta)}.
\end{equation}

\paragraph{Monte Carlo Policy Gradient vs. Actor-Critic}

\begin{figure}[!t]
	\centering
	\subfigure[Monte Carlo policy gradient method.]{
		\begin{minipage}[b]{0.35\textwidth}
			\includegraphics[width=1\textwidth]{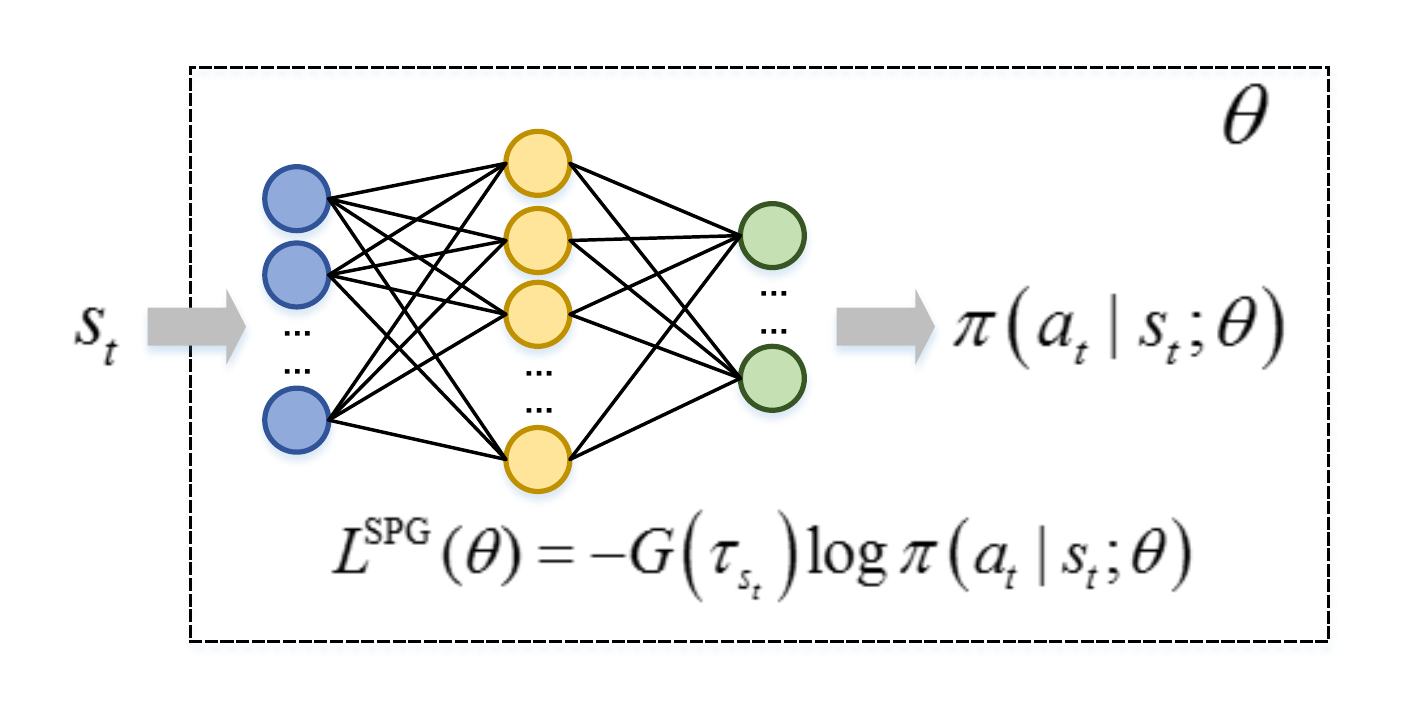}
		\end{minipage}
		\label{mcpg}}
	\subfigure[Actor-critic methods for SPG.]{
		\begin{minipage}[b]{0.35\textwidth}
			\includegraphics[width=1\textwidth]{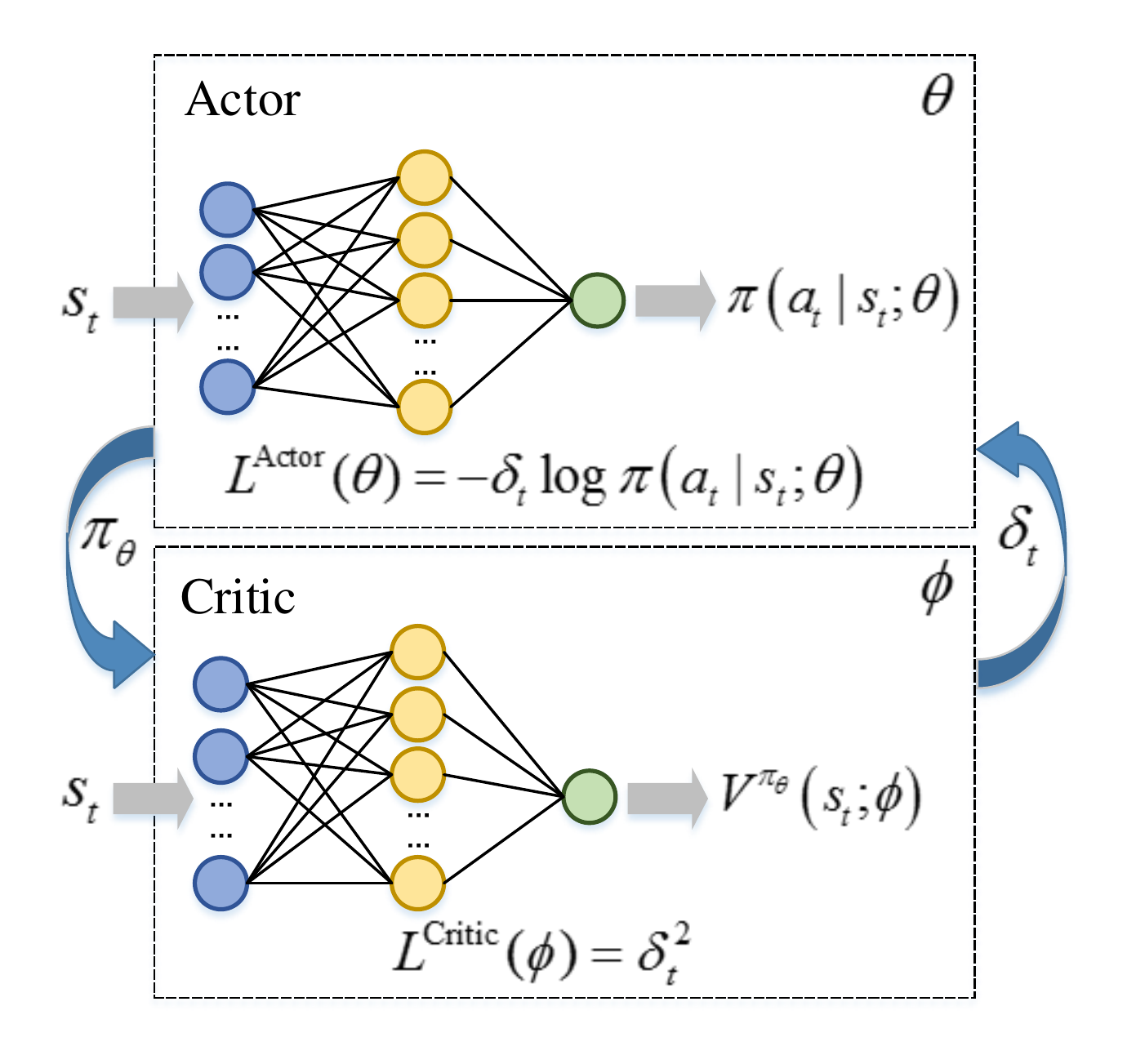}
		\end{minipage}
		\label{actorcritic}}
	\subfigure[Actor-critic methods for DPG.]{
		\begin{minipage}[b]{0.33\textwidth}
			\includegraphics[width=1\textwidth]{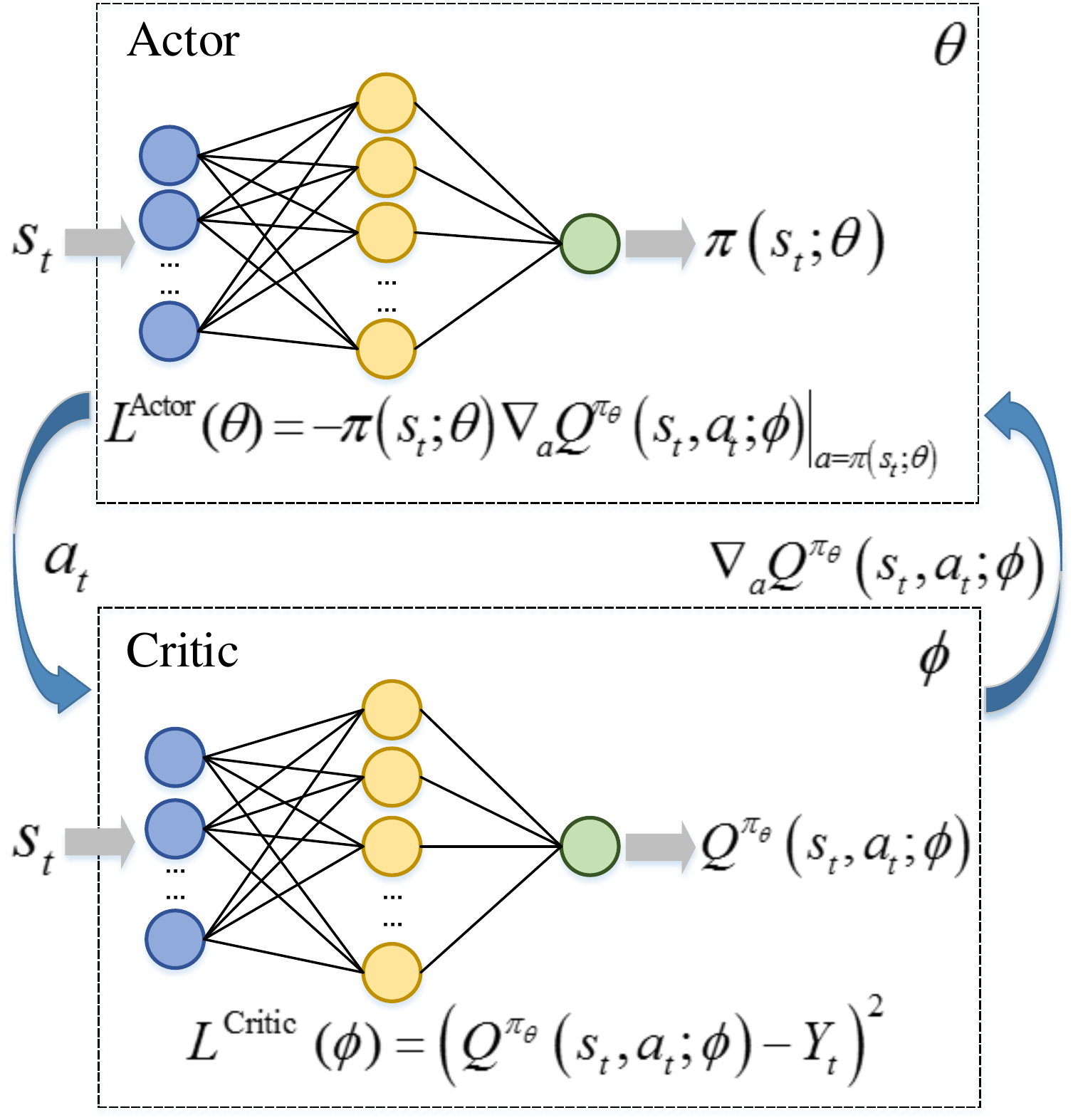}
		\end{minipage}
		\label{dpg}}
	\caption{Monte Carlo policy gradient versus actor-critic methods.}
	\label{pgac}
	
\end{figure}

In \eqref{equ14} and \eqref{equ19}, the value of $G(\tau_{s_{t}})$ and $\nabla_{a} Q^{\pi_{\theta}}$ need to be derived to update the policy parameters $\theta$ in SPG and DPG, respectively. For SPG, this can be achieved either by Monte Carlo policy gradient method or actor-critic method, as is illustrated in Fig. \ref{mcpg} and Fig. \ref{actorcritic}, respectively. For DPG, this is normally achieved by actor-critic method as is shown in Fig. \ref{dpg}.\par

The Monte Carlo policy gradient method tries to evaluate $G(\tau_{s_{t}})$ through Monte Carlo simulation. A typical Monte Carlo algorithm of the SPG methods is the REINFORCE algorithm proposed in \cite{williams1992simple}. Based on the Monte Carlo approach, a trajectory $\tau_{s_{0}}$ is firstly sampled by running the current policy from an initial state $s_{0}$. Then for each time step $t=0,1,\cdots,T$, the total reward $G(\tau_{s_{t}})$ starting from time step $t$ is calculated, which is multiplied with the policy gradient $\nabla_{\theta} \log\pi(a_{t}|s_{t};\theta)$ to update the parameters $\theta$ according to (\ref{equ13}). The above procedure is repeated over multiple runs, while in each run a different trajectory is sampled. \par

%\begin{figure}[!t]
%	\centering
%	\includegraphics[width=0.4\textwidth]{mcpg.pdf}
%	\caption{Monte Carlo policy gradient method.}
%	\label{mcpg}
%\end{figure}
%
%\begin{figure}[!t]
%	\centering
%	\includegraphics[width=0.4\textwidth]{ac_arch.pdf}
%	\caption{Actor-critic methods for SPG.}
%	\label{actorcritic}
%\end{figure}
%
%\begin{figure}[!t]
%	\centering
%	\includegraphics[width=0.4\textwidth]{dpg.pdf}
%	\caption{Actor-critic methods for DPG.}
%	\label{dpg}
%\end{figure}

Moreover, in order to reduce the variance of the policy gradient, a baseline function $b(s_t)$ which is independent of $a_t$ is introduced. Based on this, the REINFORCE algorithm with baseline is introduced, and the loss function of it can be formulated as 
\begin{equation}
{L}^{\mathrm{SPG\_BASE}}(\theta)=-(G(\tau_{s_{t}})-b(s_t))\log\pi(a_{t}|s_{t};\theta).
\label{equ15}
\end{equation}

\newtheorem{remark2}[remark]{Remark}
\begin{remark2}[\textbf{Pros and cons of Monte Carlo policy gradient DRL methods}]	
In contrast to value-based DRL methods, the policy gradient methods for DRL is a direct mapping from state to action, which leads to better convergence properties and higher efficiency in high-dimensional or continuous action spaces \cite{franccois2018introduction}. Moreover, it can learn stochastic policies, which have better performance than deterministic policies in some situations. However, Monte Carlo policy gradient methods suffer from high variance of estimations. As on-policy methods, they require on-policy samples, which made them very sample intensive.
\end{remark2}

Actor-critic methods are well-known for combining the advantages of both Monte Carlo policy gradient and value-based methods, and they have been widely studied in DRL. As illustrated in Fig. \ref{actorcritic} and Fig. \ref{dpg}, an actor-critic method is generally realized by two NNs, i.e., the actor network and the critic network, which share parameters with each other. The actor network is similar to the NN of the policy gradient method, while the critic network is similar to the NN of the value-based method. During the learning process, the critic updates the parameters of value functions, i.e, $\phi$, according to the policy given by the actor. Meanwhile, the actor updates the parameters of policy, i.e., $\theta$, according to the value functions evaluated by the critic. Generally, two learning rates are required to be predefined respectively for the updates of $\phi$ and $\theta$ \cite{konda2000actor}.\par

In the actor-critic method for SPG, the critic network is used obtain the value of $G(\tau_{s_{t}})$ in \eqref{equ14}. Specifically, the baseline $b(s_t)$ in \eqref{equ15} is set to the value function $V^{\pi_{\theta}}\left(s_{t};\phi\right)$, which is approximated by the critic network with a loss function as given in \eqref{equ6}. In a state $s_t$, the agent selects an action $a_t$ according to the current policy $\pi_{\theta}$ given by the actor network, receives a reward $r_{t+1}$, and the state transits to $s_{t+1}$. Similar to \eqref{equ6} in value-based method, the loss function for the critic network can be expressed as
\begin{equation}
{L}^{\rm{Critic}}(\phi)={\delta_t}^{2},
\end{equation}
\noindent where
\begin{equation}
\delta_t = r_{t+1} +\gamma V^{\pi_{\theta}}\left(s_{t+1};\phi\right)-V^{\pi_{\theta}}(s_t;\phi),
\label{equ16}
\end{equation}

Similar to \eqref{equ8} in DQN, the parameters of the critic network are updated as

\begin{equation}
\phi \leftarrow \phi + \beta \delta_t \nabla_{\phi} V^{\pi_{\theta}}(s_t;\phi),
\end{equation} 
\noindent where $\beta$ is the learning rate for the critic.\par

Note that $G(\tau_{s_{t}})$ is an estimate of $Q^{\pi_{\theta}}(s_{t},a_{t};\phi)=r_{t+1} +\gamma V^{\pi_{\theta}}\left(s_{t+1};\phi\right)$. Therefore, given the value functions evaluated by the critic network, the value of $G(\tau_{s_{t}})-b(s_t)$ in \eqref{equ15} can be replaced by $\delta_t$ in \eqref{equ16}, which can be seen as an estimate of the advantage of action $a_{t}$ in state $s_{t}$ \cite{wang2016sample}. The loss function of the actor network can be defined similar to \eqref{equ15}, i.e.,
\begin{equation}
{L}^{\rm{Actor}}(\theta) =-\delta_t \log \pi\left(a_{t} | s_{t};\theta\right).
\end{equation} \par
Similar to \eqref{equ13} in the policy gradient method, the parameters of the actor network are updated as
\begin{equation}
\theta \leftarrow \theta + \alpha \delta_t \nabla_{\theta} \log \pi(a_t | s_t;\theta),
\end{equation}
\noindent where $\alpha$ is the learning rate for the actor. \par

Through the update processes in the actor-critic algorithm, the critic can make the approximation of value functions more accurately, while the actor can choose better action to get higher reward.\par

Typical actor-critic methods for SPG include the asynchronous advantage actor-critic (A3C) algorithm and soft actor-critic (SAC). The former mainly focuses on the parallel training of multiple actors that share global parameters \cite{mnih2016asynchronous}. The latter involves a soft Q-function, a tractable stochastic policy and off-policy updates \cite{haarnoja2018soft}. SAC achieves good performance on a range of continuous control tasks.\par

One typical actor-critic method for DPG is the deep deterministic policy gradient (DDPG) algorithm. The DDPG algorithm is a model-free off-policy actor-critic algorithm, which combines the ideas of DPG and DQN. It is first proposed by Lillicrap \textit{et al.} in 2015 \cite{lillicrap2015continuous}. Besides the online critic network $Q$ with parameters $\phi$, and the online actor network $\pi$ with parameters $\theta$, the target networks $Q^{\prime}$ and $\pi^{\prime}$ in the DDPG algorithm are specified with $\phi^{\prime}$ and $\theta^{^{\prime}}$, respectively. The parameters of these four NNs are required to be updated in the learning process. The gradient ${\nabla_a}{Q^{{\pi_\theta }}}(s_t,a_t;\phi)$ is obtained by the critic network.\par

Based on DDPG, several algorithms are proposed in recent years, such as Distributed Distributional Deep Deterministic Policy Gradients (D4PG) \cite{barth2018distributed}, Twin Delayed Deep Deterministic (TD3) \cite{fujimoto2018addressing}, Multi-Agent DDPG (MADDPG) \cite{lowe2017multi}, and Recurrent Deterministic Policy Gradients (RDPG) \cite{heess2015memory}.\par

\newtheorem{remark3}[remark]{Remark}
\begin{remark3}[\textbf{Pros and cons of Actor-Critic DRL methods}]	
Actor-critic methods combine the advantages of both value-based and Monte Carlo policy gradient methods. They can be either on-policy or off-policy. Compared with Monte Carlo methods, they require far less samples to learn from and less computational resources to select an action, especially when the action space is continuous. Compared with value-based methods, they can learn stochastic policies and solve RL problems with continuous actions. However, it is prone to be unstable due to the recursive use of value estimates.
\end{remark3}

From the above discussion, we know that in Monte Carlo methods, the policy gradient is unbiased but with high variance; while in actor-critic methods, it is deterministic but biased. Therefore, an effective way is to combine these two types of methods together. Q-prop is such an efficient and stable algorithm proposed by S. Gu, T. Lillicrap \textit{et al.} in 2016 \cite{gu2016q}. It constructs a new estimator that provides a solution to high sample complexity and combines the advantages of on-policy and off-policy methods. \par

Q-prop can be directly combined with a number of prior policy gradient DRL methods, such as DDPG and TRPO. Compared with actor-critic methods such as DDPG algorithms, Q-prop has achieved higher stability in DRL tasks in real-world problems. One limitation with Q-prop is that the computation speed will be slowed down by the critic training when the speed of data collection is fast.\par

%\newtheorem{remark4}[remark]{Remark}
%\begin{remark4}[\textbf{Pros and cons of DPG DRL methods}]	
%DPG methods are a special type of actor-critic methods that focus on deterministic policies. Compared with their SPG counterparts, they require less samples in computing the deterministic policy gradient, especially if the action space has many dimensions. Different from value-based methods that can only solve RL problems with discrete actions, DPG-based methods focus on and work well for high dimensional continuous action problems. It usually works off-policy to guarantee enough exploration unless there is sufficient noise in the environment. Also, when combined with DQN, DPG-based method is good only when the Q function is accurate.
%\end{remark4}

\paragraph{Simple Policy Gradient vs. Natural Policy Gradient}
The policy gradient methods discussed above all use a simple gradient of loss function $\nabla_{\theta} L(\theta)$ to update the parameters of NN. On the other hand, NPG method updates the parameters in NN using the natural gradient $\nabla_{\theta}^{\mathrm{N}}L(\theta)$ as discussed in Section II.B instead of simple gradient to provide a more efficient solution \cite{kakade2002natural}.\par

%The loss function of NPG is the same as that of the SPG, i.e.,
%\begin{equation}
%{L}^{\mathrm{NPG}}(\theta) =  V^{\pi_{\theta}}(s).
%\end{equation}

The loss function of NPG is the same as that of SPG, whose general expression is given in \eqref{equ17}. The parameters are updated as
\begin{equation}
\theta \leftarrow \theta \!+\! F_{\theta}^{-1} \nabla_{\theta} V^{\pi_{\theta}}(s),
\end{equation}

\noindent where

\begin{equation}
F_{\theta}=\mathrm{E}_{\pi_{\theta}}\left[\nabla_{\theta} \log \pi\left(a_{t} | s_{t} ; \theta\right)\left(\nabla_{\theta} \log \pi\left(a_{t} | s_{t} ; \theta\right)\right)^{T}\right]
\end{equation}

\noindent is the Fisher information matrix used to measure the step size for update \cite{schulman2015trust}. \par

NPG method defines a new form of step size that specifies how much those parameters should be adjusted, and therefore provides a more stable and effective update. However, the drawback of NPG is that when complicated NN is used to approximate the policy where the number of parameters is large, it is impractical to calculate the Fisher information matrix or store them appropriately \cite{franccois2018introduction}. Methods originated from NPG, such as Trust Region Policy Optimization (TRPO) \cite{schulman2015trust} and Proximal Policy Optimization (PPO) \cite{schulman2017proximal} solve the above problem to some extent and are widely used for DRL in practice. Moreover, there are algorithms applying NPG to actor-critic methods, such as Actor Critic using Kronecker-Factored Trust Region (ACKTR) \cite{wu2017scalable} and Actor-Critic with Experience Replay (ACER) \cite{wang2016sample}.\par

%A set of mini-batch data is randomly sampled in the replay memory buffer and is regarded as the training set of $Q$ and $\pi$. Based on the Monte-Carlo approach, an un-biased estimate of ${\nabla_a}{Q^{{\pi_\theta }}}(s,a)$ is obtained \cite{wang2016sample}.

\subsection {Advanced DRL Algorithms}
\subsubsection{POMDP-based DRL}
In the previous sections, we consider RL in a Markovian environment, which implies that knowledge of the current state is always sufficient for optimal control. However in many real-world problems, total environment information cannot be observed by the agent accurately, usually due to the limitations in sensing and communications capabilities. An agent acting under situation with partial observability can model the environment as a POMDP \cite{shani2013survey}. RL tasks in realistic environments need to deal with those incomplete and noisy state information resulting from POMDP.\par

POMDP can be seen as an extension of MDP by adding a finite set of observations and a corresponding observation model \cite{dai2013pomdp}. A POMDP is usually defined as a six-tuple $<\mathcal{S}, \mathcal{A}, P, r, \Omega, \mathcal{O}>$, where state space $\mathcal{S}$, action space $\mathcal{A}$, transition probability $P$, and reward $r$ are defined previously as elements in MDP,

\begin{itemize}
	\item $\Omega$ is the observation space, where $o\in\Omega$ is a possible observation.
	\item $\mathcal{O}\left(o|s', a\right)$ is the conditional probability that taking an action $a$ leading to a new state $s'$ will result in an observation $o$.
\end{itemize}

Similar to MDP, an agent chooses an action $a\in\mathcal{A}$ according to policy $\pi(a|s)$ which results in the environment transiting to a new state $s'\in\mathcal{S}$ with probability $P\left(s'|s, a\right)$ and the agent receives a reward $r(s,a)$. Different from MDP, the agent cannot directly observe system states, but instead receives an observation $o\in\Omega$ which depends on the new state of the environment with probability $\mathcal{O}\left(o|s', a\right)$. Also, the policy and Q-function are modified as $\pi(a|o)$ and $Q(o,a)$ respectively.\par

Since the agent cannot directly observe the underlying state, it needs to exploit history information to reduce uncertainty about the current state \cite{wierstra2007solving}. The observation history at time step $t$ can be defined as $h_{t}=\{(o_{1},a_{1}),\cdots,(o_{t-1},a_{t-1}),(o_{t},-)\}$. \par

%When the environment model is known to the RL agent, the optimal approach is for the agent to compute a belief $b_{t}=P(x_{t}|h_{t})$ that provides a probability distribution over states. By introducing belief state, POMDP problem can be converted to a belief-based MDP problem. On the other hand, if the environment model is not available to the agent, it is straightforward to use the last $k$ observations as input to the policy. However, using a finite history may result in important information in the past being forgotten and overlooked. In order to overcome this challenge, RNN appears to be a good solution, because it is designed to deal with time series and can maintain long-term memory \cite{murphy2000survey}.\par

Several typical existing methods of solving POMDP problems are listed as follows.\par

\paragraph{Deep Recurrent Q-Network (DRQN)}
%An agent is able to chose actions in complex tasks by value-based methods for DRL, e.g. DQN and DDQN. However, it is hard for those methods to get outstanding performance when the agent cannot perceive the complete information of the state. 

To address the partial observable problem, Hausknecht \textit{et al.} proposed Deep Recurrent Q-Network (DRQN) in 2015 to integrate information through time and enhance DQN's performance \cite{hausknecht2015deep}. DRQN adds recurrency to DQN by replacing DQN's first fully-connected layer with a LSTM layer. 

In the partially observed cases, the agent does not have access to state $s_{t}$. So Q-function in terms of history $h_{t}$ is defined as $Q(h_{t}, a_{t})$, which is the output of NN \cite{heess2015memory}. The input to NN is $o_{t}$, while the rest of the information in $h_{t}$ apart from $o_{t}$, i.e., $\{(o_{1},a_{1}),\cdots,(o_{t-1},a_{t-1})\}$ is captured by the hidden states in RNN. \par

%Here, the Bellman optimality equations for Q function is
%
%\begin{equation}
%\begin{split}
%&Q^{*}(h_{t}, a_{t})=  \\ 
%&r_{t+1}+\gamma\sum_{h_{t+1}}P(h_{t+1}|h_{t},a_{t}) \max _{a_{t+1}}Q(h_{t+1},a_{t+1}).
%\end{split}
%\end{equation}
%
%Compared with DQN where tuples $(s_t , a_t , r_{t+1} , s_{t+1})$ are stored in memory and sampled for training, in DRQN, the tuples are modified as $(h_t , a_t , r_{t+1} , h_{t+1})$ and sampled for two types of updates. Those two types of updates are referred to as bootstrapped sequential updates and bootstrapped random updates, respectively. In both methods, episodes are selected randomly from the replay memory. In bootstrapped sequential updates, updates start at the beginning of the episode and sample experiences sequentially through the entire episode, while the RNN’s hidden state is carried forward throughout the episode. In bootstrapped random updates, updates start at random points in the episodes, while the RNN’s initial state is zeroed at the start of the update. The sequential updates method learns faster but violates the DQN's random sampling policy. Both methods show good performance in experiments.\par

\paragraph{Recurrent Policy Gradients (RPG)}
RPG methods belong to policy gradient methods where NNs are used to approximate policies  \cite{wierstra2007solving}. As mentioned in Section II-D, in policy gradient methods, $\pi(s)$ or $\pi(a|s)$ is a direct mapping from state $s$ to action $a$. But in RPG, the goal of the agent is to learn a policy that maps history $h$ to action $a$, which is denoted as $\pi(h)$ or $\pi(a|h)$.\par

%Similar to \eqref{equ13} and \eqref{equ18}, the parameters are updated as 
%
%\begin{equation}
%\theta \leftarrow \theta + \alpha G(\tau_{h_{t}}) \nabla_{\theta} \log\pi(a_{t}|h_{t};\theta),
%\end{equation}
%
%\noindent for stochastic policies, and
%
%\begin{equation}
%\theta \leftarrow \theta \!+\! \alpha \left[\nabla_{\theta} \pi_{\theta}(h) \nabla_{a} Q^{\pi_{\theta}} \!\!\! \left.(h, a)\right|_{a=\pi_{\theta}(h)} \!\right]
%\end{equation}
%
%\noindent for deterministic policies, respectively, where $\tau_{h_t}$ refers to the sampling trajectory of history $h_{t}$. Here, RNN is trained to obtain information from $h$ by its recurrent state and compute $\pi(h)$ as well as $Q(h,a)$.

RPG methods are applied to many partially observed physical control problems i.e. system identification with variable and unknown information, short-term integration of sensor information to estimate the system state, as well as long-term memory problems. A typical algorithm, Recurrent Deterministic Policy Gradient (RDPG), is proposed by N. Heess, J. J. Hunt et.al based on RPG methods \cite{heess2015memory}.

\paragraph{Memory, RL, and Inference Network (MERLIN)}
%In RL algorithms, extensive memory can be used to solve POMDP tasks. 

MERLIN algorithm focuses on memory-dependent policies which output the action distribution based on the entire observation sequence in the past \cite{wayne2018unsupervised}. The ideas for MERLIN, including predictive sensory coding, hippocampal representation theory and temporal context model, mainly originate in neuroscience and psychology. It is mainly composed of two basic components: a memory-based predictor and a policy. \par

%The memory-based predictor is mainly used to compress the input observation $o$ into low-dimensional state variables $z$ to represent a state. In each time step, the recurrent network outputs a prior distribution $p\left(z_{t} | z_{1}, a_{1}, \dots, z_{t-1}, a_{t-1}\right)$ to predict the next state variable. Next, a posterior distribution $q\left(z_{t} | z_{1}, a_{1}, \dots, z_{t-1}, a_{t-1}, o_{t}\right)$ is obtained based on the observation $o_t$ and the prior distribution $p$. The posterior distribution $q$ has corrected the prior distribution $p$ to form a more accurate estimate of state variable. Then, $z_{t}$ is sampled from distribution $q$. $z_{t}$ is used to select action by the other basic component and stored in the memory next step prediction.\par

\paragraph{Deep Belief Q-Network (DBQN)}
%As an agent cannot directly observe the current state of the environment, a belief $b$ can be defined to provide a probability distribution over states. By introducing belief state, POMDP problem can be converted to a belief-based MDP problem. 

%Based on a belief state $b$, which is a probability distribution describing the probability of being at every state, a belief - based MDP can be created.

DBQN is a model-based method that uses DQN to map a belief $b_t$ to an action. When $P$, $o$ and $r$ in a POMDP model are known, $b_t$ can be estimated accurately with Bayes' theorem and sent to NN as input
\cite{egorov2015deep}. During updating, this approach usually leads to divergence. To stabilize the learning, techniques like experience replay, target network and an adaptive learning method are used.\par

%The Bellman optimality equation for beliefs is
%
%\begin{equation}
%\begin{split}
%&Q^{*}(b_{t}, a_{t})=  \\ 
%&r_{t+1}+\gamma\sum_{b_{t+1}}P(b_{t+1}|b_{t},a_{t}) \max _{a_{t+1}}Q(b_{t+1},a_{t+1}).
%\end{split}
%\end{equation}

% For experience replay, tuples $\left(b_{t},a_{t},r_{t+1},b_{t+1}\right)$ are stored in memory and sampled uniformly. The adaptive learning method is used to regulate the parameter adjustment rate of the network \cite{egorov2015deep}.\par

Besides, there are also other methods of solving POMDP problems, some of which are developed based on RNN and typical methods for DRL, such as Action-specific Deep Recurrent Q-Network (ADRQN) \cite{zhu2018improving}, and Deep Distributed Recurrent Q-Networks (DDRQN) \cite{foerster2016learning}.\par

\subsubsection {Multi-Agent DRL}

In the previous sections, we mainly discuss the DRL methods for single-agent cases. In practice, there are situations where multiple agents need to work together, e.g. the manipulation in multi-robot systems, the cooperative driving of multiple vehicles. In these cases, DRL methods for MA systems are designed. \par

An MA system consists of a group of autonomous, interacting agents sharing a common environment, and has a good degree of robustness and scalability \cite{bu2008comprehensive}. The multiple agents in the system can interact with each other in cooperative or competitive settings, and hence the concept of stochastic game is introduced to extend MDP into the MA setting. A stochastic game or MA-MDP with $N$ agents is defined as a tuple $<\mathcal{S},\mathcal{A}_{1},\cdots,\mathcal{A}_{N},P,r_{1},\cdots,r_{N}>$, where 
\begin{itemize}
	\item $\mathcal{S}$ is the discrete set of states,
	\item $\mathcal{A}_{i}$, $i=1,\cdots,N$ are the discrete sets of actions available to the agents, yielding the joint action set $\mathcal{A}=\mathcal{A}_{1}\times\cdots\times\mathcal{A}_{N}$,
	\item $P:\mathcal{S}\times\mathcal{A}\times\mathcal{S}\rightarrow[0,1]$ is the state transition probability function,
	\item $r_{i}:\mathcal{S}\times\mathcal{A}\rightarrow\mathbb{R}$, $i=1,\cdots,N$ are the reward functions for the agents. 
\end{itemize}

In MA-MDP, the state transitions depend on the joint action $a=[a_{1},\cdots,a_{n}]$ of all the agents, where $a\in\mathcal{A}$ and $a_{i}\in\mathcal{A}_{i}$. In the fully-collaborative problems, all the agents share the same reward, i.e., $r_{1}=\cdots=r_{N}$. In the fully-competitive problems, the agents have opposite rewards with $r_{1}+\cdots+r_{N}=0$. Therefore, $r_{1}=-r_{2}$ in the typical scenario with two agents \cite{bu2008comprehensive}. MA-MDP problems that are neither fully collaborative nor fully competitive are mixed games.\par 

%In a competitive formulation, each agent considers the reward that is integrated not only with respect to its own policy but with respect to the policies of all agents. In the collaborative problems, each agent takes the rewards of other agents into consideration, and the reward of interest to each agent is the expected reward accumulated over time and across all agents. Similar to \eqref{equ1} in the single-agent cases, the value function for a multi-agent system is presented as follows,
%\begin{equation}
%V^{\Pi}\left(s_{0}\right)=\left\{\begin{array}{c}{\mathrm{E}_{\tau_{s_{0}} \sim \Pi}\left[G_{n}\left(\tau_{s_{0}}\right)\right]},\rm{competitive}, \\ {\mathrm{E}_{\tau_{s_{0}}\sim\Pi}\left[\sum\limits_{n = 1}^N G_{n}\left(\tau_{s_{0}}\right)\right]},\rm{cooperative},\end{array}\right.
%\end{equation}
%where $G_n$ is the total reward of agent $n (n=1,2,…,N)$, and $\Pi  = \{ {\pi _1},{\pi _2},...,{\pi _N}\}$
%is the overall policy of all $N$ agents \cite{khan2018scalable}. 

In MA RL, each agent learns to improve its own policy by interacting with the environment to obtain rewards. For each agent, the environment is usually complex and dynamic, and the system may encounter the action space explosion problem. Since multiple agents are learning at the same time, for a particular agent, when the policies of other agents change, the optimal policy of itself may also change. This may affect the convergence of the learning algorithm and cause instability. \par

%The simplest approach to learning in multi-agent settings is to use independently learning agents. For example, independent-Q learning is an algorithm in which each agent independently learns its own policy, treating other agents as part of the environment \cite{Tan1997}. However, independent-Q learning cannot deal with the non-stationary environment problem. Combining game theory with RL, some typical algorithms for Multi-Agent RL (MARL) have been studied, aiming to solve the problems mentioned above. Some examples are the Minimax-Q algorithm \cite{littman1994markov}, the Nash Q-Learning algorithm \cite{hu2003nash}, the Friend-or-foe Q-Learning (FFQ) algorithm \cite{littman2001friend}, and  WoLF-PHC \cite{bowling2001rational}. \par

In recent years, the DRL methods for single-agent cases have been extended to the MAs cases as discussed below.\par

\paragraph{Multi-Agent Value-Based Methods}
The experience replay mechanism in DQN algorithm is not designed for the non-stationary environment in MA systems. Several variants of DQN have been proposed to deal with this problem.\par  
%{\color{blue}Foerster et al. \cite{Foerster2017} introduced two methods for stabilising experience replay of DQN in MADRL. The firrst method uses the importance sampling approach to naturally decay obsolete data whilst the second method disambiguates the age of the samples retrieved from the replay memory using a fingerprint. comment: Above is directly copied. Add other algorithms.}
Foerster \textit{et al.} \cite{Foerster2017} introduced two methods for stabilizing experience replay of DQN in MA DRL. In the MA importance sampling (MAIS) algorithm, off-environment importance sampling is introduced to stabilize experience replay, where obsolete data is supposed to decay naturally. In the MA fingerprints (MAF) algorithm, each agent needs to be able to condition on only those values that actually occur in its replay memory to stabilize experience replay.\par
In \cite {van2016coordinated}, a coordinated MA DRL method is designed based on DQN. Faster and more scalable learning is realized by using transfer planning. To coordinate between multiple agents, the global Q-function is factorized as a linear combination of local sub-problems. Then, the max-plus coordination algorithm is applied to optimize the joint global action over the entire coordination graph.\par
%In multi-agent shaped Q-learning (MASQL), each agent maintains a cooperative tendency table (CTT), and observe partners' actions when taking an action. When an agent perceives a state, the corresponding cooperative tendency value, and the Q-value are merged to a Shaped-Q value. The action with maximal Shaped-Q value is selected, then the agents update their own CTTs. As a result, a consensus is achieved more quickly, and the learning efficiency is improved.

\paragraph{Multi-Agent Policy Gradient Methods}
Policy gradient methods usually exhibit very high variance when coordination of multiple agents is required. In order to overcome this challenge, several algorithms adopt the framework of centralized training with decentralized execution.\par

%{\color{blue} Comment: Add other algorithms, e.g., Counterfactual multi-agent (COMA) policy gradients ...} \par
In the counterfactual MA policy gradient (COMAPG) algorithm \cite{foerster2018counterfactual}, a centralized critic is used to estimate the Q-function, and decentralized actors are used to optimize the policies of multiple agents. The core idea of the COMAPG algorithm is to apply a counterfactual baseline, which can marginalize out a single agent’s action and keep the other agents’ actions fixed. Moreover, a critic representation is introduced for efficiently evaluating the counterfactual baseline in a single forward pass. \par
MA deep deterministic policy gradient (MADDPG) \cite{lowe2017multi} is essentially a DPG algorithm that trains each agent with a critic that requires global information and an actor that requires local information. It allows each agent to have its own reward function, so that it can be used for cooperative or competitive tasks. \par

%The core idea of the MADDPG algorithm is the centralized training and the distributed execution. The training process uses centralized learning to train critic and actor. When executing, the actor only needs to know the local information. Critic requires policy information from other agents. \par

Based on the above tutorial, we list the classical algorithms for each type of DRL methods in Table \ref{table_al} and summarize their pros and cons. \par

\begin{table*}[]

	\newcommand{\tabincell}[2]{\begin{tabular}{@{}#1@{}}#2\end{tabular}}

	\centering

	\renewcommand{\arraystretch}{1.2}

	\caption{Classical algorithms for DRL.}

	\label{table_al}

	\begin{tabular}{|p{1cm}|p{1cm}|p{1.5cm}|p{1.8cm}|p{6cm}|p{4cm}|}

		\hline

		\textbf{Feature}          & \multicolumn{3}{c|}{\textbf{Type}}                                                                                                                                                                   & \textbf{Classical Algorithms}                                                                                                                                                                                                                                                                                                                                                                                                                                                  & \textbf{Pros \& Cons}                                                                                                                           \\ \hline

		\multirow{8}{*}{basic}    & \multicolumn{3}{c|}{Value-Based}                                                                                                                                                                     & Deep Q-network (DQN) \cite{mnih2015human}, Double Deep Q-network (DDQN) \cite{van2016deep}, DDQN with duel architecture \cite{wang2015dueling}, DDQN with Proportional Prioritization \cite{schaul2015prioritized}                                                                                                                                                                                                                                                             & simplicity and good performance; only suitable for discrete action space                                                                         \\ \cline{2-6}

		& \multirow{7}{*}{\begin{tabular}[c]{@{}c@{}}Policy \\ Gradients\end{tabular}} & \multirow{2}{*}{\begin{tabular}[c]{@{}c@{}}classified \\ according to \\ natures of\\ policy \\ functions\end{tabular}}             & Stochastic Policy Gradient (SPG)     & REINFORCE \cite{williams1992simple}, Soft Actor-Critic (SAC) \cite{haarnoja2018soft}, Asynchronous Advantage Actor Critic (A3C) \cite{mnih2016asynchronous}                                                                                                                                                                                                                                                                                                                    & /                                                                                                                                             \\ \cline{4-6}

		&                                  &                                                                                                                             & Deterministic Policy Gradient (DPG) & Deep Deterministic Policy Gradient (DDPG) \cite{lillicrap2015continuous}, Distributed distributional deep deterministic policy gradients (D4PG), Twin Delayed Deep Deterministic (TD3) \cite{fujimoto2018addressing}                                                                                                                                                                                                                                                           & requiring less samples; only suitable for continuous action space                                                                                \\ \cline{3-6}

		&                                  & \multirow{3}{*}{\begin{tabular}[c]{@{}c@{}}classified \\according to \\ ways of \\policy\\ evaluation\end{tabular}}               & Monte Carlo                         & REINFORCE \cite{williams1992simple}, Trust Region Policy Optimization (TRPO) \cite{schulman2015trust}, Proximal Policy Optimization (PPO) \cite{schulman2017proximal}, Trust Region Policy Optimization (TRPO) \cite{schulman2015trust}, Proximal Policy Optimization (PPO) \cite{schulman2017proximal}                                                                                                                                                                        & better convergence properties, higher efficiency in high-dimensional or continuous action spaces; high variance of estimations, sample intensive \\ \cline{4-6}

		&                                  &                                                                                                                             & Actor-Critic                        & Soft Actor-Critic (SAC) \cite{haarnoja2018soft}, Asynchronous Advantage Actor Critic (A3C) \cite{mnih2016asynchronous}, Deep Deterministic Policy Gradient (DDPG) \cite{lillicrap2015continuous}, Distributed distributional deep deterministic policy gradients (D4PG), Twin Delayed Deep Deterministic (TD3) \cite{fujimoto2018addressing}, Trust Region Policy Optimization (TRPO) \cite{schulman2015trust}, Proximal Policy Optimization (PPO) \cite{schulman2017proximal} & requiring less samples and less computational resources; unstable due to the recursive use of value estimates                                    \\ \cline{4-6}

		&                                  &                                                                                                                             & Monte Carlo \& Actor-Critic        & Q-Prop \cite{gu2016q}                                                                                                                                                                                                                                                                                                                                                                                                                                                          & higher stability; slow computation speed when the speed of data collection is fast                                                               \\ \cline{3-6}

		&                                  & \multirow{2}{*}{\begin{tabular}[c]{@{}c@{}}classified \\ according to \\ learning or \\ parameter\\ update\\ techniques\end{tabular}} & Simple Policy Gradient              & REINFORCE \cite{williams1992simple}, Soft Actor-Critic (SAC) \cite{haarnoja2018soft}, Asynchronous Advantage Actor Critic (A3C) \cite{mnih2016asynchronous}, Deep Deterministic Policy Gradient (DDPG) \cite{lillicrap2015continuous}, Distributed distributional deep deterministic policy gradients (D4PG), Twin Delayed Deep Deterministic (TD3) \cite{fujimoto2018addressing}                                                                                              & /                                                                                                                                             \\ \cline{4-6}

		&                                  &                                                                                                                             & Natural Policy Gradient (NPG)       & Trust Region Policy Optimization (TRPO) \cite{schulman2015trust}, Proximal Policy Optimization (PPO) \cite{schulman2017proximal}, Actor Critic using Kronecker-Factored Trust Region (ACKTR) \cite{wu2017scalable}, Actor-Critic with Experience Replay (ACER) \cite{wang2016sample}                                                                                                                                                                                           & more stable and effective update; impractical to calculate the Fisher information matrix when complicated NN is used                             \\ \hline

		\multirow{2}{*}{advanced} & \multicolumn{3}{c|}{POMDP}                                                                                                                                                                           & Deep Belief Q-network (DBQN) \cite{zhu2018improving}, Deep Recurrent Q-network (DRQN) \cite{hausknecht2015deep}, Recurrent Deterministic Policy Gradients (RDPG) \cite{heess2015memory}                                                                                                                                                                                                                                                                                        & \multirow{2}{*}{/}                                                                                                                            \\ \cline{2-5}

		& \multicolumn{3}{c|}{MA}                                                                                                                                                                              & Multi-agent Importance Sampling (MAIS) \cite{Foerster2017}, Coordinated Multi-agent DQN \cite{van2016coordinated}, Multi-agent Fingerprints (MAF) \cite{Foerster2017}, Counterfactual Multi-agent Policy Gradient (COMAPG) \cite{foerster2018counterfactual}, Multi-agent DDPG (MADDPG) \cite{lowe2017multi}                                                                                                                                                                   &                                                                                                                                                  \\ \hline

	\end{tabular}

\end{table*}

\section{General RL/DRL Model for Autonomous IoT}
\begin{figure}[!b]
	\centering
	\includegraphics[width= 0.4\textwidth]{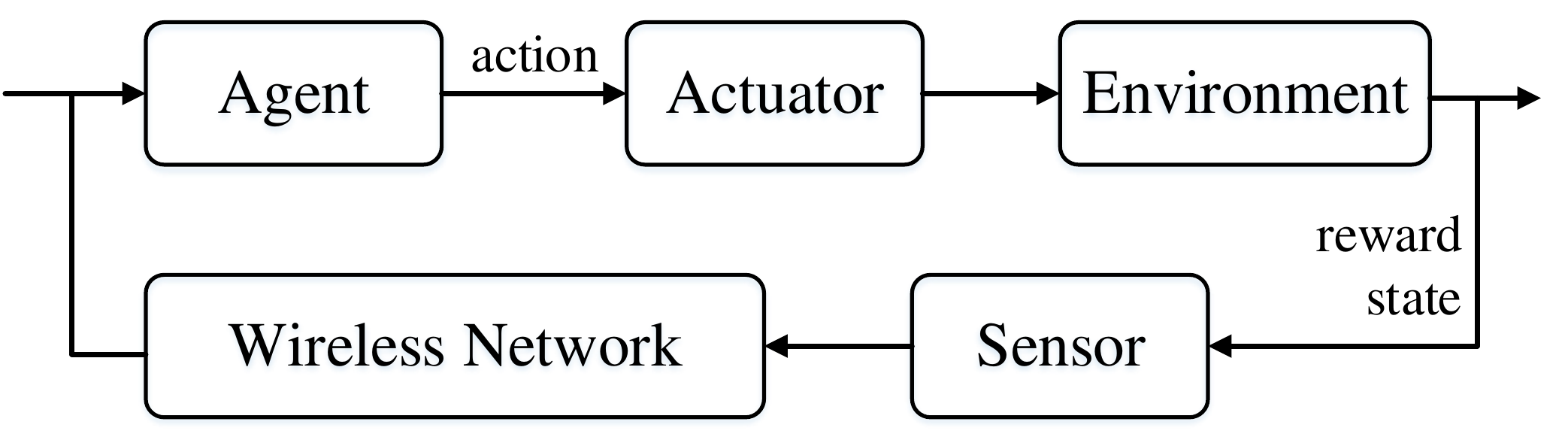}
	\caption{The RL/DRL model for WSAN.}
	\label{AIoT_RL}
\end{figure}
Before we discuss the RL/DRL model for the AIoT system, we first examine that of a wireless sensor and actuator network (WSAN), which can be considered as an element or a simplified version of AIoT. A WSAN consists of a group of sensors that gather information about their environment, and a group of actuators that interact with and act on the environment. All elements communicate wirelessly. In the RL/DRL model for WSAN as illustrated in Fig. \ref{AIoT_RL}, an agent obtains aspects of its environment through sensors, and chooses control actions that are implemented by the actuators. The chosen action determines the value of the immediate reward as well as influences the dynamics of its environment. The agent communicates with the sensors and actuators to receive state information and send control commands.   \par

\begin{figure}[!t]
	\centering
	\includegraphics[width= 0.5\textwidth]{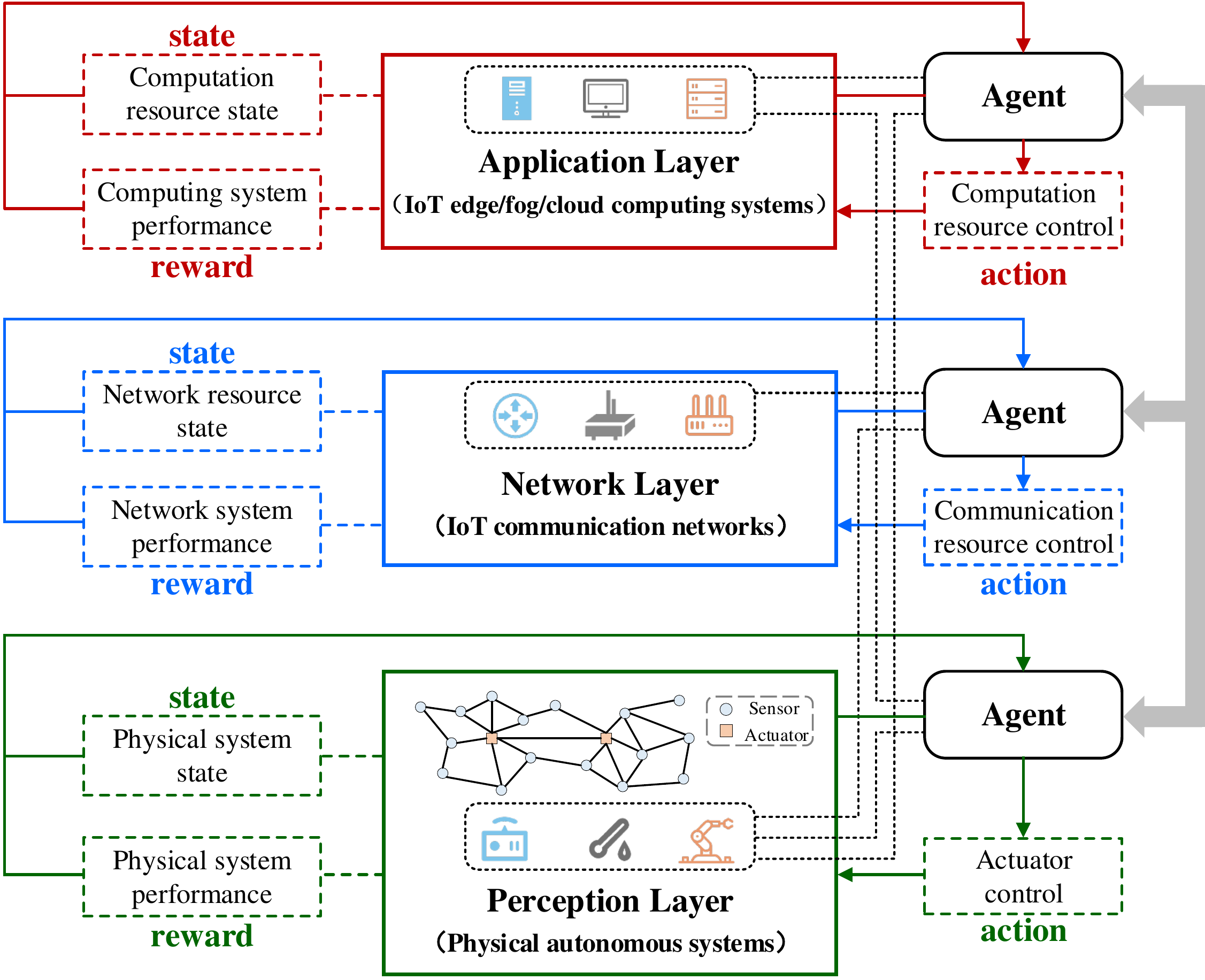}
	\caption{General RL/DRL model for autonomous IoT.}
	\label{aiot_rl}    
\end{figure}

Compared with WSAN, the AIoT has a more complex ecosystem that encompasses identification, sensing, communication, computation, and services. A typical AIoT architecture consists of three fundamental building blocks as shown in Fig. \ref{aiot_rl}:
\begin{itemize}
	\item \textbf{Perception layer}: corresponds to the \textbf{physical autonomous systems} in which IoT devices with sensors and actuators interact with the environment to acquire data and exert control actions;
	\item \textbf{Network layer}: corresponds to the \textbf{IoT communication networks} including wireless access networks and the Internet that discover and connect the IoT devices to the edge/fog servers and cloud servers for data and control command transmission;
	\item \textbf{Application layer}: corresponds to the \textbf{IoT edge/fog/cloud computing systems} for data processing/storage and control actions determination.
\end{itemize} 

Due to the more sophisticated system architecture, the RL/DRL models for AIoT systems are more complex than those of WSAN as illustrated in Fig. \ref{AIoT_RL}. The environment can include one or more layers in the AIoT architecture. The agent(s) can locate at the IoT devices, the edge/fog/cloud servers, and wireless APs. In the following, we first define the basic RL/DRL elements such as state, action, and reward for each layer, respectively. Then, we define the RL/DRL elements when the environment includes all the three layers as an integrated part.\par

\subsection{Perception Layer}
When the environment only includes the perception layer, the physical system dynamics are modeled by a controlled stochastic process with the following state, action, and reward.
\begin{itemize}
	\item \textbf{Physical system state} $(s_{\mathrm{phy}})$, e.g., the on-off status of the actuators, the RGB images of the system, the locations of the agents; 
	\item \textbf{Actuator control action} $(a_{\mathrm{actu}})$, e.g., controlling the movement of a robot, adjusting the driving speed and direction of a vehicle, turning on/off a device;
	\item \textbf{Physical system performance} $(r_{\mathrm{phy}})$, e.g., energy consumption in a power grid, how fast a mobile agent such as a robot or a vehicle can move, or whether it is away from obstacles.
\end{itemize}

\subsection{Network Layer}
When the environment only includes the network layer, the network dynamics are modeled by a controlled stochastic process with the following state, action, and reward.
\begin{itemize}
	\item \textbf{Communication network state} $(s_{\mathrm{net}})$, e.g., the amount of allocated bandwidth, the signal to interference plus noise ratio, the channel vector of a finite state Markov channel model;
	\item \textbf{Communication resource control action} $(a_{\mathrm{comm\_re}})$, e.g., the power allocation, the multi-user scheduling, the subchannel allocation in OFDM system;    
	\item \textbf{Communication network performance} $(r_{\mathrm{net}})$, e.g., the transmission delay, the transmission error probability, the transmission power consumption. 
\end{itemize}

\subsection{Application Layer}
When the environment only includes the application layer, the edge/fog/cloud computing system dynamics are modeled by a controlled stochastic process with the following state, action, and reward.
\begin{itemize}
	\item \textbf{Computing system state}  $(s_{\mathrm{comp}})$, e.g., the number of virtual machines (VMs) that currently run, the number of tasks buffered in the queue for processing;
	\item \textbf{Computing resource control action} $(a_{\mathrm{comp\_re}})$, e.g., the caching selection, the task offloading decisions, the virtual machine allocation;    
	\item \textbf{Computing system performance} $(r_{\mathrm{comp}})$, e.g., utilization rate of the computing resources, the processing delay of the offloading tasks.
\end{itemize}

\subsection{Integration of Three Layers}
When the environment includes all the three layers of AIoT architecture, the RL/DRL models generally include elements defined as follows.
\begin{itemize}
	\item \textbf{AIoT state} ($s_{\mathrm{aIoT}}$) includes the aggregation of physical system state, network resource state, and computation resource state, i.e., $s_{\mathrm{aIoT}}=\{s_{\mathrm{phy}},s_{\mathrm{net}},s_{\mathrm{comp}}\}$;
	\item \textbf{AIoT action} ($a_{\mathrm{aIoT}}$) includes the aggregation of actuator control action, communication resource control action, and computing resource control action, i.e., $a_{\mathrm{aIoT}}=\{a_{\mathrm{actu}},a_{\mathrm{comm\_re}},a_{\mathrm{comp\_re}}\}$;
	\item \textbf{AIoT reward} ($r_{\mathrm{aIoT}}$) is normally set to optimize the physical system performance, which can be expressed as a function of the network performance and computing system performance, i.e., $r_{\mathrm{aIoT}}=r_{\mathrm{phy}}(r_{\mathrm{net}},r_{\mathrm{comp}})$.   
\end{itemize}

As the agent in RL/DRL is a logical concept, the RL/DRL problem in each layer can be solved by the agent in its respective layer - observing the states and rewards from its environment and learning polices to determine corresponding actions as shown in Fig. \ref{aiot_rl}. However, the physical location of an agent can be different from its logical layer. We classify the devices that an agent may locate in according to the physical locations of the devices as 
\begin{itemize}
	\item perception layer devices, i.e., IoT devices;
	\item network layer devices, i.e, wireless APs;
	\item application layer devices, i.e., edge/fog/cloud servers.
\end{itemize} 

As shown in Fig. \ref{aiot_rl}, the mapping of the logical layer of an agent and its physical locations are given. A perception layer agent may locate in IoT devices and/or edge/fog/cloud servers. A network layer agent may locate in wireless APs and/or IoT devices (e.g., for Device-to-Device (D2D) communications). An application layer agent may locate in edge/fog/cloud servers and/or even IoT devices (e.g., to perform task offloading). \par

When the environment of an RL/DRL problem includes more than one layer, the agents of different layers need to share information and jointly optimize their polices. For example, the network layer may provide transmission delay information to the perception layer to be included as part of the system state; or, the perception layer may provide its optimization objective to the network layer to formulate the reward function. When the physical locations of the agents of different layers are the same, e.g., when both perception layer agent and application layer agent locate at the cloud servers, a single logical agent combining agents of different layers can be considered for the RL/DRL problem. \par

\begin{figure*}[!t]
	\centering
	\includegraphics[width=0.9\textwidth]{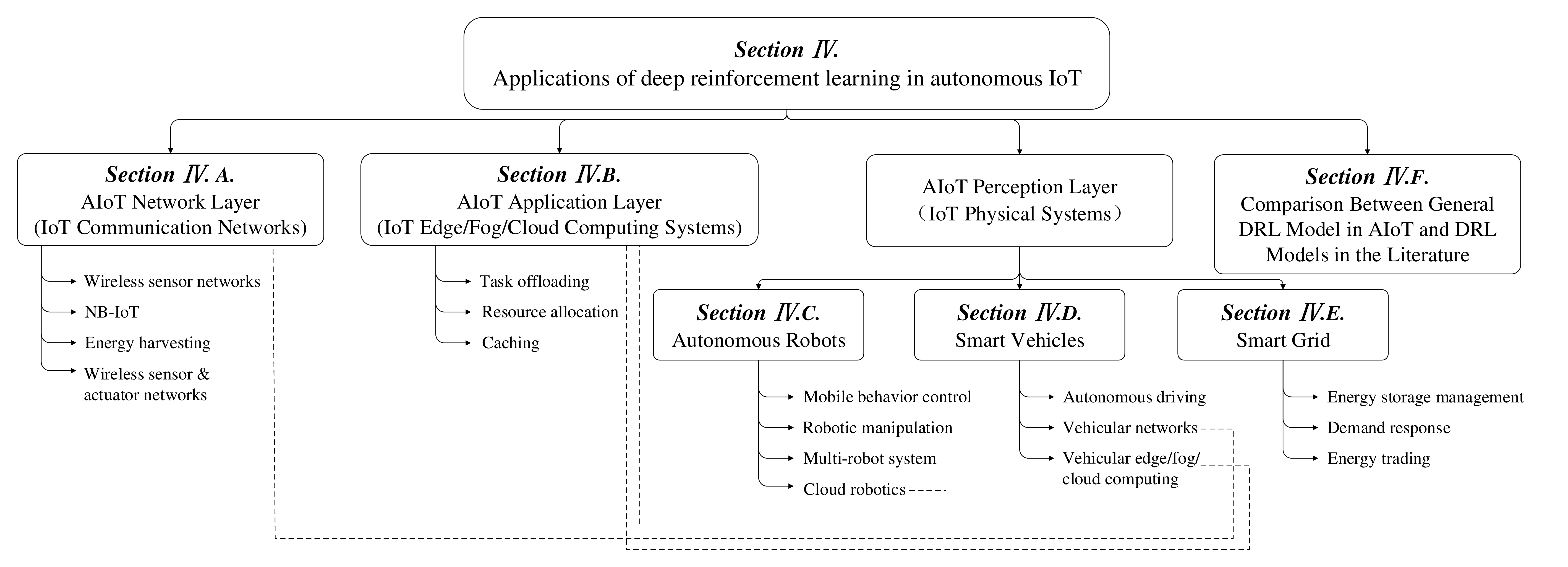}
	\caption{Applications of deep reinforcement learning in autonomous IoT.}
	\label{app}
\end{figure*}

\begin{table*}[t]
	
	\centering
	
	\renewcommand{\arraystretch}{1.2}
	
	\caption{Summary of states in RL/DRL Model for AIoT Systems.}
	
	\label{table_state}
	
	%    \resizebox{\textwidth}{!}{
	
	\begin{tabular}{|p{1.5cm}|p{1.6cm}|p{3.2cm}|p{0.5\textwidth}|} 
		\hline
		\multicolumn{2}{|c|}{\textbf{AIoT Layer}} & \textbf{Examples} & \textbf{Details} \\
		\hline
		& Autonomous & kinematic state
		& position, heading direction of the robot(s)    
		\\     \cline{3-4}    
		& Robots &manipulation state&
		angle of end-effector, opening status of the gripper, day-time or night-time mode, object grasping state 
		\\     \cline{3-4}        
		&  &surrounding environment&
		camera image and/or laser measurements of environment
		\\     \cline{2-4}    
		& Smart Vehicles & driving environment & 
		camera image of environment, relative position or distance to other vehicles, traffic signal state
		\\ \cline{3-4}    
		Physical system state&  &kinematic state&
		velocity and/or position of the agent vehicle, distance between multiple agent vehicles, state of the reservoir of UAV
		\\             \cline{2-4}        
		& Smart Grid &  battery SoC &
		amount of energy stored in the ESS
		\\ \cline{3-4}    
		& & time state &
		information on the time period relevant for the dynamics of the system, e.g. quarter of the day, day of the week and season of the year
		\\ \cline{3-4}
		& &RE generation state & 
		amount of renewable energy produced by PV panels, wind turbines, etc
		\\ \cline{3-4}    
		& &energy demand state& 
		energy demand in smart grid by critical load
		\\ \cline{3-4}
		& &price state & 
		real-time electricity price
		\\ \cline{3-4}        
		& &DR device on/off state& 
		on/off state of the DR device
		\\             \hline                
		\multicolumn{2}{|c|}{} & channel state & 
		SINR, pathloss, channel gain, data transmission rate, transmission successful indicator of wireless channel
		\\ \cline{3-4}    
		\multicolumn{2}{|c|}{} &topology state &
		number of nodes in the network, locations of moving sensors in the field, whether an area is covered by a sensor
		\\             \cline{3-4}
		\multicolumn{2}{|c|}{Communication network state} & sensor state& 
		sleep, active, idle, process, TxRx, state estimation error
		\\             \cline{2-4}
		& \multirow{3}{*}{} &queue state & 
		number of tasks/packets/bits waiting to be transmitted or processed
		\\             \cline{3-4}
		& \multirow{3}{*}{} &energy queue state & 
		amount of available energy for tasks/packets/bits transmission or processing
		\\             \cline{3-4} 
		\multicolumn{1}{|c|}{} && energy consumption state& 
		the amount of energy consumed by the system
		\\ \cline{1-1}  \cline{3-4}  
		&& task state & 
		remaining time to finish, waiting time, data size, CPU cycles, deadline, completion reward
		\\ \cline{2-4}  
		\multicolumn{2}{|c|}{Computing system state} & edge/fog/cloud server state&
		remaining computation resources; prices and CPU frequencies of different virtual machines (VMs) levels; number of VMs run in physical machines (PMs); whether a content is stored             
		\\ 
		\cline{3-4}    
		\multicolumn{2}{|c|}{} &content state & 
		popularity of requested content; number of requests for the content
		\\
		\hline                    
		
	\end{tabular}
	
	%    }
	
\end{table*}

\section{Applications of Deep Reinforcement Learning in Autonomous IoT}

Although AIoT is a new trend in IoT that has not been adequately studied by existing research works, the respective applications of DRL in each of the three layers of AIoT architecture have been widely studied by recent works. Therefore, we provide a literature review of the applications of DRL in the perception layer (physical autonomous systems), the network layer (IoT communication networks), and the application layer (IoT edge/fog/cloud computing systems) in this section. Most literature discussed in this section was published in the past decade, and was collected by searching in the IEEE Xplore using the words of related application and "deep reinforcement learning" as keywords. Moreover, we also found some other works in Google Scholar as a supplement. As there are a great variety of physical autonomous systems, we focus on three types of systems that have received most attention in DRL research for the perception layer, i.e., autonomous robots, smart vehicles, and smart grid.\par

In this section, we first discuss the applications of DRL in IoT communication networks and IoT edge/fog/cloud computing systems for general autonomous physical systems in Section IV.A and Section IV.B, respectively. Then, we focus on the three types of physical autonomous systems, i.e., autonomous robots, smart vehicles, and smart grid, in Section IV.C, Section IV.D, and Section IV.E, respectively. Note that some IoT communication technologies and IoT edge/fog/cloud computing technologies are designed specifically for a particular physical autonomous system, e.g, vehicular edge/fog/cloud computing and vehicular networks for smart vehicles, and cloud robotics for autonomous robots. These technologies are discussed in the respective physical autonomous system subsections. Finally, in Section IV.F, we compare the general DRL model in AIoT proposed in Section III with the reviewed DRL models in the existing literature. The framework of the literature review is given in Fig. \ref{app}.
	
\begin{figure}[!t]
		
		\centering
		
		\includegraphics[width= 0.5\textwidth]{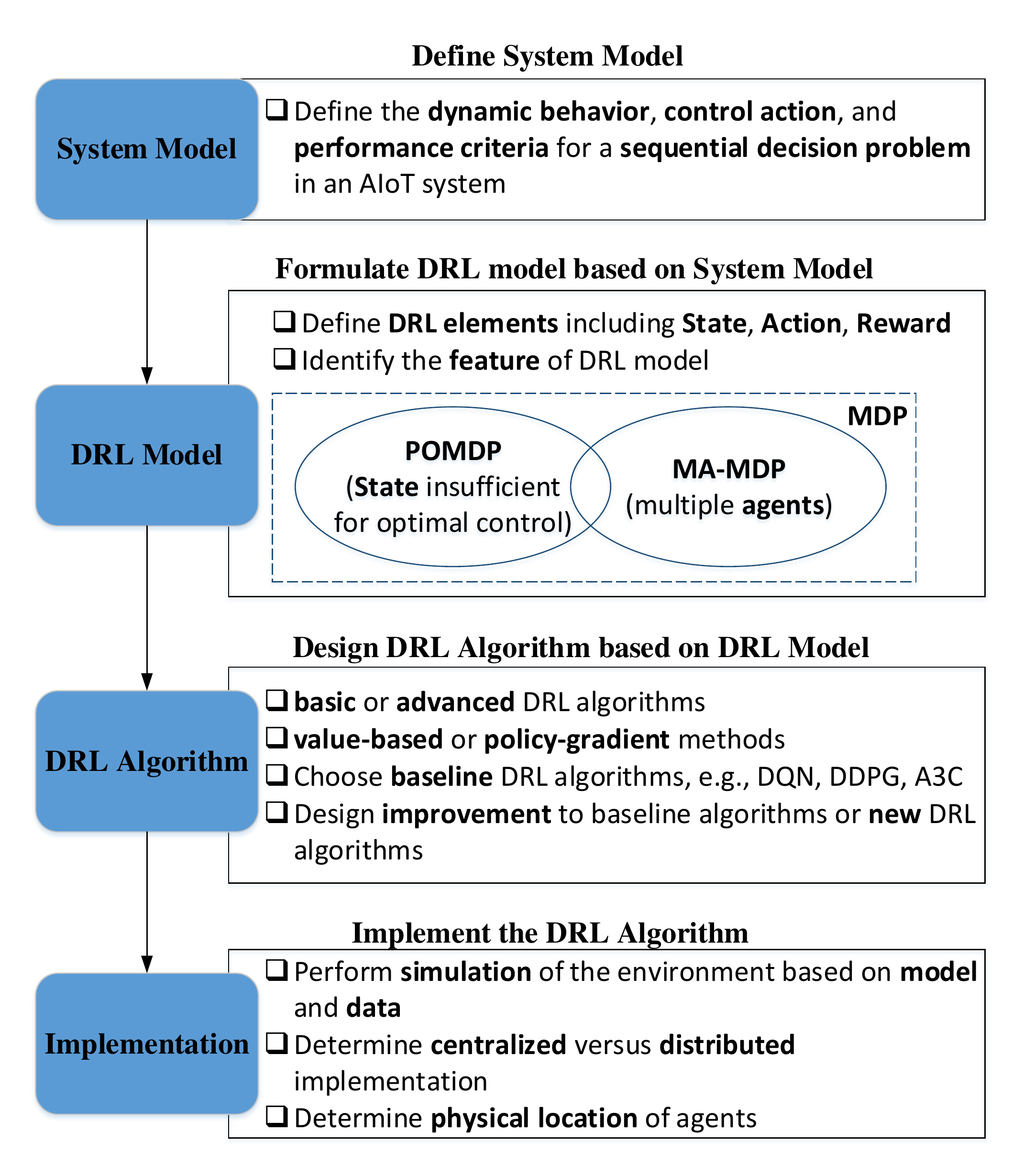}
		
		\caption{General procedure for solving optimal control problems in AIoT by DRL theory.}
		
		\label{drl_general_proc}             
		
	\end{figure}

	Before we review the existing literature, we first provide the general procedure for solving optimal control problems in AIoT by DRL theory as shown in Fig. \ref{drl_general_proc}. Firstly, a system model needs to be given, which defines the dynamic behavior, control action, and performance criteria for a sequential decision problem in an AIoT system. Secondly, a DRL model is formulated based on the system model. The most essential part is to define the DRL elements including state, action, and reward functions. The transition probability of the states can also be given whenever available. Moreover, it is important to identify the features of the DRL model. Generally speaking, the DRL model can be either a basic MDP model, a POMDP model or an MA-MDP model as introduced in Section II. Finally, DRL algorithms to solve the DRL model need to be developed and implemented. \par
	
	%\begin{figure}[!t]
	
	%            \centering
	
	%            \includegraphics[width= 0.5\textwidth]{DRL_general_procedure.pdf}
	
	%            \caption{General procedure for solving optimal control problems in AIoT by DRL theory.}
	
	%            \label{drl_general_proc}             
	
	%\end{figure}
	
	Some typical examples of states, actions, and rewards in DRL models for AIoT systems in existing literature are summarized and classified according to the AIoT layers in Table \ref{table_state}, Table \ref{table_action}, and Table \ref{table_reward}, respectively. In the rest of this section, we will review, compare and summarize the related research works from the following perspectives: \par
	\begin{itemize}
		
		\item review and compare \textbf{system models} considered for AIoT and identify promising system models for future research;
		
		\item  review and compare \textbf{DRL models} according to the DRL elements and features. Some guidelines for the formulation of DRL models are provided with respect to different system models;
		
		\item review and compare \textbf{DRL algorithms} used for AIoT with special attention on how to select different methods, e.g., value-based and policy gradient, according to the different characteristics of DRL models. As the baseline DRL algorithms such as DQN and DDPG are already introduced in Section II, we focus more on the new DRL algorithms that are different from the baseline algorithms or with proposed improvement over baseline algorithms;
		
		\item discuss the \textbf{implementation} considerations for the DRL algorithms, especially the physical location of the agent(s) to implement the algorithm and whether the centralized or distributed implementation is considered;
		
	\end{itemize}

\begin{table*}
	\centering
	\renewcommand{\arraystretch}{1.2}
	\caption{Summary of actions in RL/DRL Model for AIoT Systems.}
	\label{table_action}
	%    \resizebox{\textwidth}{!}{
	\begin{tabular}{|p{1.3cm}|p{1.8cm}|p{3.2cm}|p{0.5\textwidth}|} 
		\hline
		\multicolumn{2}{|c|}{\textbf{AIoT Layer}} & \textbf{Examples} & \textbf{Details} \\
		\hline
		Actuator control  & Autonomous Robots & kinematic control & turn left/right, go straight/back, rotating left/right, reach specific object, steering angle, velocity    \\     \cline{3-4}
		action&   & task manipulation & position of end-effector, change in position, change in azimuthal angle, gripper open and close, termination, sweep/pick up/put down objects\\     \cline{3-4}        
		&   & charging& wander, recharge at home\\     \cline{2-4}    
		
		& Smart Vehicles  & velocity control &
		brake, accelerate, maintain (discrete) 
		\\ & & &
		velocity (continuous)
		\\ \cline{3-4}  
		& &direction control     &
		change lane (discrete);
		\\ & & & turn left, turn right, maintain (discrete)
		\\ & & & steering angle (continuous)                     
		\\             \cline{2-4}    
		& Smart Grid & ESS management&amount of charging/discharging power of the ESS.
		\\ \cline{3-4}  
		& & DG energy dispatch &
		energy dispatch decision of DGs
		\\ \cline{3-4}  
		& & energy trading amount&
		amount of energy trading with main grid or other MGs
		\\ \cline{3-4}  
		& & energy trading price&
		the price to sell/buy.
		\\ \cline{3-4}  
		& & DR devices on/off change &
		whether to change the on/off state of the DR devices    
		\\             \cline{2-4}    
		& WSAN & actuator node movement &forward, back, left, right, and stop
		\\ \cline{3-4}  
		& & power mode control &
		turn of/off the sensor, select the high/low power mode of the sensor
		%        \\ \cline{3-4}  
		%        & &sensor node control &
		%        node movement such as moving forward, making a turn.
		\\             \hline        
		\multicolumn{2}{|C{3cm}|}{Communication resource control action}  & communication resource allocation &
		bandwidth/subchannel allocation, energy allocation, IoT device scheduling, BS selection for offloading, route selection, relay selection, relay activation
		\\ \hline            
		\multicolumn{2}{|C{3cm}|}{Computation resource}&offloading decision & offload or not (binary); 
		\\ \multicolumn{2}{|C{3cm}|}{control action} & & the proportion of data to be offloaded (discrete or continuous);
		\\ \multicolumn{2}{|C{3cm}|}{} & & the number of offloaded bits (continuous).
		%        \begin{minipage}[t]{0.4\textwidth}
		%            \begin{itemize}
		%                \item binary variable: whether to offload or not;
		%                \item discrete variable: the proportion of data to be offloaded;
		%                \item continuous variable: the number of offloaded bits. 
		%            \end{itemize}
		%        \end{minipage}             
		\\
		\cline{3-4} 
		\multicolumn{2}{|c|}{} &caching decision  &whether to cache a content, which existing content to replace with
		\\
		\cline{3-4}    
		\multicolumn{2}{|c|}{} &computation resource allocation& 
		number of CPUs cores (per second), VM level, edge/cloud server selection for offloading, serve or reject a request        
		\\
		\hline
		
	\end{tabular}
	%    }
\end{table*}

\begin{table*}
	\centering
	\renewcommand{\arraystretch}{1.2}
	\caption{Summary of rewards in RL/DRL Model for AIoT Systems.}
	\label{table_reward}
	%    \resizebox{\textwidth}{!}{
	\begin{tabular}{|p{1.3cm}|p{1.8cm}|p{3.2cm}|p{0.5\textwidth}|} 
		\hline
		\multicolumn{2}{|c|}{\textbf{AIoT Layer}} & \textbf{Examples} & \textbf{Details} \\
		\hline
		
		Physical system  & Autonomous Robots &  task completion    & reach correct destination, correct operation (correct pushing/pulling, putting down/picking up of objects)        \\ \cline{3-4}    
		performance&   & task completion efficiency & backward penalty, cumulative distance traveled, overall completion time\\ \cline{2-4}    
		
		& Smart  & driving safety&
		collision avoidance
		\\ \cline{3-4}  
		& Vehicles &driving smoothness         &
		keeping on the same driving lane, avoiding unnecessary velocity adjustment
		\\ \cline{3-4}  
		& &driving efficiency             &
		driving speed, moving direction, junction waiting time, flow rate
		\\ \cline{3-4}  
		& &environmental benefits                 &
		vehicle fuel consumption, power consumption
		\\             \cline{2-4}    
		& Smart Grid & energy balance    & power balance within the MG, taking into account the charge/discharge of ESS
		\\ \cline{3-4}  
		& &DG generation cost& the cost of energy generation by DG
		\\ \cline{3-4}  
		& &ESS operation cost &losses for batteries based on charge/discharge operations
		\\ \cline{3-4}  
		& &energy trading cost/profit&the cost/profit caused by energy transaction 
		\\ \cline{3-4}  
		& &load shedding cost&the cost to meet the requirement of controllable load
		\\ \cline{3-4}  
		& &consumer's satisfaction &  level of consumer's satisfaction about the DR action
		%        \\ \cline{3-4}  
		%        & &consumer's cost & consumer's cost for energy purchasing
		%        \\ \cline{3-4}  
		%        & &broker's cost&broker's cost for energy purchasing and energy selling
		%        \\ \cline{3-4}  
		%        & &profit for entities or MG & the amount of profit that the consumer/broker/producer/MG can obtain in energy trading
		\\ \hline                
		\multicolumn{2}{|c|}{}  & reliability &
		decoding error probability, quality of the selected channel, packet loss rate, the expected correct bits per packet 
		\\ \cline{3-4}    
		\multicolumn{2}{|c|}{Communication network }& throughput &sum data rate of all the transmissions
		\\ \cline{3-4}    
		\multicolumn{2}{|c|}{performance }& connectivity &number of connected nodes
		\\
		\cline{3-4}    
		\multicolumn{2}{|c|}{} & sensing coverage & covered area, number of sensing events, field estimation error, information gain
		\\
		%        \cline{3-4}
		%        \multicolumn{2}{|c|}{}& net bit rate &
		%        \\
		\cline{2-4}    
		& \multirow{3}{*}{}  & energy consumption  & 
		energy consumption related to task/content transmission and task processing
		\\
		\cline{3-4}    
		\multicolumn{1}{|c|}{} && delay & 
		task/content transmission delay and task processing delay \\
		\cline{1-1}\cline{3-4}    
		\multicolumn{1}{|c|}{} && task drop rate & 
		probability that a task is dropped due to task queue being saturated    
		\\   \cline{3-4}  
		\multicolumn{1}{|c|}{} && load balance & 
		efficient distribution of network or application traffic across multiple servers 
		\\
		\cline{2-4}    
		\multicolumn{2}{|c|}{Computing system}& edge/cloud service cost  & 
		payment for purchasing edge/cloud service or PMs/VMs
		\\
		\cline{3-4}    
		\multicolumn{2}{|c|}{performance} & server utilization  & 
		utilization rate of edge/fog/cloud server or PMs/VMs
		\\
		\cline{3-4}    
		\multicolumn{2}{|c|}{} & content freshness  & 
		popularity of stored content
		\\    
		\hline     
		
	\end{tabular}
	%    }
\end{table*}

\subsection{AIoT Network Layer - IoT Communication Networks}

\begin{figure}[!t]
	
	\centering
	
	\includegraphics[width=0.45\textwidth]{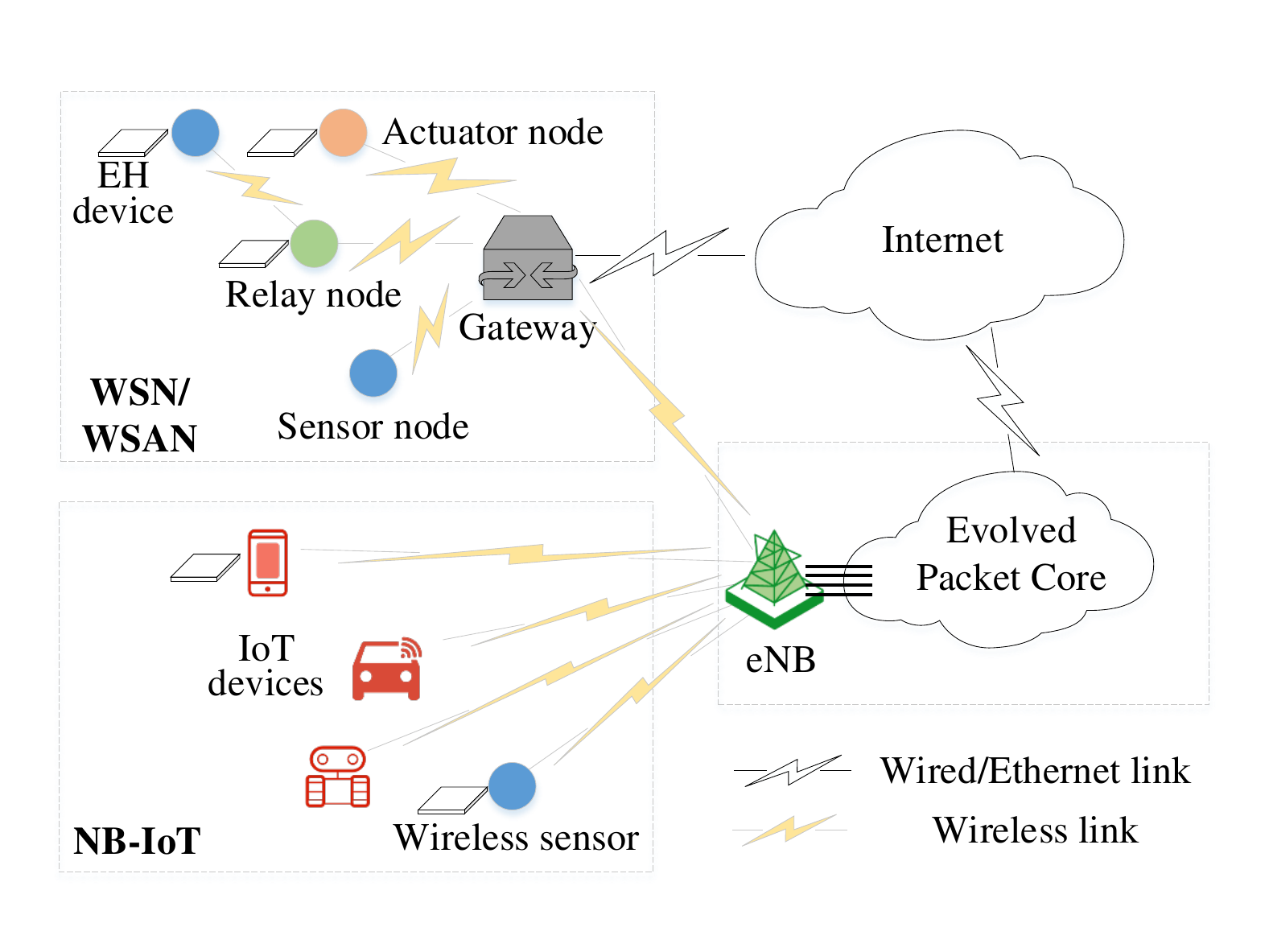}
	
	\caption{IoT communication networks system models.}
	
	\label{IoTComm1}
	
\end{figure}

A reliable and efficient wireless communication network is an essential part of the IoT ecosystem. Such wireless networks range from short range local area networks such as Bluetooth, Zigbee/IEEE 802.15.4, and IEEE 802.11 to long range wide area networks such as Narrowband Internet of Things (NB-IoT) and LoRaWAN. When designing resource control mechanisms to efficiently utilize the scarce radio resources in transmitting the huge amount of IoT data, the IoT networks need to consider the characteristics of IoT devices such as massive in number, limited in energy, memory and computation resources. Moreover, the requirements of IoT applications such as low latency and high reliability have to be taken into account as well. One of the promising approaches to develop resource control mechanisms tailored for IoT is to enable IoT devices to operate autonomously in a dynamic environment by using learning frameworks such as DRL \cite{park2016learning}. The existing works in this area mainly include studies on wireless sensor networks (WSNs), wireless sensor and actuator networks (WSAN), NB-IoT, and energy harvesting (EH). The system models of existing works are illustrated in Fig. \ref{IoTComm1}. Table \ref{IoTComm2} lists the related research works and their DRL models and DRL algorithms.

\begin{table*}[t]
	
	\centering
	
	\renewcommand{\arraystretch}{1.2}
	
	\caption{Summary of DRL Models and Algorithms in Research Works for IoT Communication Networks.}
	
	\label{IoTComm2}
	
	%    \resizebox{\textwidth}{!}{
	
	\begin{tabular}{|p{0.8cm}|c|p{2.4cm}|p{2.6cm}|p{2.6cm}|p{0.8cm}|p{1.9cm}|p{2.3cm}|} 
		
		\hline
		
		\textbf{Theme} & \textbf{Ref.} &\multicolumn{4}{c|}{\textbf{DRL Model}} & \textbf{DRL} & \textbf{Agent }\\
		
		\cline{3-6}
		
		&    & \textbf{State} & \textbf{Action} & \textbf{Reward} & \textbf{Feature} & \textbf{Algorithm}&  \textbf{Location} \\
		
		\hline
		
		\multirow{5}*{\rotatebox[origin=c]{270}{WSN}}& \cite{kwon2019intelligent} & topology state & communication resource allocation & throughput, energy consumption & basic &DDQN & IoT device (centralized) \\
		
		\cline{2-8}
		
		& \cite{renaud2006coordinated}& topology state & power mode control& energy consumption, sensing coverage& MA & fully distributed Q-Learning, etc. &  sensor controller (distributed) \\
		
		\cline{2-8}
		
		& \cite{chen2019deep} & topology state & sensor node control & sensing coverage & basic &deep reinforced learning tree (DRLT) & sensor controller (centralized) \\
		
		\cline{2-8}
		
		& \cite{su2019cooperative} & channel state, energy consumption state & communication resource allocation & energy consumption, sensing coverage & basic &DQN & sensor controller (centralized) \\
		
		\cline{2-8}
		
		& \cite{zhu2017new} & task state, channel state & communication resource allocation & throughput & basic &Deep Q-Learning & relay node (centralized) \\
		
		\hline
		
		\multirow{3}*{\rotatebox[origin=c]{270}{WSAN}}&  \cite{oda2017design} & topology state & actuator node movement& sensing coverage & basic &  DQN & actuator node (centralized) \\ \cline{2-8}
		
		& \cite{kunzel2018weight}& topology state&actuator node movement& reliability, delay & basic & Q-Learning & routing agent (centralized) \\
		
		\cline{2-8}
		
		&\cite{leong2018deep}& sensor state, channel state & communication resource allocation & reliability& basic & DQN& network controller (centralized) \\
		
		\hline
		
		\multirow{2}*{\rotatebox[origin=c]{270}{NB-IoT}}& \cite{jiang2019cooperative} & channel state & communication resource allocation& reliabilty & MA, POMDP &  CMA-DQN & NB-IoT devices (distributed) \\ \cline{2-8}
		
		& \cite{chafii2018enhancing} & channel state&communication resource allocation& reliability& basic & upper confidence band (UCB) & NB-IoT device (centralized) \\
		
		\hline
		
		\multirow{4}*{\rotatebox[origin=c]{270}{EH}}&  \cite{lei2016iwsn} & task queue state, channel state, energy queue state & communication resource allocation& reliability, delay & basic &  AMDP+OSL & fusion center (centralized) \\ \cline{2-8}
		
		&  \cite{chu2018reinforcement}  &  channel state, energy queue state& communication resource allocation & reliability & basic & LSTM-based DQN & BS (centralized) \\ \cline{2-8}
		
		& \cite{li2019partially} & energy queue state & communication resource allocation& energy consumption & POMDP &  DDQN & BS (centralized) \\  \cline{2-8}
		
		& \cite{qiu2019deep} & energy consumption state, sensor state & communication resource allocation& reliability & basic &  DDPG & sensor controller (centralized) \\ 
		
		\hline
		
	\end{tabular}
	
	%    }
	
\end{table*}

\subsubsection{Wireless Sensor Networks}

Wireless sensor network (WSN) is a wireless network of many tiny disposable low power sensors. WSNs are expected to be integrated into the IoT systems, where sensor nodes join the Internet dynamically and use it to collaborate and accomplish their tasks.

	\begin{figure}[!t]
		
		\centering
		
		\includegraphics[width=0.45\textwidth]{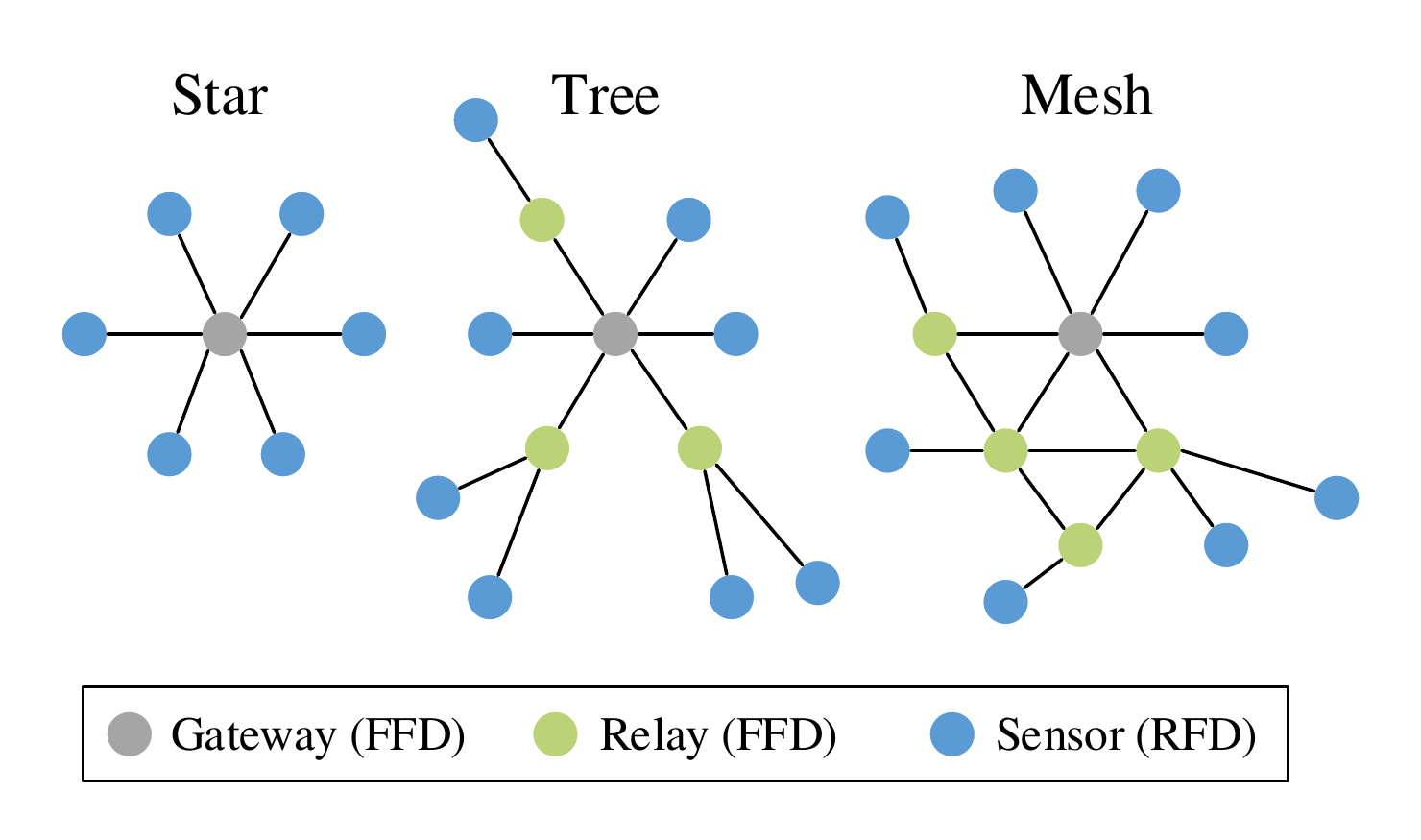}
		
		\caption{Three types of network topology types for WSN.}
		
		\label{wsn}
		
	\end{figure}

	%Wireless sensor networks (WSNs) offer practical applications that can directly benefit from artificial intelligence technology. For a large scale IoT application, sensors are needed in huge number.
	
	%In \cite{kumar2018power}, RL is used for modeling the sensors in the physical, routing and network layer. The routing and networking layer deals with the communication capabilities of the sensors. The resource scheduling issues among the sensors are solved in order to optimize the lifetime of the sensors, energy usage, and communication costs.
	
	%A multi-agent system approach on wireless sensor networks is able to tackle the resource constraints in these networks by efficiently coordinating the activities among the nodes. 
	
	%In \cite{renaud2006coordinated}, the authors consider the coordinated sensing coverage problem and study the behavior and performance of four distributed DRL algorithms, i.e., fully distributed Q-learning, distributed value function (DVF), optimistic DRL, and frequency maximum Q-Iearning (FMQ). Their performance in terms of communication and computational costs, energy consumption, and sensor coverage levels are evaluated and compared. The authors in \cite{ding2019deep} leverage DRL for router selection in a wireless network with heavy traffic. Compared with existing routing algorithms, the proposed algorithms achieve higher network throughput due to the low congestion probability.

	A WSN mainly consists of three types of nodes, i.e., sensor, relay and gateway, which can be organized into three types of network topology as shown in Fig. \ref{wsn}. In the star topology, each sensor node connects to the gateway, which means the network is highly dependent on the central gateway. In the tree topology, the system is arranged in a top-down structure. When a parent node fails to work effectively, all its child nodes will be affected. In the mesh topology, the network nodes are interconnected with one another forming a mesh structure, where the data from sensor nodes can be forwarded by the relay nodes in a multi-hop fashion to the gateways.\par

	In a mesh topology, network connectivity has an important effect on network performance such as throughput. Generally speaking, better connectivity can be achieved by activating more relay nodes in the network at the cost of larger energy consumption. Therefore, it is important to adjust the wireless transmission range of relay nodes to maximize the throughput while minimizing the energy consumption for the network. \cite{kwon2019intelligent} proposes an autonomous network formation solution that enables an IoT device to make the decision on whether to activate its transmission or not based on the multi-hop ad hoc network topology state. Distributed implementation is considered where each IoT device is an agent that applies the DDQN algorithm to make decisions based only on the local system state, i.e., the number of nodes in its transmission range. \par

	Cooperative communication is recognized as an important technology to improve the WSNs' performance in data transmission rate and node coverage. A typical WSN system model with cooperative communication consists of a sensor node, multiple relay nodes, and a gateway, where the best relay node to forward data from sensor to gateway needs to be determined. In \cite{su2019cooperative}, the process of cooperative communication with relay selection is modeled as an MDP, where the channel state and energy consumption for signal broadcasting are used as state and an appropriate relay is selected to participate in the cooperative communication by a DQN type algorithm. The reward is determined by energy consumption and information gain. \par

	The cognitive network is envisioned as one of the key enablers for the IoT, which can tackle the problem of the crowded spectrum for the rapidly increasing amount of IoT applications. A cognitive network consists of cognitive nodes that can sense the environment and make intelligent decisions to seize the opportunities to transmit. \cite{zhu2017new} leverage the DRL technique to enhance the packet transmission efficiency in cognitive IoT. The system model considers one relay node which gathers packets from multiple sensor nodes and sends to the gateway via $M$ channels. According to the queue state and channel state, the relay makes channel selection decisions for packet transmission. The goal of the policy is to maximize the throughput and minimize the energy consumption, and a deep Q-Learning based algorithm is proposed to solve the problem.\par

	One important application of the sensor nodes is to properly cover an area in order to make sure that all important events which occur in that area can be accurately detected by at least one sensor. This is referred to as the sensor coverage problem. An MA system approach on WSNs is able to tackle the resource constraints in these networks by efficiently coordinating the activities among the nodes. \cite{renaud2006coordinated} tackles the coordinated sensing coverage problem in an MA system, where each sensor acts as an agent. The state obtained by the sensor is the binary sensing status of its area. There are three discrete actions in the action space: turn off the sensor, turn on the sensor in low power mode and turn on the sensor in high power mode. The reward function for each agent is based on both the gain of sensing coverage and energy consumption resulting from the action. The authors study the behavior and performance of four distributed DRL algorithms, i.e., fully distributed Q-Learning, distributed value function (DVF), optimistic DRL, and frequency maximum Q-Learning (FMQ).\par

	\cite{chen2019deep} also studies the sensing coverage problem in WSNs with a focus on mobile robotic sensor nodes. The position of the moving sensor is taken as the state and an action corresponds to the movement of the mobile sensor. Higher information gain according to the location of the sensors can lead to higher rewards in the DRL model. To accelerate the exploration process and find near-optimal sampling locations for the mobile sensors, deep reinforced exploring learning tree (DRLT) is designed and outperforms other field exploration algorithms, such as rapidly exploring random tree (RRT) and rapidly exploring tree with linear reduction (RRLR).\par

%{\color{red} Since the massive device connectivity in mesh topology will produce more network traffic data in the future, the existing routing protocols are becoming unsuitable for the network with high data rate and ultrashort latency. The main reason for this is that those routing protocols only calculate the shortest path without considering the remaining buffer size of every router. When the amount of data increases, the network may suffer from heavy traffic load and even network congestion. In \cite{ding2019deep}, DRL is used to solve the router selection problem in the network with heavy traffic. The agent is the data transmission task which receives the packet size, remaining buffer size and data source as the state. The action is defined as the selection for the next router. The reward considers the network congestion and transmission path length. Two DQN-based algorithms are designed to solve the DRL model. One is source-destination multi-task DQN (SDMT-DQN) as illustrated in Fig. \ref{sdmt}, which can learn from past experiences and update routing policies in real-time. The other is destination-only multi-task DQN (DOMT-DQN) which has better performance in shortening the path length while maintaining the congestion probability at a relatively low level.}

%\begin{figure}[!t]

%    \centering

%    \includegraphics[width=0.45\textwidth]{SDMT-DQN.pdf}

%    \caption{{\color{red}SDMT-DQN algorithm.}}

%    \label{sdmt}

%\end{figure}

\subsubsection{Wireless Sensor and Actuator Networks}

Wireless sensor and actuator networks (WSANs) are composed of a large number of sensor nodes with low power, one or more actuators, and a processing unit. They are envisioned as an important part of the IoT ecosystem and have a panoply of applications, ranging from industrial automation to homeland security. Unlike conventional WSNs, sensor and actuator nodes must work together closely to collect and forward data, and act on any sensed data collaboratively, promptly and reliably to react to the physical world. \par

	Scheduling transmissions is one of the challenges in WSAN from a networking perspective because of the volatile nature of wireless channels. Wireless transmission is scheduled from sensors to the gateway and from the gateway to the actuators over a shared medium. To solve this problem, \cite{leong2018deep} formulates a DRL-based sensor scheduling problem for allocating wireless channels to sensors for remote state estimation of dynamical systems. The problem is formalized as an MDP and solved by the DQN algorithm. In this work, the network controller allocates communication resources based on the sensor state and channel state. The reward function is calculated by the reliability.\par

	WSANs, e.g., ISA SP100.11a and WirelessHART, have special devices known as network managers that perform tasks such as admission control of devices, the definition of routes, and allocation of communication resources. The state-of-art routing algorithms used in these protocols usually have different weights for different route preferences. Weight adjustment can be challenging because of the dynamicity of wireless networks. RL/DRL models can be used for weight adjustment with a consideration of current application requirements and communication conditions. In \cite{kunzel2018weight}, a global routing agent with Q-Learning is proposed for weight adjustment of the state-of-the-art routing algorithm, aiming at achieving a balance between the overall delay and the lifetime of the network. The routing agent receives network topology state information including the weight of the number of hops and the weight of the energy source. It makes decisions on whether to change into a neighbor state or to keep the current state. The reward is determined by the expected network lifetime and average network latency.\par

	Similar to the mobile sensor movement control in WSNs \cite{chen2019deep}, automatic control of node mobility is also essential in WSANs. Several performance metrics, such as connectivity, coverage, energy consumption and accuracy, can be improved by moving the nodes in the networks. \cite{oda2017design} focuses on the connectivity performance which is essential to conducting collaborative tasks among the actuator nodes. Specifically, the authors present the design and implementation of a simulation system based on DQN for mobile actor node control in a WSAN. The actuator node takes the network topology state as the input state and makes the decision on mobile actor node movement from five discrete patterns, including stop, forward, back, left and right. The reward is measured by sensing coverage for each action.\par

%The authors in \cite{oda2017design} present the design and implementation of a simulation system based on DQN for mobile actor node control in a WSAN. In \cite{kunzel2018weight}, a global routing agent with Q-Learning is proposed for weight adjustment of the state-of-the-art routing algorithm, aiming at achieving a balance between the overall delay and the lifetime of the network.

%The study in \cite{leong2018deep} focuses on a DRL-based sensor scheduling problem for allocating wireless channels to sensors for the purposes of remote state estimation of dynamical systems. The algorithm can be run online, and is model-free with respect to the wireless channel parameters.

\subsubsection{NB-IoT}

NB-IoT is a technology proposed by 3GPP in Release-13. It offers low energy consumption and extensive coverage to meet the requirements of a variety of social, industrial and environmental IoT applications. Compared to legacy LTE technologies, NB-IoT chooses to increase the number of repetitions of transmission to serve users in deep coverage. However, large repetitions can reduce system throughput and increase the energy consumption of IoT devices, which can shorten their battery life and increase their maintenance costs.

	Radio resource allocation in NB-IoT specifies the number of radio resources allocated to each group of devices in a Transmission Time Interval (TTI). NB-IoT includes two types of uplink channels, namely, Narrowband Physical Random Access CHannel (NPRACH) and Narrowband Physical Uplink Shared CHannel (NPUSCH). At the beginning of each uplink TTI, the evolved Node B (eNB) selects a configuration that specifies the radio resource allocation in order to accommodate the NPRACH procedure with the remaining resources used for data transmission. However, it is a challenge to balance the channel resource allocation between the NPRACH procedure and data transmission. To solve this uplink resource configuration problem, a Cooperative MA DQN (CMA-DQN) approach is developed in \cite{jiang2019cooperative}, in which each DQN agent independently controls a configuration variable for each group, in order to maximize the long-term average number of working IoT devices in NB-IoT. As the eNB can only observe the channel state at the end of each TTI, the problem is formalized as POMDP and historical information is used for current state prediction. The state is represented by channel state information of the last $M$ TTIs. Each agent receives the same common reward at the end of the current TTI. The common reward can ensure that all the agents are aiming at maximizing the number of devices that transmit data successfully in NB-IoT. Multiple agents are trained in parallel in CMA-DQN and the weight matrix is updated by using DDQN.\par

	Enhancing the coverage and reducing energy consumption are the key targets for NB-IoT. The major state-of-art solutions are repeating transmission data and control signals, which lead to system throughput reduction as well as spectral efficiency loss. In \cite{chafii2018enhancing}, the authors propose a new method based on the RL algorithm to enhance NB-IoT coverage. Instead of employing a random spectrum access procedure, dynamic spectrum access can reduce the number of required repetitions, increase the coverage, and reduce energy consumption. The agent receives two values 0 or 1 as the reward, which means that the selected channel is occupied or vacant.

%In \cite{chafii2018enhancing}, the authors propose a new method based on the RL algorithm to enhance NB-IoT coverage. Instead of employing a random spectrum access procedure, dynamic spectrum access can reduce the number of required repetitions, increase the coverage, and reduce energy consumption.

%{\color{blue}In \cite{he2018green}, the authors present a green resource allocation algorithm based on DRL to improve the accuracy of QoE adaptively in the green services of content-centric IoT. Comment: Is this NB-IoT?}

\subsubsection{Energy Harvesting}

Energy Harvesting (EH) is a promising technology for the long-term and self-sustainable operation of the IoT devices. While EH is a promising technique to extend the lifetime of IoT devices, it also brings new challenges to resource control due to the stochastic nature of the harvested energy.

	The uncertainty of the harvested energy poses challenges to the reliability of EH systems, which is essential for a number of industrial applications. In \cite{lei2016iwsn}, the energy management policy in an industrial WSN is investigated to minimize the weighted packet loss rate under the delay constraint, where the packet loss rate considers the lost packets both during the sensing and transmission processes. A centralized fusion center (FC) takes task queue state, channel state and energy queue state as the input state. At each time slot, a sensor scheduling action and a transmission energy allocation action are chosen from the discrete action space. The problem is formulated into an MDP model, and stochastic online learning is applied to derive a distributed energy allocation algorithm with a water-filling structure and a scheduling algorithm by an auction mechanism. \par

	One way to deal with the uncertainty of harvested energy is through battery level prediction in EH-based systems. \cite{chu2018reinforcement} considers an uplink transmission scenario with multiple EH user equipment (UEs) and a BS with limited access channels. The authors model the access control based on battery prediction as an MDP. The channel state and energy queue state are employed as the input state to the BS, which then outputs the action according to the scheduling policy. As the performance of the model relies on both the battery prediction result and the access control policy, the reward takes the sum rate of the transmissions into account. A two-layer LSTM-based DQN control network is proposed to solve the problem. The first layer is an LSTM-based network to perform the battery prediction. The second layer takes the battery prediction result, channel state and energy queue state as the input and outputs the action for producing the access control policy.\par

	The uncertainty of harvested energy can also be captured by formulating a POMDP problem for the EH-based systems. In \cite{li2019partially}, BS is considered as an agent to schedule the IoT devices after receiving the energy queue states of some of the nodes. The amount of energy consumption is considered as the reward. DDQN algorithm is adopted to solve this POMDP-based problem.\par

	While most algorithms for EH are value-based methods, \cite{qiu2019deep} proposes an algorithm based on DDPG which can tackle the energy management problem in a continuous space. The sensor controller receives the energy consumption state and bit error rate of the sensor and determines the transmission energy allocation for the sensor. The reward is measured by the net bit rate.\par

%\cite{chu2018reinforcement} studies the joint access control and battery prediction problem in a small-cell IoT system including multiple EH user equipments (UEs) and a base station (BS) with limited uplink access channels. A DQN-based scheduling algorithm that maximizes the uplink transmission sum rate is proposed. For the battery prediction problem, using a fixed round-robin access control policy, an RL-based algorithm is developed to minimize the prediction loss without any model knowledge about the energy source and energy arrival process. In \cite{lei2016iwsn}, the energy management policy in an industrial wireless sensor network is investigated to minimize the weighted packet loss rate under the delay constraint, where the packet loss rate considers the lost packets, both during the sensing and delivering processes. The problem is formulated into an MDP model, and stochastic online learning with a post-decision state is applied to derive a distributed energy allocation algorithm with a water-filling structure and a scheduling algorithm by an auction mechanism.\par

	\subsubsection{Comparison and Insights}
	
	By summarizing and comparing the above literature, the following insights can be obtained.

	\begin{itemize}
		
		\item \textit{System model}: Most of the research work focus on star topology since the agent control in a single-hop network is relatively simple. On the other hand, DRL-based solutions for mesh topology including cellular networks with D2D communications can be studied more.
		
		\item \textit{DRL model}: In terms of the system state, the channel state is usually included due to the time-varying nature of the wireless channel. Examples of typical channel state include SINR, pathloss, channel gain, data transmission rate as given in Table \ref{table_state}. Another common system state is the transmission queue state, especially when the data packets arrive according to a dynamic process. When EH is considered, the energy queue state is normally an important component of the system state. In routing problems for mesh networks, the topology state usually has overall information about the nodes in the network which can be helpful for the routing agent \cite{kunzel2018weight}. In terms of action, most research works focus on communication resource allocation. Some literature in WSAN takes actuator control as action. Typical reward functions include throughput, reliability, energy consumption, sensing coverage.
		
		\item \textit{DRL algorithm}: DQN and novel DQN-based DRL algorithms are most frequently adopted in existing literature as can be observed from Table \ref{IoTComm2}. This is partly due to the fact that the action space in most existing works is discrete. As discussed in Section II, value-based methods are simpler than actor-critic methods and easier to converge. However, value-based methods cannot be applied to continuous action space unless the continuous actions are discretized, which results in loss of performance and curse-of-dimensionality problem. 
		
		\item \textit{Implementation}: As shown in Table \ref{IoTComm2}, most DRL algorithms are centrally implemented at the BS, gateway, etc., while a few MA-based DRL algorithms are distributively implemented at sensors or IoT devices. As sensors and IoT devices normally have limited energy, computation, and storage resources, the algorithm design needs to consider the trade-off between performance and complexity. The energy spent on communication and computation by the five agents on Crossbow Mica2 motes during the learning phase (first 20,000 iterations) is compared between different DRL algorithms in \cite{renaud2006coordinated}. It is shown that although DRL algorithms based on the global system state can achieve the best performance, the incurred communication and computation overhead is too large. It is suggested that performing DRL algorithms based on local system state in resource constraint IoT devices is a more viable option.
		
		%    \item \textit{AIoT layer}: Most existing research focuses only on the network layer, while a few works of literature consider both the perception layer and the network layer in WSAN. It will be interesting to further explore the resource control of IoT communication network for some specific real-world IoT applications.

	\end{itemize}

\subsection{AIoT Application Layer - IoT Edge/Fog/Cloud Computing Systems}
\begin{figure}[!t]
	\centering
	\includegraphics[width=0.45\textwidth]{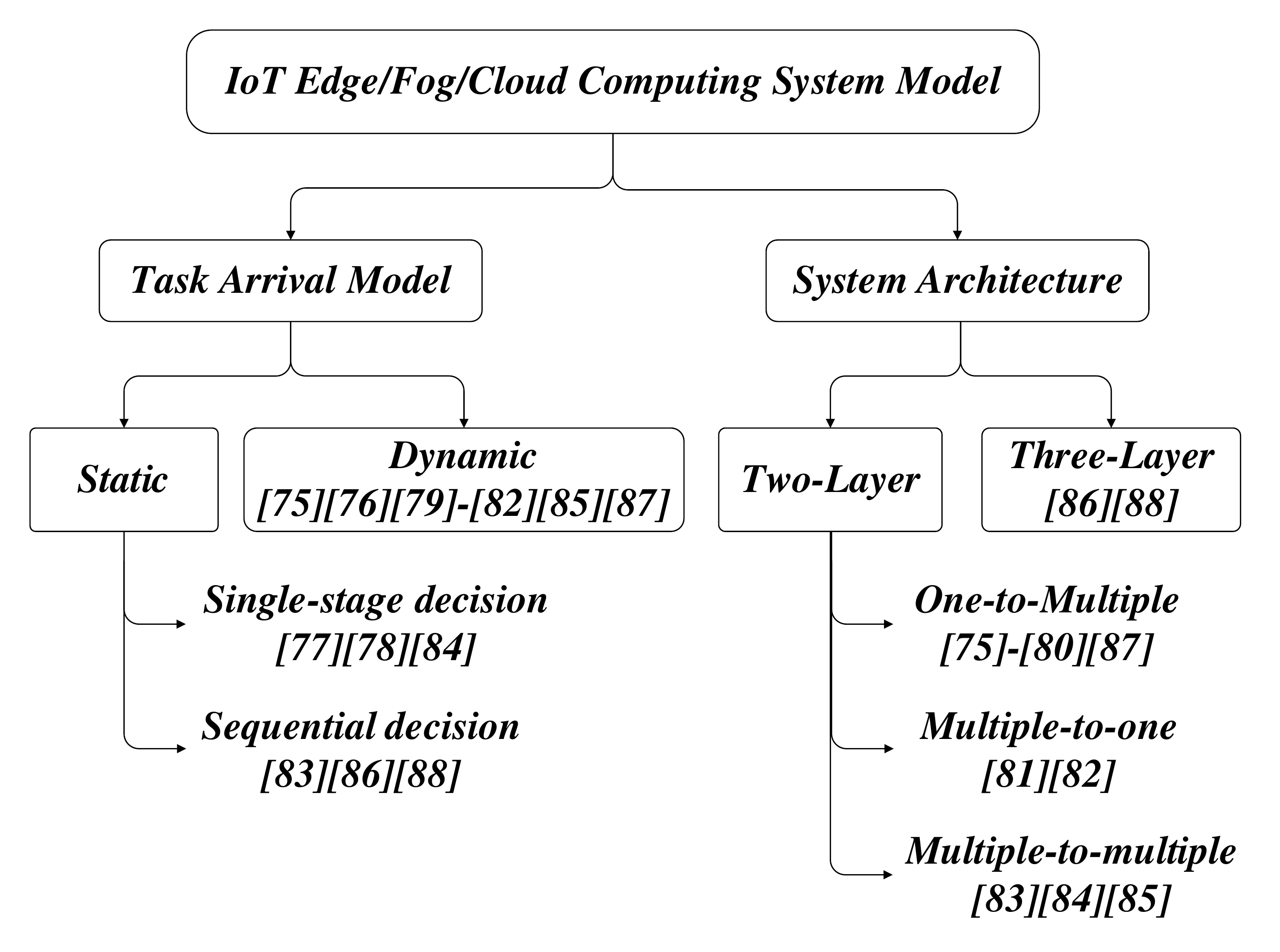}
	\caption{Classification of IoT edge/fog/cloud computing system models.}
	\label{edge1}
\end{figure}

	Edge/fog/cloud computing is a crucial technique to process and analyze the huge amount of sensory data in IoT. In such systems, IoT devices can offload the computationally intensive tasks to the edge/fog/cloud servers. Moreover, caching IoT data at the network edge is considered to be able to alleviate the congestion and delay in transmitting IoT data through wireless networks. The above problems have been widely studied by applying DRL techniques. We classify the existing research based on the different considerations in system models as shown in Fig. \ref{edge1}. Table \ref{table_edge2} lists the related research works and their respective DRL models and algorithms. \par

	\begin{table*}[t]
		\centering
		\renewcommand{\arraystretch}{1.2}
		\caption{Summary of DRL Models and Algorithms in Research Works for IoT Edge/Fog/Cloud Computing Systems.}
		\label{table_edge2}
		%    \resizebox{\textwidth}{!}{
		\begin{tabular}{|p{0.8cm}|c|p{2.5cm}|p{2.5cm}|p{2cm}|p{2cm}|p{2.2cm}|p{1.7cm}|} 
			\hline
			\textbf{Theme} & \textbf{Ref.} &\multicolumn{4}{c|}{\textbf{DRL Model}} & \textbf{DRL Algorithm} & \textbf{Agent }\\
			\cline{3-6}
			&    & \textbf{State} & \textbf{Action} & \textbf{Reward} & \textbf{Environment} & &  \textbf{Location} \\
			\hline
			\multirow{13}*{\rotatebox[origin=c]{270}{Task Offloading and Resource Allocation}}& \cite{He2018} & task state, queue state & offloading decision (discrete) & delay & application layer &  Deep Q-Learning & edge server (centralized) \\ \cline{2-8}
			& \cite{Liu2019} & channel state, task queue state, edge server state & offloading decision (discrete) & delay, energy consumption& application layer & Deep Q-Learning & IoT devices (distributed) \\
			\cline{2-8}
			& \cite{huang2019deep} & offloading decision, communication resource allocation & movement among two neighboring states (discrete) & delay, energy consumption & application layer, communication layer  & DQN & edge server (centralized) \\
			\cline{2-8}
			& \cite{Huang2019} & channel state & offloading decision (discrete) & delay & application layer & deep actor network with supervised learning & edge server (centralized) \\
			\cline{2-8}
			%            & \cite{chen2018decentralized} & queue state, channel state & offloading, continuous &  delay, energy consumption & application layer, communication layer  & DDPG & IoT device (distributed) \\
			%            \cline{2-8}
			& \cite{Lei2019} & queue state, task state & offloading decision, communication resource allocation (discrete) & delay, energy consumption & application layer, communication layer  & value-based DRL with value function decomposition & edge server, IoT device (semi-distributed) \\
			\cline{2-8}
			& \cite{Qiu2019} & channel state, queue state, topology state, task state, edge server state   & offloading decision, computation resource allocation (discrete) & delay, energy consumption, cloud service cost & application layer  & actor-critic with ACA for exploration & edge server, centralized \\
			\cline{2-8}
			& \cite{Chen2019c} & channel state, queue state, energy queue state  &  offloading decision, computation resource allocation, energy allocation (discrete) & delay, task drop rate, edge service cost & application layer  & DDQN with Q-function decomposition & centralized network controller (centralized) \\
			\cline{2-8}
			& \cite{Min2019} & channel state, energy queue state  &  offloading decision (discrete) & delay, energy consumption & application layer  & DQN & edge server, centralized  \\
			\cline{2-8}
			& \cite{Chen2019d} & edge server state, task state, network state  &  offloading decision, communication resource allocation, computation resource allocation (discrete),  & delay, energy consumption & application layer, communication layer, application layer  & MCTS+MLT & centralized network controller (centralized)  \\
			\cline{2-8}
			& \cite{wang2018smart} & computation resource allocation  &  adjustment of states (discrete) & delay, load balance & application layer, communication layer  & DQN & SDN controller (centralized)  \\
			\cline{2-8}
			& \cite{Ren2019} & channel state, queue state, energy queue state  &  offloading decision, communication resource allocation (discrete),  & delay, energy consumption & application layer, communication layer  & DDQN+FL & edge servers (distributed)  \\
			\cline{2-8}
			& \cite{cheng2019space} & task state, channel state  &  offloading decision (discrete) &  delay, energy consumption, server utilization & application layer  & deep actor-critic & cloud server (centralized) \\
			\cline{2-8}
			%            & \cite{quan2018novel} &  task state, cloud server state  &  offloading, computation resource allocation (discrete) &  delay, server usage & application layer   &  DQN &cloud server, centralized  \\
			%%            \cline{2-8}
			%            & \cite{Nassar2019} & fog server state, task state  &  computation resource allocation (discrete) &  delay & application layer  &  Q-learning, SARSA, Expected-SARSA & fog server (centralized) \\
			\hline
			\multirow{2}*{\rotatebox[origin=c]{270}{Caching}} & \cite{zhu2018caching} &   content state &   caching decision, discrete & delay & application layer   & deep actor-critic & edge server (centralized)  \\
			\cline{2-8}
			& \cite{wei2018joint} &  task state, content state, channel state, edge router state   & offloading decision, caching decision, communication resource allocation, computation resource allocation (discrete) &delay, content freshness & application layer, communication layer & natural actor-critic & edge router (centralized)  \\    
			\hline
		\end{tabular}
		%    }
	\end{table*}

    \subsubsection{Task Offloading and Resource Allocation}
Reasonable decisions are required to be made on whether to offload the computation tasks to the edge/fog/cloud servers or perform them locally at the IoT devices. Moreover, a proper amount of communication and computation resources need to be allocated for the transmission and processing of each task. In the following, we will identify several important system model considerations that have an important impact on the formulation of the corresponding DRL models.\par

\begin{figure}[!t]
	\centering
	\includegraphics[width=0.5\textwidth]{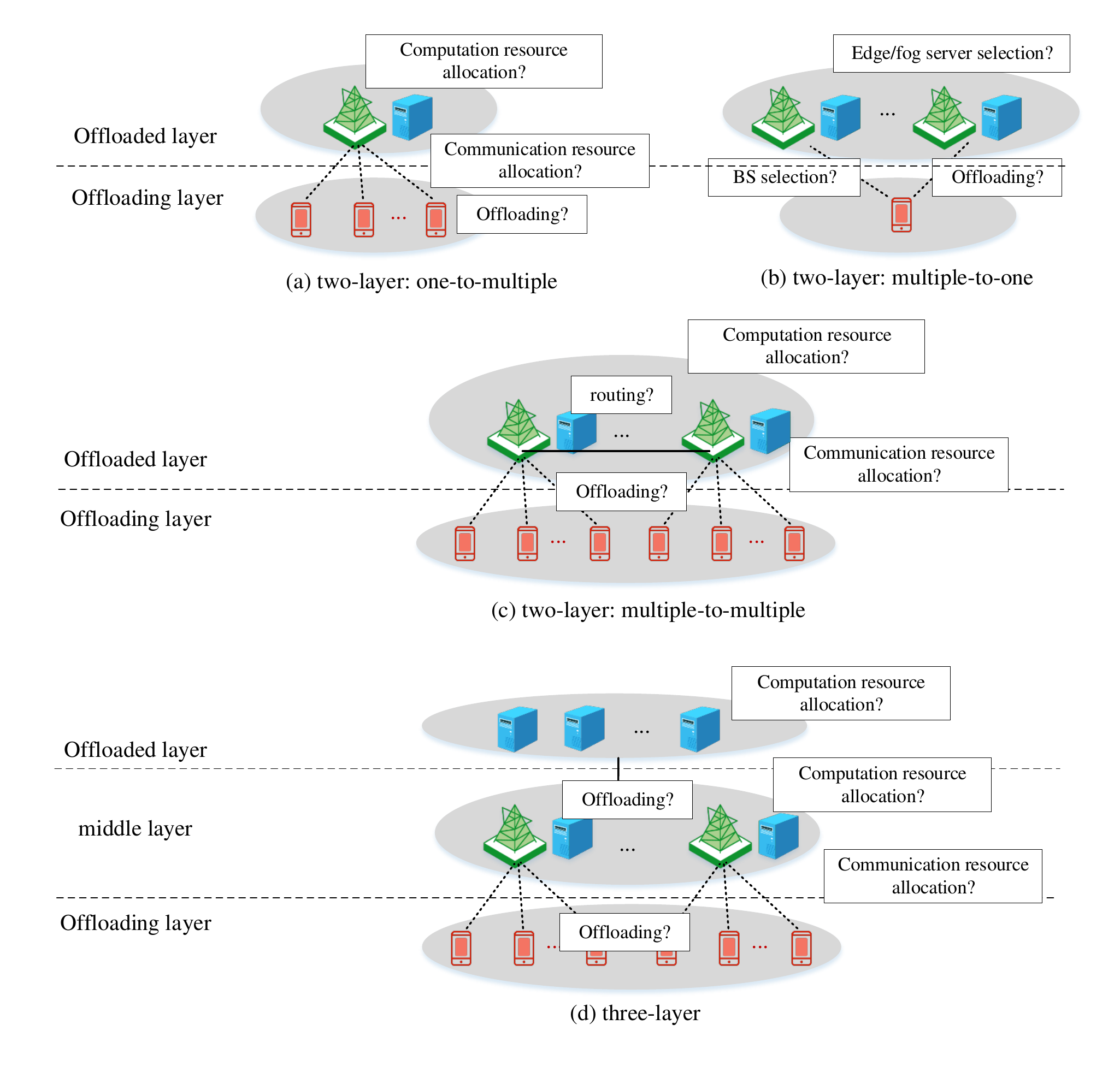}
	\caption{IoT edge/fog/cloud computing system architecture.}
	\label{edge2}
\end{figure}

\paragraph{System architecture}
The system architectures considered in existing works can be classified into two-layer and three-layer as illustrated in Fig. \ref{edge2}. A two-layer architecture has an offloading layer and an offloaded layer. Some examples of offloading vs. offloaded layers are IoT device layer vs. edge/fog/cloud layer and edge/fog layer vs. cloud layer. Fig. \ref{edge2} uses the IoT device layer vs. edge/fog layer as an example. On the other hand, a three-layer architecture has a middle layer (i.e., edge/fog layer) in addition to the offloading layer (i.e., IoT device layer) and offloaded layer (i.e., cloud layer). An example of three-layer architecture is considered in \cite{cheng2019space}, where a DRL-based computing offloading approach is proposed to learn the optimal offloading policy in a space-air-ground integrated network (SAGIN). The flying unmanned aerial vehicles (UAVs) and satellites provide access to edge computing and cloud computing, respectively. The offloading decisions need to determine whether the IoT devices should offload data to the edge server, and also whether the edge server should further offload data to the cloud.  \par

The two-layer architecture can be further classified according to the number of nodes in the offloaded and offloading layers, namely, one-to-multiple, multiple-to-one, and multiple-to-multiple as shown in Fig. \ref{edge2}.     In one-to-multiple architecture, resource allocation is usually determined jointly with the offloading decisions for each node in the offloading layer (e.g., IoT device). A joint task offloading decision and bandwidth allocation optimization method based on DQN is designed for the edge computing system in \cite{huang2019deep}. The overall offloading cost is evaluated in terms of energy cost, computation cost, and delay cost. The authors in \cite{Huang2019} consider wireless powered edge computing system that optimally adapts offloading decisions and communication resource allocations to the time-varying wireless channels. Only the offloading decisions are determined by DRL algorithms, while the wireless resource allocations are derived by convex optimization. Some works focus only on offloading decisions. In \cite{Liu2019}, the offloading decision is determined by each IoT device as an agent based on the channel state, task queue state, and its remaining computation resources. A $\epsilon$-greedy Q-Learning algorithm is adopted to optimize the delay and energy consumption. The computing offloading problem in an LTE-U-enabled network is considered in \cite{He2018}, which determines whether the task on an IoT device is carried out locally or is offloaded to the LTE-U BS based on queue state and task priority. A Deep Q-Learning algorithm is adopted to solve the problem. A blockchain-empowered edge computing system is considered in \cite{Qiu2019}, where the offloading decisions are made for both mining tasks as well as data processing tasks.\par 

%     A one-to-one architecture is considered in \cite{Nassar2019} for a Fog RAN system, where there is one fog node determining whether to serve a request locally or offload it to the cloud. Several RL methods, e.g., Q-learning, SARSA, Expected SARSA, and Monte Carlo is applied to solve the above problem. \par

In multiple-to-one architecture, the main objectives are usually to determine for a representative IoT device whether and to which edge/cloud servers to offload its tasks. In \cite{Min2019}, a DQN-based offloading scheme is proposed to select the edge server and proportion of data to be offloaded for an IoT device with EH. On the other hand, \cite{Chen2019c} considers a representative mobile user served by multiple BSs connected to a single edge server in an ultra-dense RAN. The problem of selecting the proper BS via which to offload data from mobile users to the edge server is tackled by a DQN-type algorithm. \par

%    In \cite{quan2018novel}, a cloud computing system with multiple physical machines (PMs) serving a representative end-user is studied. The objective is to select the best PM for data offloading considering the delay and utilization rate of the PMs.

In multiple-to-multiple architecture, routing and load balance between different nodes in the offloaded layer need to be considered. A collaborative edge computing system is considered in \cite{Chen2019d}, where multiple edge computing servers collaboratively perform distributed computing. Each edge computing server serves multiple mobile devices (MDs), and when it received an offloaded task form an MD, it can choose to further offload it to other collaborative edge computing servers. The state of the DRL model includes network state and task characteristics. The action involves determining MD offloading rate, edge computing server offloading rate, communication resource allocation, and computation resource allocation. The reward is designed to optimize the performance over delay and power consumption. Instead of using baseline DRL algorithms, a DNN is trained to predict the resource allocation action in a self-supervised learning manner, where the training data is generated from the searching process of Monte Carlo tree search (MCTS) algorithm. Moreover, Multitask Learning (MTL)-Based Action Prediction method is adopted, which improves the traditional DNN by splitting the last layers of DNN to construct a sub NN for supporting higher action dimensions. A similar system model is considered in \cite{wang2018smart}, where a DQN-based scheme is proposed to allocate computation and communication resources for an edge computing system with multiple edge servers and mobile users, in order to reduce the delay and achieve load balance. The authors in \cite{Ren2019} study an edge computing system where there are multiple IoT devices with EH capabilities. Each IoT device determines where to perform a task, i.e., whether and which edge computing server to offload; and how many energy resources should be allocated based on the channel state, task queue state, and energy queue state. In order to reduce the transmission costs between the IoT devices and edge nodes, federated learning (FL) is used to train DRL agents in a distributed fashion.    \par

\paragraph{Task arrival model}    
The computation task arrival models considered in the existing literature can be divided into static versus dynamic. The static task arrival models consider that the number of tasks in the system is a fixed value, while the more practical dynamic task arrival models \cite{Liu2019,He2018,Lei2019,Qiu2019,Chen2019c,Min2019,Ren2019} consider that tasks arrive according to a stochastic/deterministic process and are buffered in a queue if cannot be processed immediately upon arrival. The static task arrival model can be further divided into two types. In single-stage decision static models \cite{huang2019deep,Huang2019,wang2018smart}, single-stage decisions for all the tasks are determined in one shot simultaneously. In sequential decision static model \cite{Chen2019d,cheng2019space}, offloading and resource allocation decisions for the fixed number of tasks in the system are determined sequentially until all the tasks are processed or all the resources are occupied.  \par 

For dynamic and sequential decision static models, the control decisions for a particular task need to consider their impacts on the future tasks in terms of the long-term average performance of the system. They belong to the sequential decision problems, where DRL techniques provide a powerful tool in dealing with them. On the other hand, the single-stage optimization problems in a single-stage static model are usually formulated as the mixed linear programming problems, where DRL is considered as a better tool than the heuristic algorithms. \par

\paragraph{Centralized vs. distributed implementation}  
Most of the above DRL algorithms are considered to be centrally implemented in the edge/fog/cloud server or central control unit. This means that the IoT devices need to report their local states to the central control unit so that the latter can perceive the global system state. As the state space of the DRL model grows exponentially with the number of IoT devices, the computation complexity and communication overhead can be overwhelming when the number of IoT devices are large.\par 

From this perspective, it seems that a distributed DRL algorithm where each IoT device makes independent decisions based on its local system state is a promising solution as in \cite{Liu2019}. However, the mutual exclusion nature in multi-user resource allocation makes it hard to design a fully distributed solution in general. Moreover, as DRL algorithms normally take tens of thousands or even millions of time steps to train, the computation complexity and energy consumption are likely to forbid their implementation on the resource-constrained IoT devices. \par

%    For example,  A DRL algorithm for distributed implementation at the IoT devices based on local system state is designed in \cite{chen2018decentralized}. The offloading and resource allocation problem is reduced to a power allocation problem by considering a data-partition task model. 

\begin{figure}[!t]
	\centering
	\includegraphics[width=0.5\textwidth]{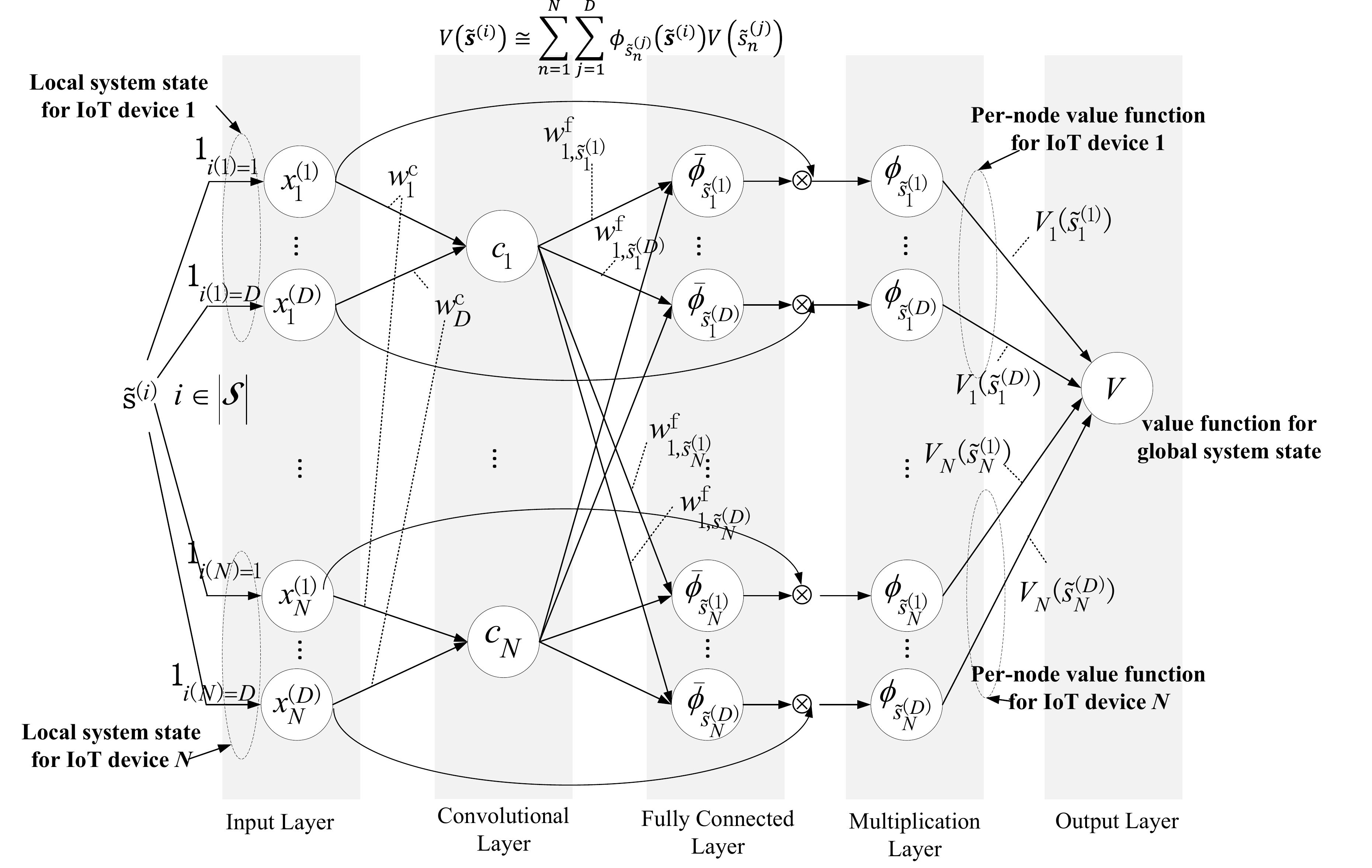}
	\caption{A value function approximation architecture of DRL Algorithm for IoT edge computing that facilitates semi-distributed implementation.}
	\label{edge4}
\end{figure}

In contrast to a fully centralized or distributed DRL algorithm, a semi-distributed implementation where the edge server and IoT devices cooperate to determine the optimal action is proposed in \cite{Lei2019}. A multi-user edge computing system with NB-IoT wireless network is considered, where the offloading decision and user scheduling are optimized to minimize delay and energy consumption. A value function approximation architecture as illustrated in Fig. \ref{edge4} for DRL algorithm is proposed. The global system state is first decomposed into local system states, which are used as input to the NN to approximate the value functions. The output to the NN are value functions for the global system states, which are derived from the NN as the sum of per-node value functions of local system states. Specifically, the value function for the $i$-th global system state $\mathbf{s}^{(i)}$ can be derived as
\begin{align}
\label{eq26}
V(\tilde{\mathbf{s}}^{(i)})\cong\sum_{n=1}^{N}\sum_{j=1}^{D}\phi_{\tilde{s}_{n}^{(j)}}(\tilde{\mathbf{s}}^{(i)})V_{n}(\tilde{s}_{n}^{(j)}),
\end{align}
where $D$ is the cardinality of the local system space of any device $n\in\{1,2,\cdots,N\}$, and $V_{n}(\tilde{s}_{n}^{(j)})$ is the per-node value function of IoT device $n$ for its local system state $\tilde{s}_{n}^{(j)}$. $\{\phi_{\tilde{s}_{n}^{(j)}}(\tilde{\mathbf{s}}^{(i)})\}_{n=1}^{N}$ is the feature vector of the global system state $\tilde{\mathbf{s}}^{(i)}$.\par

\begin{figure}[!t]
	\centering
	\includegraphics[width=0.5\textwidth]{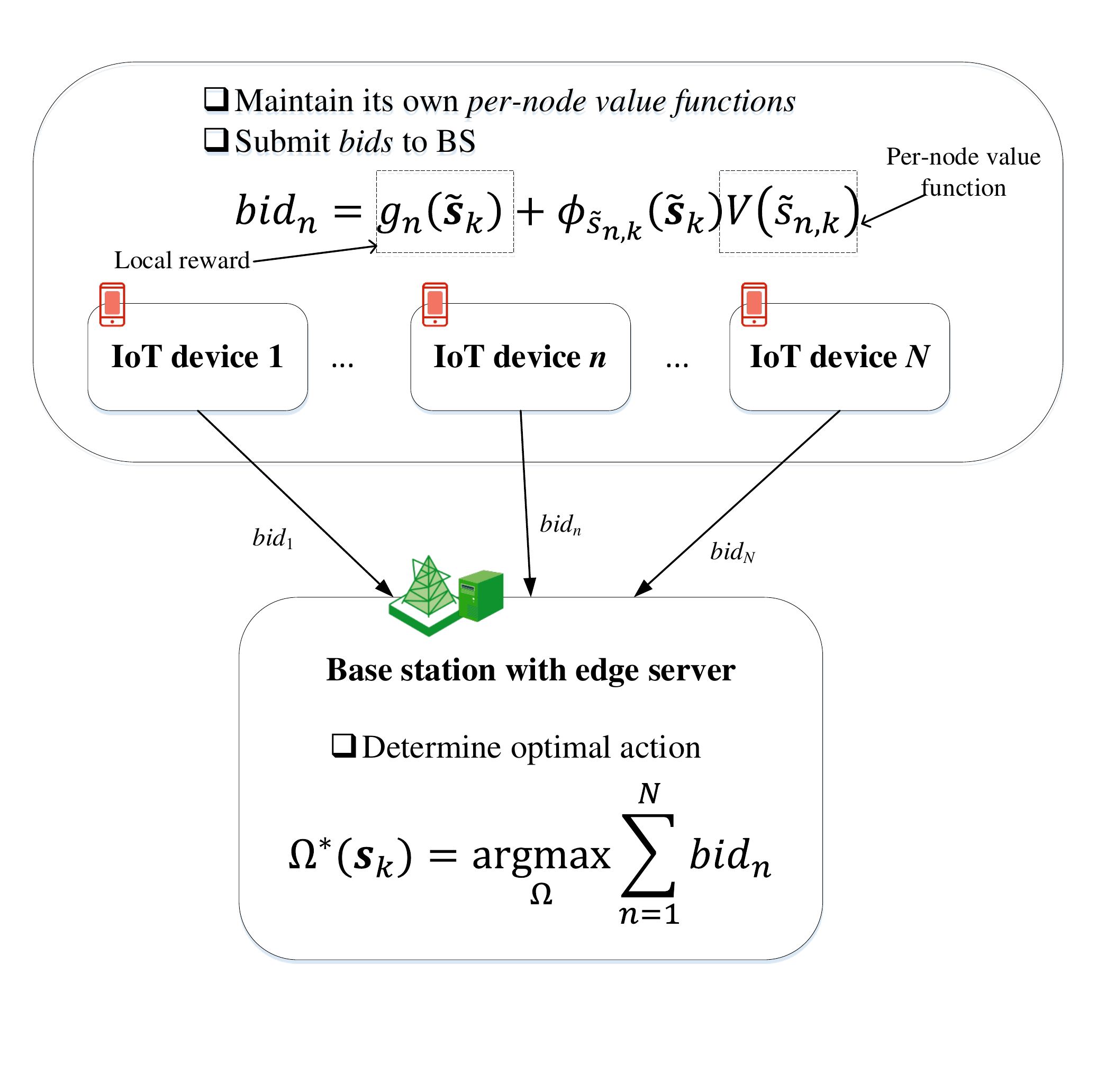}
	\caption{A semi-distributed implementation of DRL algorithm for IoT edge computing.}
	\label{edge3}
\end{figure}

The above proposed NN architecture can facilitate semi-distributed auction-based implementation as shown in Fig. \ref{edge3}. Specifically, each IoT device maintains its per-node value functions, based on which it can distributively calculate and submit bids to the BS and edge server. The BS centrally determines the optimal action based on the bids submitted by all the IoT devices. In this way, IoT devices help to alleviate the computational and storage burdens from the BS, while BS makes control decisions to control the scarce spectrum resources in the license band.\par

\subsubsection{Caching}
Caching decision mainly involves content placement, such as whether to cache content at the edge server, and which existing content should be replaced when new content is stored as listed in Table table\_edge2. The research in \cite{zhu2018caching} solves the problem of caching IoT data at the edge with the help of DRL. The proposed data caching policy aims to strike a balance between the communication cost and the loss of data freshness. In \cite{wei2018joint}, the issue of caching strategy is tackled together with the offloading policy and resource allocation. The natural actor-critic algorithm is adopted for this purpose.\par

\subsubsection{Comparison and Insights}
By summarizing and comparing the above literature, the following insights can be obtained.

\begin{itemize}
	\item \textit{System model}: Most of the research works focus on two-layer one-to-multiple and two-layer multiple-to-one system architectures. On the other hand, only a few literatures are devoted to two-layer multiple-to-multiple or three-layer architectures due to their complexities. Therefore, it is worthwhile to devote more attention to these system architectures in the future. 
	\item \textit{DRL model}: A general rule of thumb when defining the system state is that if the wireless channel fading is considered, the channel state usually needs to be included in the system state. On the other hand, if the dynamic task arrival model is considered, the queue state is usually necessary. Also note that when DRL is used to solve mixed integer programming problems under single-stage static task arrival models \cite{huang2019deep,Huang2019,wang2018smart}, the system states are usually defined as the control decisions to be optimized (e.g., offloading decision, resource allocation), which are usually formulated as actions in DRL models for dynamic and sequential-decision static task arrival models. It can be observed from Table \ref{table_edge2} that the mostly considered rewards are delay and energy consumption.
	\item \textit{DRL algorithm}: DQN, deep Q-Learning or similar value-based DRL methods are adopted most frequently as can be observed from Table \ref{table_edge2}. Actor-critic algorithms are adopted for those few works with continuous action space. Moreover, some works with discrete action spaces also adopt the actor-critic algorithms, as they promise better performance than their value-based counterparts. It should also be pointed out that most existing works usually directly apply the well-known DRL algorithms such as DQN and DDPG. However, there are also some research works that propose to improve the existing DRL algorithms considering the specific characteristics of studied problems. The authors in \cite{Qiu2019} use the adaptive genetic algorithm (ACA) for exploration. Moreover, recent works in \cite{Chen2019c,Lei2019} designed the Q or value function approximation architectures according to the structures of the specific problems. More research efforts can be devoted to this aspect in future works. 
	\item \textit{Implementation}: Most DRL algorithms are considered to be centrally implemented in the edge/fog/cloud server or central control unit, where the computation complexity and communication overhead increase with the increasing number of IoT devices. On the other hand, fully distributed implementation of DRL algorithms on the resource-constrained IoT devices may result in large processing delay and energy consumption. It is interesting to explore efficient collaboration between edge/fog/cloud server and IoT devices to achieve the most efficient implementation of the DRL algorithms.
	%        \item \textit{AIoT layer}: Most existing research focuses on the application layer, while a few literatures consider both the application layer and the communication layer. It will be interesting to include the perception layer in the DRL model as well, which means that the edge/fog/cloud computing system for some specific IoT applications are considered.

\end{itemize}

%\subsection{Others}
%In \cite{mohammadi2018semisupervised}, a semi-supervised DRL model is proposed, which fits smart city applications as it consumes both labeled and unlabeled data to improve the performance and accuracy of the learning agent. The model utilizes variational auto-encoders (VAE) as the inference engine for generalizing optimal policies. The authors state that the proposed model is the first investigation that extends DRL to the semi-supervised paradigm.
%In \cite{kolomvatsos2017reinforcement}, the authors define the concept of a query controller (QC) that receives queries for analytics and assigns each of them to a processor placed in front of each data partition. RL schemes are adopted to discuss an intelligent process for query assignments.\par
%Based on the above literature review, we summarize and list some typical values of states, actions, and rewards in Table \ref{table}, arranged in different categories as given in Section III corresponding to the three layers in AIoT architecture. \par

\subsection{AIoT Perception Layer - Autonomous Robots}

The applications of DRL methods in autonomous robots have been widely discussed. Researches include mobile behavior control of robots, robotic manipulation, management in multi-robot systems, and cloud robotics issues. Table \ref{table_robot} lists the related research works and their respective DRL models and DRL algorithms.

\begin{table*}[!t]	
	
	\newcommand{\tabincell}[2]{\begin{tabular}{@{}#1@{}}#2\end{tabular}}
	
	\centering
	
	\renewcommand{\arraystretch}{1.2}
	
	\caption{Summary of DRL Models and Algorithms in Research Works for IoT Autonomous Robots.}
	
	\label{table_robot}
	
	\begin{tabular}{|p{0.8cm}|p{0.5cm}|p{1.5cm}|p{2cm}|p{2.8cm}|p{1.5cm}|p{0.8cm}|p{1.2cm}|p{1.5cm}|} 
		
		\hline
		
		\textbf{Theme} & \textbf{Ref.} &\multicolumn{5}{c|}{\textbf{DRL Model}} & \textbf{DRL} & \textbf{Agent}\\
		
		\cline{3-7}
		
		&	& \textbf{State} & \textbf{Action} & \textbf{Reward} & \textbf{Environment} &\textbf {Feature} & \textbf {Algorithm} & \textbf {Location}\\
		
		\hline
		
		\multirow{4}*{\rotatebox[origin=c]{270}{Mobile Behavior Control}}& \cite{sasaki2017study}& surrounding environment & kinematic control& collision avoidance, task completion efficiency & perception layer &basic & DQN & robot (centralized) \\ \cline{2-9}
		
		& \cite{xin2017application,mIoT} & surrounding environment & kinematic control& task completion & perception layer &basic & DQN & robot (centralized) \\
		
		\cline{2-9}
		
		& \cite{yan2018path} & kinematic state & kinematic control & collision avoidance & perception layer &basic & DDPG & robot (centralized) \\
		
		\cline{2-9}
		
		& \cite{tongloy2017asynchronous} & surrounding environment, kinematic state& kinematic control & collision avoidance, task completion efficiency& perception layer &basic & A3C & robot (centralized) \\	
		
		\hline

		\multirow{4}*{\rotatebox[origin=c]{270}{ Robotic Manipulation}} & \cite{yang2018hierarchical}&   surrounding environment & kinematic control & collision avoidance, task completion efficiency & perception layer  & basic & DDPG & robot (centralized)\\
		
		\cline{2-9}
		
		& \cite{gu2017deep} & manipulation state, kinematic state & task manipulation action & task completion & perception layer & basic & DDPG & robot (centralized)\\
		
		\cline{2-9}
		
		& \cite{kalashnikov2018scalable} & surrounding environment, manipulation state & task manipulation action & task completion & perception layer & basic & deep Q-Learning& robot (centralized)\\
		
		\cline{2-9}
		
		& \cite{tsurumine2019deep} & surrounding environment & task manipulation action & task completion & perception layer & basic & deep P-network & robot (centralized)\\	
		
		\hline

		\multirow{5}*{\rotatebox[origin=c]{270}{ Multi-Robot System}} & \cite{yasuda2018collective} & surrounding environment & kinematic control & task completion & perception layer & basic & DQN & robot (centralized)\\
		
		\cline{2-9}
		
		& \cite{sun2009cooperative} & task completion state & task manipulation action & task completion efficiency & perception layer & MA & deep Q-Learning & robot (distributed)\\
		
		\cline{2-9}
		
		& \cite{sartoretti2019distributed} & surrounding environment & task manipulation action & task completion efficiency, task completion & perception layer & MA & actor-critic & robot (distributed)\\
		
		\cline{2-9}
		
		& \cite{long2018towards} & surrounding environment, kinematic state, targeted positions & kinematic control & collision avoidance, task completion efficiency & perception layer & POMDP, MA & policy gradient & robot (distributed)\\
		
		\cline{2-9}
		
		& \cite{mataric1997reinforcement} &kinematic state, manipulation state & task manipulation action, charging action & task completion & perception layer & MA & deep Q-Learning & robot (distributed)\\
		
		\hline

		\multirow{2}*{\rotatebox[origin=c]{270}{Cloud Robotics}} & \cite{liu2019lifelong}& content state & kinematic control, resource allocation & collision avoidance, task completion & perception layer, application layer & basic & DQN & cloud server (centralized)\\
		
		\cline{2-9}
		
		& \cite{liu2018reinforcement} & server state, task state & offloading decision, resource allocation & server utilization & perception layer, application layer & basic & Q-Learning & cloud server (centralized)\\
		
		\hline
		
	\end{tabular}
	
\end{table*}

\subsubsection{Mobile Behavior Control}

The mobile behavior control mainly refers to the path planning, navigation, and general movement control of robots. DRL approaches have been applied in many existing works for this purpose. In DRL models for mobile behavior control of robots, the actions mainly focus on the kinematic control of the autonomous robots, i.e., defining the moving velocity and direction in discrete or continuous form. The states are related to the surrounding environment and the kinematic state of robots. The surrounding environment can be obtained by sensors on the robot, for example, taking images of environment by camera, or measuring distance to obstacles by lasers. As for the reward, collision avoidance in robot movement is often considered, as collisions may cause damage of the robot and failure in the task of arriving the destination. Moreover, in some studies, the efficiency and effectiveness of robot movement are also taken into account. For example, in \cite{sasaki2017study}, a penalty is given to the robot when it moves backward. The authors apply DQN to the robot behavior learning simulation environment, so that mobile robots can learn to obtain good mobile behavior by using high-dimensional visual information as input data. Moreover, the authors incorporate profit sharing methods into DQN to speed up learning, and the method reuses the best target network in the case of a sudden drop in learning performance. In \cite{tongloy2017asynchronous}, the cumulative distance traveled is involved when defining the reward in the DRL model of solving a mobile robot navigation problem. A hybrid A3C method is applied with the aid of convolution neural network (CNN) and LSTM. Mobile robot path planning is a common problem in mobile behavior control of autonomous robot. DQN is designed in \cite{xin2017application} and DDPG is applied in \cite{yan2018path} to solve the path planning issues. In these cases, whether the robot reaches targeted destinations is an essential concern when defining the reward function. The study in \cite{mIoT} combines Q-Learning with CNN for IoT enabled mobile robot with an arm which can reach the destination autonomously and perform suitable actions.

\subsubsection{Robotic Manipulation}

Since intelligent robots usually help to perform some operation tasks in practice, appropriate controlling schemes for them are necessary for successful manipulations. DRL theories are widely applied in solving the robotic manipulation problems. For example, the problem of controlling robots to accomplish compound tasks is solved by a hierarchical DRL algorithm in \cite{yang2018hierarchical}. In \cite{gu2017deep}, the authors demonstrate that the DRL algorithm based on off-policy training of deep Q-functions can be applied to complex three-dimensional (3D) operation tasks, and can effectively learn DNN strategies to train real physical robots. The policy updates are pooled asynchronously to decrease the training time. Similarly, the problem of learning vision-based dynamic manipulation skills is solved by using a scalable DQN approach in \cite{kalashnikov2018scalable}. \par

	In the studies related to robotic manipulation control by DRL theories, the manipulation state is usually involved in the state of the DRL model, in addition to the kinematic state and the surrounding environment. The specific form of the manipulation state is determined by the type of the task for the robot. For example, the current status of the gripper on the robot, i.e. whether it is open and the height it has reached are a part of action in \cite{kalashnikov2018scalable}. The action in the model is also related to the specific manipulation task. For example, in \cite{tsurumine2019deep}, the action is defined as picking up the handkerchief in a real robotic cloth manipulation task. As for the reward in robotic manipulation problems, the success of target achievement is a necessary criteria.

\subsubsection{Multi-Robot System}

In some cases, multiple robots are required to collaborate properly to fulfil some tasks that are difficult to be accomplished by an individual robot. A review on MA RL in multi-robot systems is provided in \cite{yang2005survey}. In multi-robot system, the action is related to the kinematic control and manipulation control of multiple robots. The research in \cite{yasuda2018collective} investigates a DRL approach to the collective behavior acquisition of swarm robotics systems. The multiple robots are expected to collect information in parallel and share their experience for accelerating the learning. In \cite{mataric1997reinforcement}, the charging action is involved, where the robot is supposed to recharge in time. The task completion efficiency is an important evaluation criteria of effectiveness of the cooperation of the multiple robots. Thus, in multi-robot system, the task completion efficiency usually contributes to the reward in the DRL model. For autonomous robots, the task completion efficiency can be evaluated by overall completion time of tasks\cite{sun2009cooperative}. \par

\begin{figure}[!t]

	\centering

	\includegraphics[width=0.5\textwidth]{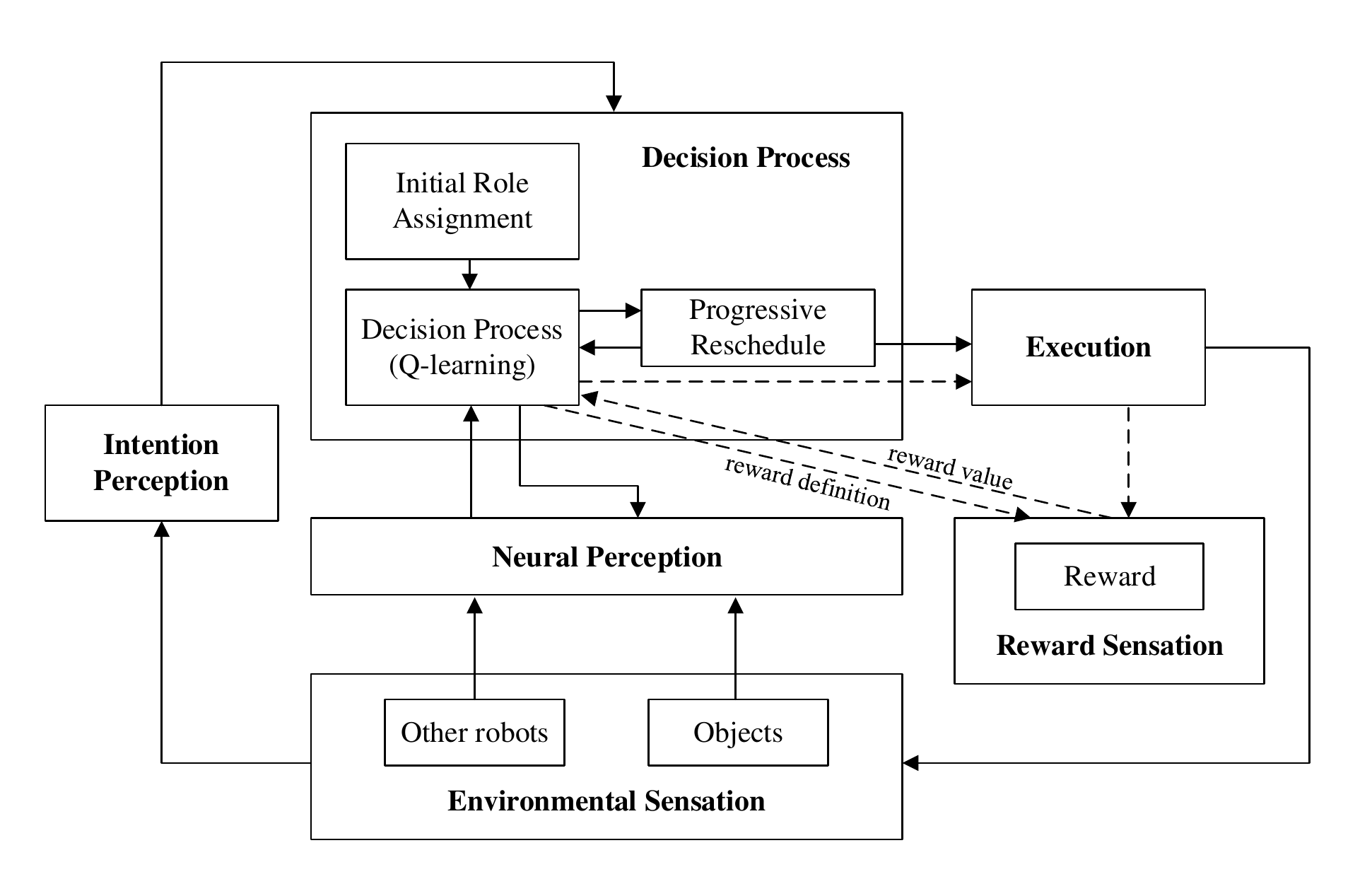}

	\caption{An MA framework for multiple robots to collaborate in a joint task space.}

	\label{robot-MA}

\end{figure}

	Distributed control schemes are introduced in some existing works. In \cite{sun2009cooperative}, the authors propose a collaborative multi-robot RL method, which realizes task learning and the emergence of heterogeneous roles under a unified framework. The method interleaves online execution and relearning to accommodate environmental uncertainty and improve performance. The multi-robot framework in Fig. \ref{robot-MA} provides an architecture for robots to collaborate in a joint task space with environmental uncertainties towards maximizing global team utility, where the components follow the environmental sensation, neural perception, decision, execution, reward feedback cycle. The study in \cite{sartoretti2019distributed} extends the A3C algorithm in single agent problems to a multi-robot scenario, where the robots work together toward a common goal. The policy and critic learning are centralized, while the policy execution is decentralized. A distributed sensor-level collision avoidance policy for multi-robot systems is proposed in \cite{long2018towards}. A multi-scenario multi-stage training framework based on policy gradient methods is used to learn the optimal policy for a large number of robots in a rich, complex environment. \par

\subsubsection{Cloud Robotics}

The concept of cloud robotics allows the robotic system to offload computing-intensive tasks from the robots to the cloud \cite{saha2018comprehensive}. Generally, cloud robotics applications include perception and computer vision applications, navigation, grasping or manipulation, manufacturing or service robotics, etc. Similar to discussions on the AIoT edge/fog/cloud computing system, the cloud robotics issues involve solving the offloading decision and resource allocation problems via DRL algorithms. For example, the authors in \cite{liu2018reinforcement} propose an RL-based resource allocation scheme, which can help the cloud to decide whether a request should be accepted and how many resources are supposed to be allocated. The scheme realizes an autonomous management of computing resources through online learning, reduces human participation in scheme planning, and improves the overall utility. \par

	Moreover, the resource allocation and kinematic control issues can be considered together in cloud robotics. For example, in \cite{liu2019lifelong}, an effective transfer learning scheme based on lifelong federated RL (LFRL) is proposed for the navigation in cloud robotic systems, where the robots can effectively use prior knowledge and quickly adapt to new environments of the system in the long run.

	\subsubsection{Comparison and Insights}

	By summarizing and comparing across the above literature, the following insights can be obtained.

	\begin{itemize}

		\item \textit{System model}: The autonomous robot problems include models with single robot or multiple robots. The multi-robot problem has been widely studied, as it is a typical issue in MA RL. 

		\item \textit{DRL model}: The state in DRL models for autonomous robot usually involves perceiving the environment of robots. The camera image is mostly adopted to represent the surrounding environment. No matter in a single robot or a system of multiple robots, if the problem involves the motion of the robot, then the kinematic state is generally a part of the state. When robot manipulation is studied, the manipulation state is an important part of the state. When intelligent robots are combined with cloud computing techniques, that is, when the cloud robotics problem is studied, the state in the DRL model includes the server computing resource state and the task execution state at cloud. The action in the DRL model for autonomous robot is mainly related to what the robot does. For example, when the main job of a robot is to move, the action in the model is related to kinematic control. For problems related to robot manipulation, the task manipulation action is considered in the DRL model. For example, when a robot wants to use its mechanical arm to pick up or drop down some objects, the possible angle, height, and switch state of the arm constitute the action set. In cloud robotics, the actions in the model may involve offloading decision and resource allocation. As for the reward in the DRL model, it is often related to whether the autonomous robot can successfully complete the task.

		\item \textit{DRL algorithm}:  DQN and DDPG algorithms are widely applied in autonomous robots. For continuous action space, DDPG is mostly adopted \cite{yan2018path, yang2018hierarchical}. In the research of autonomous robots, the multi-robot problem involves MA DRL theory. In this case, MA DRL algorithms are adopted. For example, in \cite{long2018towards}, a policy gradient based RL algorithm is applied among multiple robots. In the system, each robot receives its own observation at each time step and executes the action generated from the shared policy. The policy is trained with experiences collected by all robots simultaneously, which allows the multiple robots to cooperate well.

		\item \textit{Implementation}: For cloud robotics, DRL models are implemented on the cloud servers. The servers make decisions for the task execution in the robot system. On the other hand, DRL algorithms can also be implemented on the robots, which can be considered as a type of powerful IoT devices. For example, the implementation of the algorithm in \cite{long2018towards} involves a large-scale robot group with laser scanners. The multi-robot collision avoidance policy among them is trained on each robot with a computer of an i7-7700 CPU and a Nvidia GTX 1080 GPU.

		%\item \textit{AIoT layer}: In autonomous robots, most existing research focus on perception layer. In cloud robotics, both the perception layer and the application layer are involved. In future works, researchers can consider introducing network layer in the study. For example, in the multi-robot problem, communication issues between robots can be involved in the DRL model, where the communications enable better cooperation between robots.

	\end{itemize}

\subsection{AIoT Perception Layer - Smart Vehicles}

\begin{table*}[!t]

	\newcommand{\tabincell}[2]{\begin{tabular}{@{}#1@{}}#2\end{tabular}}

	\centering

	\renewcommand{\arraystretch}{1.2}

	\caption{Summary of DRL Models and Algorithms in Research Works for IoT Smart Vehicles.}

	\label{table_vehicle2}

	%	\resizebox{\textwidth}{!}{

	\begin{tabular}{|p{0.8cm}|p{0.5cm}|p{1.8cm}|p{2cm}|p{2.5cm}|p{1.7cm}|p{0.8cm}|p{1.2cm}|p{1.5cm}|} 

		\hline

		\textbf{Theme} & \textbf{Ref.} &\multicolumn{5}{c|}{\textbf{DRL Model}} & \textbf{DRL} & \textbf{Agent}\\

		\cline{3-7}

		&	& \textbf{State} & \textbf{Action} & \textbf{Reward} & \textbf{Environment} &\textbf {Feature} & \textbf {Algorithm} & \textbf {Location}\\

		\hline

		\multirow{10}*{\rotatebox[origin=c]{270}{Autonomous Driving}}& \cite{yu2016deep} & driving environment & velocity control, direction control& driving safety, driving smoothness, driving efficiency & perception layer &basic & deep Q-Learning & on board (centralized) \\ \cline{2-9}

		& \cite{vitelli2016carma} & driving environment, kinematic state & velocity control, direction control  & driving efficiency & perception layer &basic & deep Q-Learning & on board (centralized) \\

		\cline{2-9}

		& \cite{mirchevska2018high} & driving environment, kinematic state & direction control & driving efficiency & perception layer  &basic & deep Q-Learning & on board (centralized) \\

		\cline{2-9}

		& \cite{wu2017flow} & driving environment, kinematic state& velocity control, direction control & driving smoothness, driving efficiency& perception layer &basic & policy gradient & road-side unit (centralized) \\

		\cline{2-9}

		& \cite{gamage2017reinforcement} & driving environment, kinematic state & velocity control, direction control &  environmental benefits & perception layer & basic & deep Q-Learning & road-side unit (centralized) \\

		\cline{2-9}

		&\cite{you2018highway} & driving environment&velocity control, direction control& driving safty& perception layer& basic & deep Q-Learning & on board (centralized)  \\

		\cline{2-9}

		& \cite{pal2018reinforcement}& driving environment, kinematic state& velocity control, direction control & driving efficiency & perception layer, network layer & basic & deep Q-Learning & on board (centralized)  \\

		\cline{2-9}

		& \cite{wang2017formulation}& driving environment& velocity control, direction control& driving smoothness, driving efficiency& perception layer & basic & actor-critic & on board (centralized)  \\

		\cline{2-9}

		& \cite{wang2013cooperative}& driving environment, kinematic state& velocity control, direction control& driving safety& perception layer  & basic & deep Q-Learning & on board (centralized)  \\

		\cline{2-9}

		& \cite{khamis2014adaptive}& driving environment & velocity control, direction control & driving safety, driving efficiency, environmental benefits  & perception layer & MA & coordinated MA DQN& road-side unit (distributed) \\

		\hline

		\multirow{4}*{\rotatebox[origin=c]{270}{Vehicular Network}} & \cite{ye2018deep}&   channel state, vehicle state, performance requirements & transmission control & reliability, transmission efficiency& network layer  & basic & deep Q-Learning & BS (centralized)  \\

		\cline{2-9}

		& \cite{challita2019interference}&channel state, vehicle state & transmission control, velocity control, direction control& reliability, transmission efficiency, driving efficiency&perception layer, network layer & basic & deep echo state network & on board (centralized)  \\

		\cline{2-9}

		& \cite{he2018integrated} & resource state &transmission control & reliability, resource utilization &network layer, appication layer& basic & DDQN & edge server (centralized)  \\

		\cline{2-9}

		&\cite{atallah2018scheduling}&channel state, vehicle state, resource state&transmission control&reliability, driving efficiency, environmental benefits&network layer& basic & DQN &on board (centralized)  \\

		\hline

		\multirow{3}*{\rotatebox[origin=c]{270}{Vehicular Edge/Fog/Cloud Computing}} & \cite{qi2019knowledge,qi2018vehicular}&kinematic state, resource state& offloading decision& task handling efficiency& application layer& basic &A3C & edge server (centralized)  \\

		\cline{2-9}

		&\cite{zheng2015smdp}&resource state, request state& offloading decision& task handling efficiency, environmental benefits&application layer& basic & deep Q-Learning & cloud server (centralized)  \\

		\cline{2-9}

		&\cite{ydeep}&resource state& offloading decision, resource allocation& reliability, task handling efficiency&network layer, application layer& basic & deep Q-Learning & cloud server (centralized)  \\

		\cline{2-9}

		&\cite{hou2018q}&request state&offloading decision, caching strategy&task handling efficiency&  application layer & basic & deep Q-Learning & edge server (centralized)  \\			

		\hline

	\end{tabular}

	%	}

\end{table*}

The development of IoT technology has promoted the development of intelligent transportation systems (ITS). In the Internet of Vehicles (IoV), smart vehicles with IoT capabilities including sensing, communications, and data processing can possess artificial intelligence to enhance driving aid. The existing works on the applications of DRL in smart vehicles mainly include the studies on autonomous driving, vehicular networks, and vehicular edge/fog computing (VEC/VFC). Table \ref{table_vehicle2} lists the related research works and their respective DRL models and DRL algorithms.\par

\subsubsection{Autonomous Driving}

The application of DRL methods for the control of autonomous vehicles is addressed in many existing works. The authors in \cite{talpaert2019exploring} review the applications and address the challenges of real-world deployment of DRL in autonomous driving. The existing works mainly discuss the autonomous driving problem of a single vehicle or multiple vehicles. Generally, the autonomous driving problem is formulated as an MDP. In the DRL model, the state may include both the kinematic state and driving environment of vehicles. Generally, the kinematic state refers to position, velocity and other kinematic features of the vehicle whose driving behavior is required to be determined in the model. The driving environment is mainly related to other vehicles nearby, as well as the traffic facilities and obstacles on the road. The information of the driving environment can be obtained by sensing, i.e., using sensors such as radar and camera, or by communicating, i.e., communication in IoV such as vehicle-to-vehicle (V2V) and vehicle-to-Internet (V2I). Moreover, the driving environment can be in the form of an image, i.e. take an image of the surrounding traffic by the on-board camera, or be characterized in the form of vectors, i.e., use variables to represent the state of traffic signals and positions of obstacles. The velocity control and direction control are usually characterized as actions. The rewards are mostly related to assessment criteria of the driving operations, such as driving safety, driving smoothness and driving efficiency. \par

	In the problem of a single vehicle, the driving behavior of one autonomous vehicle is studied. Since the driving environment may affect the driving behavior of the vehicle, the camera image of the environment is considered as the state in the DRL model in \cite{yu2016deep}, and the deep Q-Learning method is applied to control simulated cars. In \cite{vitelli2016carma, mirchevska2018high, pal2018reinforcement}, the proposed methods are also based on deep Q-Learning, the DRL models become more complex, since the driving environment as well as the kinematic state of the vehicle are taken into account. \par

	The actions of the single vehicle include the velocity control and direction control. The velocity control can be in a discrete form as a decision of acceleration or deceleration, or in a continuous form as a decision of a specific value of velocity. The direction control can be in a discrete form, such as lane-changing, turning left or right, or maintaining in the current direction. The direction control can also be in a continuous form, where the specific steering angle is required to be obtained. In most existing works, the velocity or direction control action is in discrete form, where rough decisions on velocity and direction adjustment are made \cite{vitelli2016carma, you2018highway, pal2018reinforcement, khamis2014adaptive}. For examples, in \cite{you2018highway}, road geometry is taken into account in the MDP model in order to be applicable for more diverse driving styles, and discrete velocity and direction control are involved. The study in \cite{pal2018reinforcement} aims to optimize the driving utility of the autonomous vehicle, and enables the autonomous vehicle to jointly select the discrete motion planning action performed on the road and the communication action of querying the sensed information from the infrastructure. In some existing works, the velocity or direction control is in continuous form. For instance, the problem of ramp merging in autonomous driving is tackled in \cite{wang2017formulation}, and the specific velocity and steering angle of the vehicle is decided. \par

	As for the reward in DRL model for autonomous driving of a single vehicle, driving safety is usually concerned, and the reward function is related to the collision avoidance performance of the vehicle. For example, in \cite{mirchevska2018high}, the authors address the autonomous driving issues by presenting an RL-based approach, which is combined with formal safety verification to ensure that only safe actions are chosen at any time. A DRL agent learns to drive as close as possible to the desired velocity by executing reasonable lane changes on simulated highways with an arbitrary number of lanes. Besides, the driving efficiency and driving smoothness are also discussed in some existing works. For examples, the authors in \cite{wu2017flow} use Flow to develop reliable controllers in mixed-autonomy traffic scenarios, and the reward function is related to the average velocity of vehicles, with an added penalty to discourage accelerations or excessive lane-changes. The authors in \cite{kendall2018learning} apply a continuous, model-free DRL algorithm for autonomous driving. The distance traveled by the autonomous vehicle is used to evaluate the reward in the model. Moreover, some researchers also consider the environmental benefits in autonomous driving, and vehicle fuel consumption or power consumption is taken into account. \par

	There are also studies on the cooperative driving of multiple vehicles. The types of state, action and reward in the DRL model are similar to those in the single-vehicle cases. However, in multi-vehicle cases, the DRL algorithms may involve the MA methods, and the implementation of the system may be in distributed form. In \cite{wang2013cooperative}, the authors present a novel method of cooperative movement planning. RL is applied to solve this decision-making task of how two cars coordinate their movements to avoid collisions and then return to their intended path. An MA multi-objective RL traffic signal control framework is proposed in \cite{khamis2014adaptive}, which simulates the driver's behavior, e.g., acceleration or deceleration, continuously in space and time dimensions. \par

	When evaluating the performance of proposed DRL-based methods for autonomous driving, the researchers mostly test their methods in simulated traffic environment. A dataset can be obtained by using simulators such as JavaScript Racer \cite{yu2016deep}, Vdrift Ruddskogen track \cite{vitelli2016carma}, and Auto Drive Simulator \cite{pal2018reinforcement}, etc. Different from testing in a simulated environment, some works use real traffic dataset in their experiments, which can better reflect the real traffic environment \cite{xu2017end,mirchevska2018high}. \par	

\begin{figure}[b]

	\centering

	\includegraphics[width=0.45\textwidth]{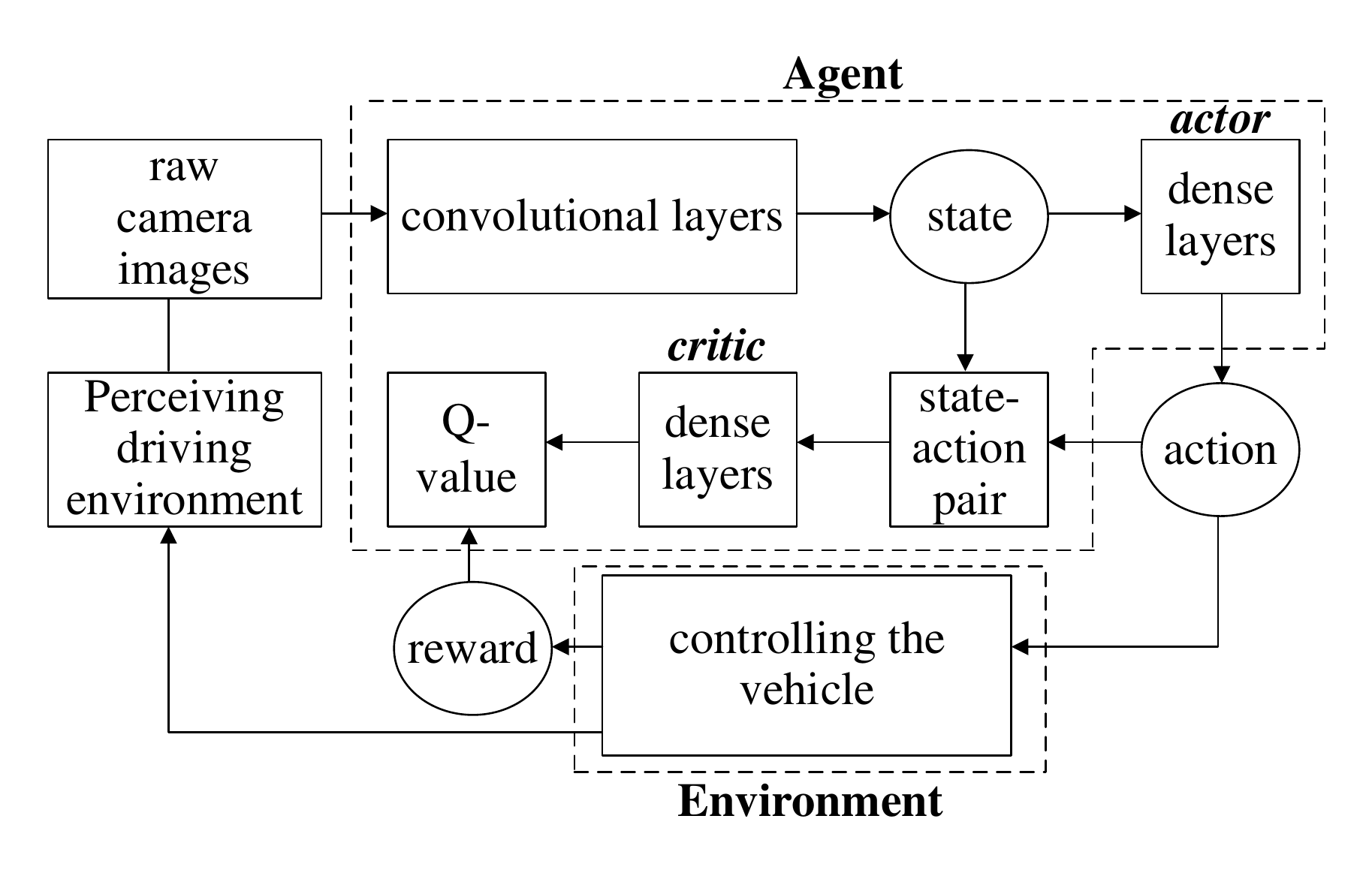}	

	\caption{Actor-critic method for vehicle control problems in autonomous driving.}

	\label{DRL-AD-AC}

\end{figure}

	Fig.\ref{DRL-AD-AC} shows an example of applying DRL algorithms in an autonomous driving problem, where actor-critic algorithm is adopted. The driving environment and kinematic state is characterized as the state vector in the DRL model. An action is selected according to the policy, and the Q-value of each possible state-action pair is estimated. The Q-values act as the critic in the model, and give a guideline for action selections. The selected action directly determines the controlling of the smart vehicle. After executing the control command, the vehicle obtains a reward which can influence the critic in the model.

	%When solving autonomous driving problems by DRL theory, a general procedure is shown in Fig.\ref{DRL-AD}. The targeted vehicle refers to the autonomous vehicle whose driving behavior is required to be decided by DRL algorithms. Other related vehicles are named surrounding vehicles. First, the targeted vehicle needs to perceive the driving environment. Possible approaches include sensing (by radar or camera) and IoV communication. After that, the state of vehicles is defined. Generally, the state information involves the targeted vehicles as well as the surrounding vehicles, in order to make sure that both the driving environment and the kinematic state are taken into consideration. Then, the action is defined in discrete of continuous form, representing the driving behavior of the targeted vehicle. The reward function is related to the driving criteria and objective such as driving safety, driving efficiency. Finally, the DRL algorithms can be applied to solve the established model, and the optimal policy can be obtained. According to the optimal policy, the targeted vehicle can decide its driving behavior in the dynamic driving environment.

\subsubsection{Vehicular Networks}

The concept of vehicular networking brings a new level of connectivity to vehicles, and has become a key driver of ITS. The control functionalities in the vehicular network can be divided into three parts according to their usages, including communication control, computing control and storage control \cite{zheng2016soft}. In \cite{liang2019toward} and \cite{ye2017machine}, the applications of ML in studying the dynamics of vehicular networks and making informed decisions to optimize network performance are discussed. In vehicular networks, problems such as resource allocation, caching, and networking can be formulated and solved via DRL. \par

	In the existing works, the states include the transmission channel status, vehicle status, resource status, and performance requirements. The actions mainly focus on message transmission control in the network. The reliability and transmission efficiency are mostly concerned when defining the reward function. For example, in \cite{challita2019interference}, a DRL algorithm based on echo state network (ESN) cells is proposed in order to provide an interference-aware path planning scheme for a network of cellular-connected UAVs. The state focuses on the external input of UAV and state of the reservoir of UAV. In \cite{ye2018deep}, the performance requirement, i.e., the delay constraint, is further considered when defining the state in the DRL model. The authors use a DRL approach to perform joint resource allocation and scheduling in V2V broadcast communications. In the system, each vehicle makes a decision based on its local observations without the need of waiting for global information. The resource state, i.e., the available BS and the available cache, is considered in \cite{he2018integrated}, where the authors develop an integration framework that enables dynamic orchestration of networking, caching, and computing resources to improve the performance of vehicular networks. The resource allocation strategy is formulated as a joint optimization problem, in which the gains of networking, caching and computing are all taken into consideration. To solve the problem, a double-dueling-deep Q-network algorithm is proposed. Similarly, deep Q-Learning is applied in \cite{atallah2018scheduling} to learn a scheduling policy, which can guarantee both safety and quality-of-service (QoS) concern in an efficient vehicular network. \par

\subsubsection{Vehicular Edge/Fog/Cloud Computing} 

Emerging vehicular applications require more computing and communication capabilities to perform well in computing-intensive and latency-sensitive tasks. Vehicular Cloud Computing (VCC) provides a new paradigm in which vehicles interact and collaborate to sense the environment, process the data, propagate the results and more generally share resources \cite{inbook}. Moreover, VEC/VFC focuses on moving computing resources to the edge of the network to resolve latency constraints and reduce cloud ingress traffic \cite{xiao2017vehicular,nobre2019vehicular,ning2019vehicular}. \par

	As studied in \cite{qi2019knowledge} and \cite{qi2018vehicular}, the vehicular edge, fog or cloud computing problems focus on the service offloading issues in the IoV. The state in the DRL model focuses on the resource and request status in the system. The determination of offloading decisions for the multiple tasks is considered as a long-term planning problem. In the existing works, service offloading decision frameworks are proposed, which can provide the optimal policy via DRL. For examples, the authors in \cite{zheng2015smdp} propose an optimal computing resource allocation scheme to maximize the total long-term expected return of the VCC system.

	With multiple access edge computing techniques, roadside units (RSUs) can provide fast caching services to moving vehicles for content providers. In \cite{hou2018q}, the authors apply the MDP to model the caching strategy, and propose a heuristic Q-Learning solution together with vehicle movement predictions based on an LSTM network.

	\subsubsection{Comparison and Insights}

	By summarizing and comparing the above literature, the following insights can be obtained.

	\begin{itemize}

		\item \textit{System model}: 

		The DRL problems for smart vehicles may be designed for a single vehicle or multiple vehicles. In autonomous driving, most existing works are single-vehicle problems, which can be solved by basic DRL algorithms such as deep Q-Learning and policy gradient methods. Some problems are multi-vehicle problems, where MA DRL algorithms are needed. In the vehicular network and vehicular edge/fog/cloud computing problems, the system usually involves multiple vehicles.

		\item \textit{DRL model}: 

		The DRL model states, actions and rewards of related works are summarized in Table \ref{table_vehicle2}. When defining the state in DRL models in autonomous driving problems, the kinematic state of the single or multiple vehicle(s) as well as the driving environment can be involved. The kinematic state is related to the kinematic features of the smart vehicle(s), and is usually be defined in the form of vectors or matrices. The driving environment refers to the traffic surrounding the smart vehicle(s), and can be characterized as images or vectors. The actions can be in discrete or continuous form, with velocity and moving direction concerned. The typical rewards in existing works are based on some driving criteria, including safety, smoothness, efficiency and environmental benefits. For the sake of feasibility and simplicity, few works consider these criteria at the same time. In future works, researchers can make efforts on an overall consideration of these criteria, and design reasonable priorities among them. 

		\item \textit{DRL algorithm}: 

		As shown in Table \ref{table_vehicle2}, when the state and action are in discrete form, deep Q-Learning is mostly adopted. When the state or action space is continuous, the actor-critic method, e.g., A3C or DQN is applied \cite{wang2017formulation,khamis2014adaptive,qi2019knowledge,qi2018vehicular}.

		\item \textit{Implementation}:

		A powerful IoT device like a smart vehicle can act as an agent in the DRL model. The vehicle-mounted servers are able to perform DRL algorithms on board. For example, in \cite{vitelli2016carma}, the authors trained their agent using CUDA-based GPU acceleration on commodity hardware and run the experiments on a Macbook Pro with an Nvidia GeForce GT 750M GPU. The hardware is relatively modest by most standards, and this indicates that such a number of computational resources is enough for implementing DQNs. 

		%\item \textit{AIoT layer}:In most existing works, the DRL problems for autonomous driving focus on the perception layer of AIoT. Sensors are most widely used for perceiving the environment. For example, the cameras on the smart vehicles are used for taking pictures of the traffic environment, and the on-board radars are used to sense the nearby vehicles or obstacles. Based on the information obtained by sensors, the agents make decisions on the controlling of the smart vehicles. On the other hand, communications in IoV enable information exchange between smart vehicles. The communications can help the smart vehicles better perceive the traffic environment, but may also introduce extra delay and affect the reliability of decision making in vehicle control. In other words, the performance of the communication network may influence the effectiveness of message transmission and thereby influence the vehicle control. In current vehicular networking problems, most works are only related to the network layer and focus on optimizing network performance such as delay and reliability. In the future work, researchers can make effort on an effective combination of autonomous driving and IoV issues, where the communication resource control and vehicle control are jointly optimized, where the reward can be determined by the driving performance of the controlled vehicle.

	\end{itemize}

\subsection{AIoT Perception Layer - Smart Grid}
\begin{figure}[!t]
	\centering
	\includegraphics[width=0.45\textwidth]{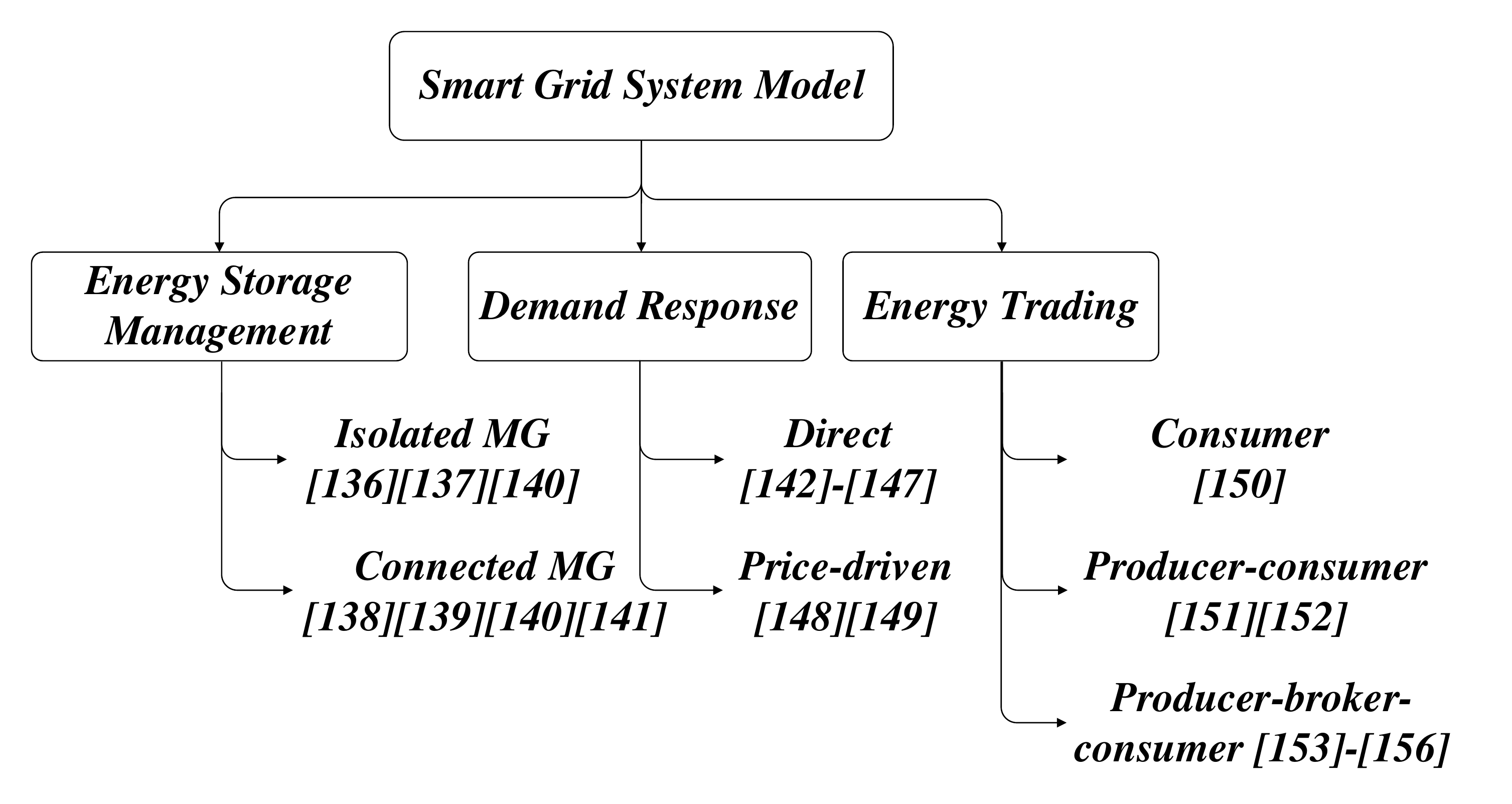}
	\caption{Classification of smart grid system models.}
	\label{sg1}
\end{figure}

The integration of distributed renewable energy sources (DRES) into the power grid introduces the need for autonomous and smart energy management capabilities in the smart grid due to the intermittent and stochastic nature of renewable energy sources (RES). With advanced metering infrastructure (AMI) and various types of sensors in the power grid to collect real-time power generation and demand data, RL and DRL provide promising methods to learn efficient energy management policies autonomously in such a complex environment with uncertainty. Specifically, the historical data can be leveraged by powerful DRL algorithms in learning optimal decisions to cope with the high uncertainty of the electrical patterns. The existing works mainly include studies on energy storage management (ESM), demand response (DR) and energy trading. We classify the existing research based on the different considerations in system models as shown in Fig. \ref{sg1}. Table \ref{table_grid2} lists the related research works and their DRL models and DRL algorithms. \par

\begin{table*}[t]
	\centering
	\renewcommand{\arraystretch}{1.2}
	\caption{Summary of DRL Models and Algorithms in Research Works for Smart Grid.}
	\label{table_grid2}
	%    \resizebox{\textwidth}{!}{
	\begin{tabular}{|p{0.8cm}|c|p{2.8cm}|p{2.6cm}|p{2.2cm}|p{0.8cm}|p{1.9cm}|p{2.3cm}|} 
		\hline
		\textbf{Theme} & \textbf{Ref.} &\multicolumn{4}{c|}{\textbf{DRL Model}} & \textbf{DRL} & \textbf{Agent }\\
		\cline{3-6}
		&    & \textbf{State} & \textbf{Action} & \textbf{Reward} & \textbf{Feature} &\textbf{Algorithm} &  \textbf{Location} \\
		\hline
		\multirow{6}*{\rotatebox[origin=c]{270}{Energy Storage Management}}& \cite{franccois2016deep} & battery SoC & ESS management & energy balance & POMDP&  DQN & EMS (centralized) \\ \cline{2-8}
		&\cite{qiu2016heterogeneous}& battery SoC & ESS management& ESS operation cost& basic & Q-Learning & EMS (centralized) \\
		\cline{2-8}
		& \cite{mbuwir2017reinforcement} & time state, RE generation state, battery SoC, energy demand state & ESS management & energy trading cost & basic & Q-Learning & EMS (centralized) \\
		\cline{2-8}
		&  \cite{zeng2018dynamic}& Energy demand state, RE generation state, price state, battery SoC & DG energy dispatch, ESS management, energy trading amount& MG operational cost & basic & ADP and RNN & EMS (centralized) \\
		\cline{2-8}
		& \cite{venayagamoorthy2016dynamic} & energy demand state, battery SoC, RE generation state & DG energy dispatch, ESS management, energy trading amount &  energy balance; & basic & ADP and actor-critic & EMS (centralized)  \\
		\cline{2-8}
		& \cite{ji2019real} & energy demand state, price state, battery SoC & DG energy dispatch, ESS management, energy trading amount& ESS operational cost & basic & DQN & MG controller (centralized) \\
		\hline
		\multirow{7}*{\rotatebox[origin=c]{270}{Demand Response}}&  \cite{mocanu2018line} & price state, energy demand state & DR devices on/off change & load shedding cost & basic &  DQN and DPG & EMS (centralized) \\ \cline{2-8}
		& \cite{wen2015optimal}& time state, energy demand state, price state &DR devices on/off change&  load shedding cost, consumer's satisfaction& basic & Q-Learning & EMS (centralized) \\
		\cline{2-8}
		&  \cite{o2010residential}& price state, energy demand state & DR devices on/off change &  load shedding cost, consumer's satisfaction& basic & Q-Learning with fixed step size & EMS (centralized) \\
		\cline{2-8}
		& \cite{claessens2016convolutional} & time state, DR device on/off state &DR devices on/off change &  load shedding cost& POMDP & DQN with CNN architecture & broker (centralized) \\
		\cline{2-8}
		& \cite{ruelens2014demand} & battery SoC &DR devices on/off change &   load shedding cost & basic & fitted Q-iteration & device controller (centralized)  \\
		\cline{2-8}
		& \cite{ruelens2016reinforcement} & time state, DR device on/off state & DR devices on/off change & load shedding cost & basic & fitted Q-iteration & device controller (centralized) \\
		\cline{2-8}
		& \cite{lu2018dynamic} & energy demand state, price state &energy trading price &   load shedding cost, energy trading cost & basic & Q-Learning & broker (centralized)  \\
		\cline{2-8}
		& \cite{lu2019incentive} & energy demand state, price state & energy trading price&   load shedding cost, energy trading cost& basic & Q-Learning & broker (centralized)  \\
		\hline
		\multirow{7}*{\rotatebox[origin=c]{270}{Energy Trading}}&  \cite{lim2014strategic} & price state & energy trading price & energy trading profit & MA &  Q-Learning & consumer (distributed) \\ \cline{2-8}
		& \cite{xiao2017energy}& energy demand state, RE generation state, price state, battery SoC &energy trading amount& energy trading profit & basic & Q-Learning & MG (centralized) \\
		\cline{2-8}
		& \cite{xiao2018reinforcement} & energy demand state, battery SoC & energy trading amount & energy trading profit& POMDP & DQN+CNN& MG (centralized) \\
		\cline{2-8}
		& \cite{kim2014dynamic}& time state, energy demand state, price state & energy trading price & energy trading profit& basic & Q-Learning with virtual experience & broker (centralized) \\
		\cline{2-8}
		& \cite{kim2016dynamic} & Energy demand state, price state & energy trading price & energy trading profit & MA & Q-Learning with virtual experience & broker and consumer (distributed) \\
		\cline{2-8}
		& \cite{reddy2011strategy} & energy demand state, price state & energy trading price & energy trading profit & basic & Q-Learning & broker (centralized) \\
		\cline{2-8}
		& \cite{foruzan2018reinforcement} & energy demand state, battery SoC, price state, time state & energy trading amount and price&  energy trading profit & MA & Q-Learning & producer and consumer (distributed) \\
		\hline
	\end{tabular}
	%    }
\end{table*}

\subsubsection{Energy Storage Management}
Microgrids (MGs) are small-scale, self-supporting power networks driven by on-site generation sources. They normally integrate RESs such as solar and wind, as well as energy storage elements such as the electrochemical battery. An MG can either connect or disconnect from the external main grid to operate in a grid-connected or isolated-mode. The most essential task in energy management of MG is to fully exploit the renewable energy power, while satisfying the critical requirement that the power generation and consumption are balanced. However, the main challenge lies in the lack of knowledge of future electricity generation and consumption. One promising method to deal with this challenge is through energy storage. Direct energy storage such as the battery is one of the energy storage options. However, the battery also brings new challenges in energy management. As the maximum amount of energy that can be charged or discharged at a certain point of time is limited by the energy storage capability and the current state-of-charge (SoC) of the battery, while the current SoC is in turn determined by the previous charging/discharging behavior, energy management becomes a sequential decision problem for a dynamic system where earlier decisions influence future available choices. \par

	The techniques using RL/DRL to determine the optimal charging/discharging policy for ESS in MG have been studied in some recent literature \cite{franccois2016deep,qiu2016heterogeneous,mbuwir2017reinforcement}. Specifically, \cite{franccois2016deep} and \cite{qiu2016heterogeneous} model the system as an isolated MG and only consider battery SoC as the state of the DRL model. The reward in \cite{franccois2016deep} uses the energy balance within the MG, taking into account the energy demand and the battery SoC level. In \cite{qiu2016heterogeneous}, on the other hand, the reward is defined by the ESS operation cost, which can be represented by the battery loss based on charge/discharge operation. Grid-connected mode MG is considered in \cite{mbuwir2017reinforcement} where MG is connected to the main grid. The state space consists of time state, renewable energy generation state, battery SoC and energy demand state. The reward is defined as the negative of the energy transaction cost.\par
	
	In \cite{zeng2018dynamic} \cite{venayagamoorthy2016dynamic} \cite{ji2019real}, except for ESS management, diesel generators (DGs) energy dispatch and energy trading are taken into account as part of the action space. There are two types of energy trading actions. One is the amount of energy that is traded with the main grid or other MGs, and the other refers to the price for energy transaction. In \cite{zeng2018dynamic},  the reward function is the negative of the MG operation cost which consists of DG cost, load shedding cost, energy transaction cost and ancillary services cost. To derive the optimal policy, approximate dynamic programming (ADP) and RNN learning are employed to solve the finite-horizon MDP problem. In \cite{venayagamoorthy2016dynamic}, the system is modeled in both isolated and connected modes and the reward functions represented by energy balance for both modes are given respectively. This work combines ADP and actor-critic framework to solve the dynamic optimization problem in EMS.\par

\begin{figure}[!t]
	\centering
	\subfigure[A CNN architecture for POMDP-based ESM problem.]{
		\begin{minipage}[b]{0.3\textwidth}
			\includegraphics[width=1\textwidth]{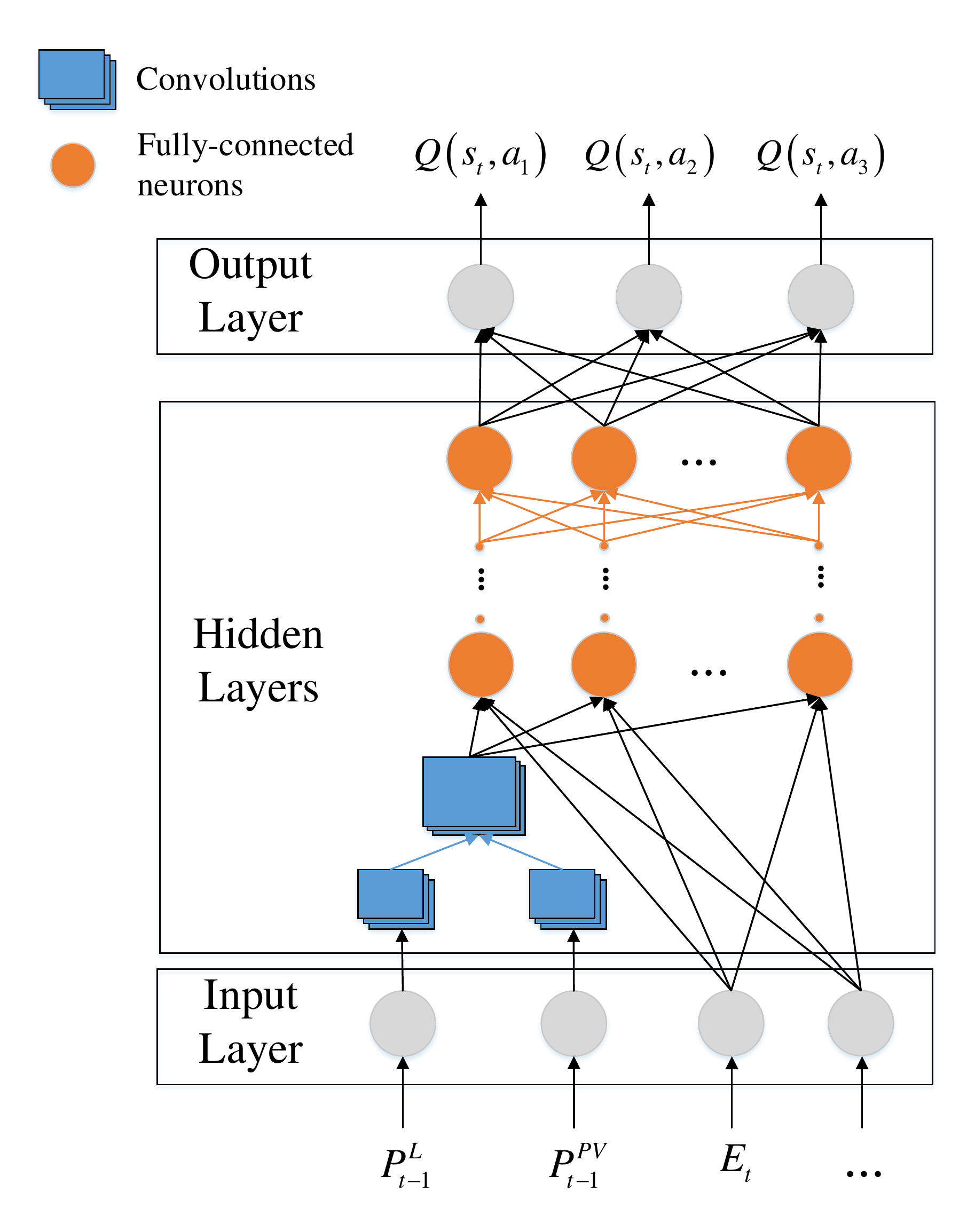}
		\end{minipage}
		\label{CNN_pomdp}
	}
	\subfigure[A RNN architecture for POMDP-based ESM problem.]{
		\begin{minipage}[b]{0.4\textwidth}
			\includegraphics[width=1\textwidth]{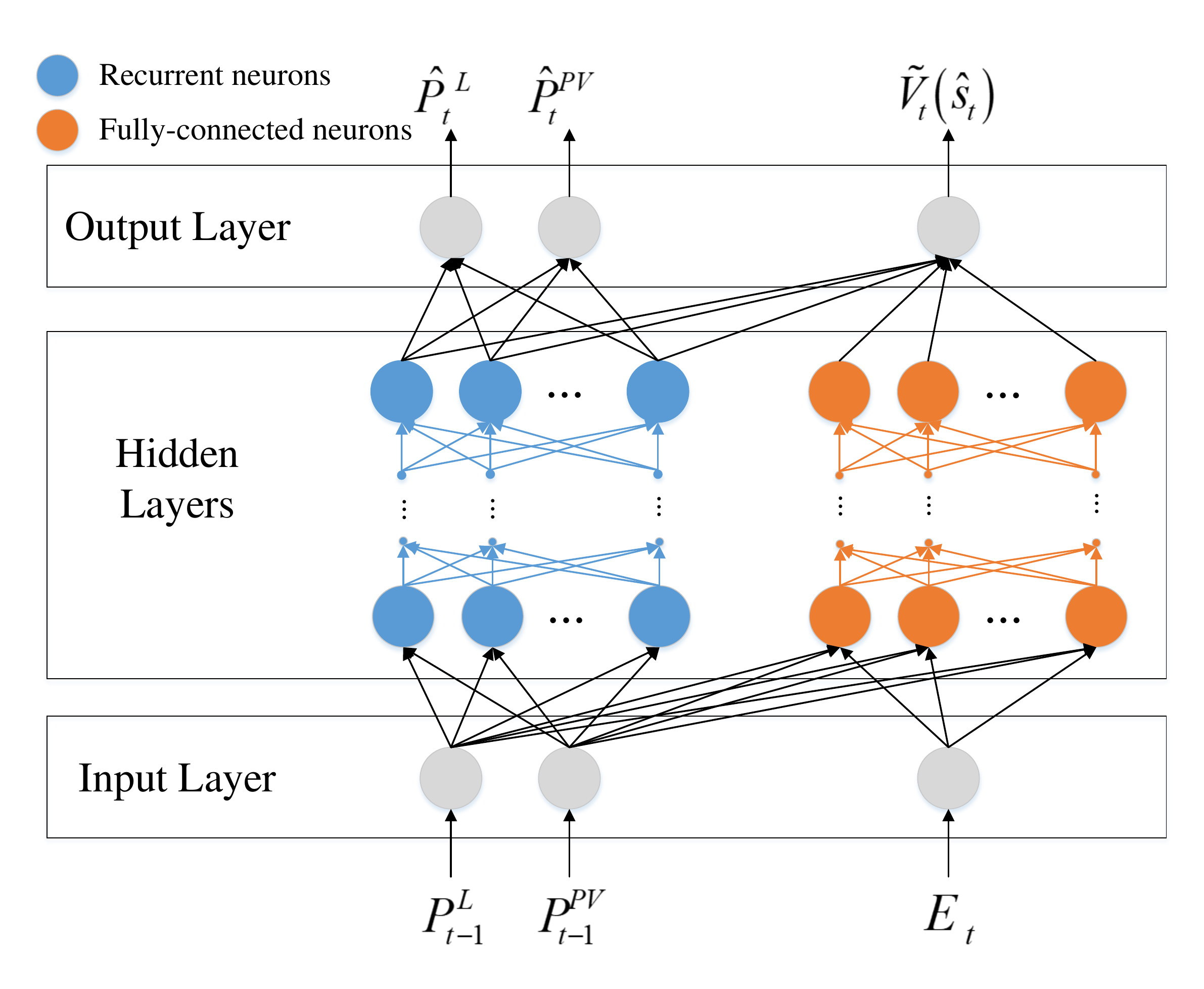}
		\end{minipage}
		\label{RNN_pomdp}
	} 
	\caption{DRL algorithms for POMDP-based ESM problem}
	\label{ESM_alg}
\end{figure}

\paragraph{leverage historical information to deal with uncertainty}
Due to the uncertainty of future renewable energy generation and load demand, MG system dynamics are often formalized as a POMDP in existing research \cite{franccois2016deep,zeng2018dynamic}. The observation of the POMDP model is normally made up of time series of historical data for renewable energy generation and load demand, along with other observable current system states such as the battery SoC level. The agent takes action according to the historical observation and obtains new observation which is then used for historical sequence updates. In \cite{franccois2016deep}, three discrete actions are considered, i.e., charge, discharge, or idle. A DQN-type algorithm is used to solve this POMDP-based DRL model. The authors tried both CNN and RNN to process the time series of historical data as shown in Fig. \ref{CNN_pomdp}, where the historical data is first processed by a set of CNN or LSTM layers, and then the output from the CNN/LSTM layers as well as other inputs are processed by the fully connected layers and output layer. The authors conclude that both NN architectures obtain close results. Compared with CNN, RNN is more widely adopted by other literature due to its capability to better capture the information in historical features. Fig. \ref{RNN_pomdp} shows a designed RNN architecture for a POMDP-based ESM problem in \cite{zeng2018dynamic}, where RNN and approximate policy iteration algorithm are used to solve the problem. The NN consists of two parts as shown in Fig. \ref{RNN_pomdp}, where the left part is an RNN that takes the renewable energy generation and load demand at previous time steps as input, and output the estimated energy generation and consumption at the current time step to derive the immediate reward. On the other hand, the right part is a feed-forward network whose input includes the historical data after it is processed by the RNN input layer of the left part, as well as other state information such as the battery SoC. The output of the right part is the approximated value functions of the input observations. \par 
	
	In the above research on MG, a mass of historical data is essential when solving a POMDP-based DRL model. Table \ref{table_grid3} lists a collection of datasets for research on smart grid. The time series data provided by these datasets include historical real-time price, energy load, renewable energy generation, etc, which can be used to simulate the real-time smart grid system.  \par

\begin{table}[t]
	\centering
	\renewcommand{\arraystretch}{1.2}
	\caption{A Collection of Datasets for Research on Smart Grid.}
	\label{table_grid3}
	%    \resizebox{\textwidth}{!}{
	\begin{tabular}{p{1.5cm}p{5.5cm}} 
		\hline
		\textbf{Dataset} & \textbf{Description} \\
		\cline{1-2}
		\hline
		PJM& public data published by regional transmission organization PJM, such as renewable energy generation data, real-Time locational marginal pricing data.\\
		\hline
		CAISO& real-time data related to the ISO transmission system and its Market, such as system demand forecasts, transmission outage and capacity status, market prices and market result data. \\
		\hline
		MISO& all aspects of real-time and day-ahead energy and ancillary services markets and reliability coordination for the region.\\
		\hline
		MIDC& an access to Oahu Solar Measurement Grid data by Measurement and Instrumentation Data Center (MIDC).
		\\ \hline
	\end{tabular}
	%    }
\end{table}
%\begin{figure}[htb]
%	\centering
%	\includegraphics[width=0.35\textwidth]{pomdp_cnn.pdf}
%	\caption{{\color{red}A Designed CNN Architecture for POMDP-based ESM Problem.}}
%	\label{CNN_pomdp}
%\end{figure}
%
%\begin{figure}[htb]
%	\centering
%	\includegraphics[width=0.45\textwidth]{pomdp_rnn.pdf}
%	\caption{{\color{red}A Designed RNN Architecture for POMDP-based ESM Problem.}}
%	\label{RNN_pomdp}
%\end{figure}

\par

%\cite{mbuwir2017battery}
\par

\subsubsection{Demand Response}
Another method to support the integration of DRES is through DR systems, which dynamically adjust electrical demand in response to changing electrical energy prices or other grid signals. Thermostatically controlled loads (TCLs) such as electric water heaters are a prominent example of loads that offer flexibility at the residential level. In fact, TCLs can be seen as a type of energy storage entity through the power to heat conversion, which is in contract to the direct energy storage entity such as a battery. DR can be divided into direct DR and price-driven DR, where the energy consumption profile of a user is adjusted according to a utility in the former while according to the price in the latter. In any case, the energy consumers need to make a continuing sequence of decisions as to either consumes energy at current (known) utility/price or to defer power consumption until later at possibly unknown utility/prices.\par

	Research works in \cite{mocanu2018line} \cite{wen2015optimal} \cite{o2010residential} \cite{claessens2016convolutional} \cite{ruelens2014demand} \cite{ruelens2016reinforcement} belong to direct DR where energy consumption profile is adjusted according the utility. Among them, \cite{mocanu2018line} \cite{wen2015optimal} \cite{o2010residential} perform optimization of schedules for residential or commercial buildings, while \cite{claessens2016convolutional} \cite{ruelens2014demand} \cite{ruelens2016reinforcement} perform optimization of schedules for a cluster by using the thermal storage. \par
	
	In \cite{mocanu2018line} \cite{wen2015optimal} \cite{o2010residential}, the state space is made up of price state, energy demand state and/or time state, and the action space is DR devices on/off change. Two components are normally being considered for the rewards: one is the controllable load cost which is an important part for demand side management, while the other is the consumer satisfaction which measures the consumer's preference over the DR.\par
	
	In \cite{claessens2016convolutional} and \cite{ruelens2016reinforcement}, the on/off state of DR device is taken into account in state space. \cite{claessens2016convolutional} formalizes the direct load control problem as a high-dimensional POMDP problem, adding history of observations to the state. The researchers in \cite{claessens2016convolutional} also proposes a novel DQN with CNN architecture to solve the problem. 
	
	Research works in \cite{lu2018dynamic} and \cite{lu2019incentive} belong to price-driven DR where energy consumption profile is adjusted according to the price. In these works, the price state is considered as an important component of the state space. The action in \cite{lu2018dynamic} is the selection of retail price while in \cite{lu2019incentive} is the incentive rate for consumers. Both of these two research works choose broker as the agent, who provides services between the utility company and consumer by purchasing energy from the utility company and selling it to the consumer. In this way, broker's cost is considered in the reward function.

\subsubsection{Energy Trading}
The integration of the DRES into the power grid blurs the distinction between an energy provider and a consumer. This is especially true for an MG, which may constantly switch its role between a provider or consumer depending on whether its generated energy exceeds or falls short of its demanded energy. In fact, a key goal of smart grid design is to facilitate the two-way flow of electricity by enhancing the ability of distributed small-scale electricity producers, such as small wind farms or households with solar panels, to sell energy into the power grid. Due to the unpredictability of the DRES, an autonomous control mechanism to ensure power supply/demand balance is essential. One promising method is through the introduction of Broker Agents, who buy electricity from distributed producers and also sell electricity to consumers. RL/DRL can be applied for the Broker Agents to learn pricing decisions to effectively maintain that balance, and earn profits while doing so, contribute to the stability of the grid through their continued participation. \par

	There are three types of entities in the scenario of energy trading where they play different roles in a smart grid system, i.e., producer, broker and consumer. Consumers' energy consumption may influence the producer's decision on the wholesale energy market price. Broker buys energy from the producer through a wholesale energy market and provides it to the consumers through a retail electricity market. 
	
	In \cite{lim2014strategic}, only consumer is considered in energy trading, where an MA system is presented for an isolated MG. Four types of agents are associated with energy generation, energy storage, energy consumption and energy bidding, respectively. \cite{xiao2017energy} and \cite{xiao2018reinforcement} consider producer and consumer as two entities in energy trading. $N$ MGs are connected with each other and the main grid, where each MG can receive energy from other MGs, main grid, local DGs, RESs and \textbf{batteries}. In \cite{xiao2017energy}, the amount of energy that MG $i$  intends to trade with MG $j$ is decided by MG $i$ according to its energy demand, energy generation, energy storage as well as price state. Researchers proposed a Hotbooting Q-Learning based energy trading algorithm to improve the utility of the MG and reduce the total energy purchased from the main grid. In \cite{xiao2018reinforcement}, MG chooses an energy trading amount by estimating the local energy demand, renewable energy generation, battery level in an experience sequence which consists of the current state and previous state-action pairs. A DQN-based energy trading scheme that uses CNN as the nonlinear function approximator is proposed to compress the state space of MG and make it easier to estimate the Q-value of each energy trading policy.\par
	
	In \cite{kim2014dynamic} \cite{kim2016dynamic} \cite{reddy2011strategy} \cite{foruzan2018reinforcement}, multiple energy trading processes among producer, broker and consumer are discussed. \cite{kim2014dynamic} and \cite{reddy2011strategy} set broker as the agent who takes energy demand state, price state and/or time state as state input, and chooses action to sell/buy in energy trading. \cite{kim2016dynamic} extends the above simple system model to a MA MG system where not only the broker but also each consumer can learn an optimal policy for its decision making. The authors also propose two improvements to allow the DRL model to be conducted online in a fully distributed manner with faster speed.\par    

	\cite{foruzan2018reinforcement} also presents an MA-based distributed energy and load management approach that involves energy producer, broker and consumer. Each agent is located in either the supply or the demand side. The aim of these agents is to optimize its utility in the auction-based market. Firstly, each agent initializes its learning parameters. Next, in each iteration of the algorithm, the producer agent predicts its production and the consumer agent selects an energy trading amount and/or price and submits it to the market. Then, the amount and price of energy to be traded are determined by the broker and each agent can obtain its reward and update the parameter. \par

%
%To overcome the challenges of implementing dynamic pricing and energy scheduling, the authors in \cite{kim2014dynamic} and \cite{kim2016dynamic} study RL algorithms that allow each service provider and each customer to learn their policy with no need of prior information about the microgrid. A microgrid energy trading scheme based on RL is proposed in \cite{xiao2018reinforcement}, which applies the DQN to improve the utility of the microgrids for the case of microgrids with a large number of connections.
%In \cite{wang2017reinforcement}, an adaptive learning algorithm is designed to find the Nash equilibrium (NE) of constrained energy trading game among individual strategic participants with incomplete information. Each player's goal is to maximize his average utility by generating a motion probability distribution based on his private information using a learning automaton scheme.
%In \cite{reddy2011strategy}, the authors employ MDP and RL to investigate the learning of pricing strategies for an autonomous Broker Agent to profitably participate in a Tariff Market.

\subsubsection{Comparison and Insights}
By summarizing and comparing the above literature, the following insights can be obtained.

\begin{itemize}
	    
		\item System model: For DR, most of the research works focus on direct DR, while price-driven DR deserves more detailed studies. For Energy Trading, most of the works focus on system models in producer-broker-consumer or producer-consumer mode, few works put the emphasis on consumer-broker or producer-broker mode.
		
		\item DRL model: For ESM problem, battery SoC is usually considered as part of the input state. When facing partial observability problems, historical features can be used as input states and POMDP can be formalized. As the MA problem is widely considered in Energy Trading, DRL models for different agents are usually built in a distributed way.
		
		\item DRL algorithm: As is shown in Table \ref{table_grid2}, Q-Learning and DQN have been widely used and developed to solve different kinds of problems due to their simplicity, good performance and the fact that the action space in smart grid problem is not relatively large and is always discrete. So, it is convenient to use these value-based methods of DRL algorithms. Classical NN architecture can be used to improve the performance of DRL methods. The work in \cite{zeng2018dynamic} and \cite{claessens2016convolutional} have taken advantage of RNN and CNN respectively to extract more features in the input state. Actor-critic methods can be used when it requires fewer samples and less computational resources to train the DRL models. Moreover, compared with value-based methods like DQN, it is possible to learn stochastic policies and solve RL problems with continuous actions by using actor-critic methods.
		
		\item Implementation: In ESM, most DRL algorithms are centralized implemented in EMS. As for DR, most research concentrate on centralized rather than distributed implementation. It is worthwhile to devote more attention to the cooperation between the DR devices.

\end{itemize}

	\subsection{Comparison Between General DRL Model in AIoT and DRL Models in the Literature}

	Comparing the general DRL model in AIoT proposed in Section III, and the DRL models in existing works reviewed in Section IV, the following facts are summarized:
	
	\begin{itemize}
		
		\item IoT Communication Networks: Table \ref{IoTComm2} shows that most existing research has focused only on the network layer, while there are a few studies in the literature that consider both the perception layer and the network layer in WSAN.
		
		\item IoT Edge/Fog/Cloud Computing Systems: Table \ref{table_edge2} shows that most existing research has focused on the application layer, while there are a few studies in the literature that consider both the application layer and the communication layer, i.e., joint communication and computation resource control. 
		
		\item Autonomous Robots: Table \ref{table_robot} shows that most existing research has focused on the perception layer. In cloud robotics, both the perception layer and the application layer have been involved. 
		
		\item Smart Vehicles: Table \ref{table_vehicle2} shows that in most existing works, the autonomous driving problems have focused on the perception layer of AIoT. In vehicular networking problems, most works have only been related to the network layer. There are a few studies in the literature that are related to both the perception layer and the network layer where the vehicle control and vehicular networking have been simultaneously considered \cite{challita2019interference,pal2018reinforcement}. As for the vehicular edge/fog/cloud computing problems, the majority of the studies have focused on the application layer, while only a few research works are related to both the network layer and the application layer \cite{he2018integrated} where the performance of not only caching and computing but also networking have been taken into consideration.
		
		\item Smart Grid: Table \ref{table_grid2} shows that current works have mainly focused on the perception layer of AIoT systems.
		
	\end{itemize}

While the majority of existing research has focused on only one of the three layers of AIoT systems, it will be interesting to explore the joint and integrated control of multiple layers of AIoT systems. For example, sensors are most widely used for perceiving the environment in autonomous driving problems. The cameras on the smart vehicles are used for taking pictures of the traffic environment, and the on-board radars are used to sense the nearby vehicles or obstacles. Based on the information obtained by sensors, the agents make decisions on the controlling of the smart vehicles. On the other hand, communications in IoV enable information exchange between smart vehicles. The communications can help the smart vehicles better perceive the traffic environment, but may also introduce extra delay and affect the reliability of decision making in vehicle control. In other words, the performance of the communication network may influence the effectiveness of message transmission and thereby influence the vehicle control. In current vehicular networking problems, most works are only related to the network layer and focus on optimizing network performance such as delay and reliability. In the future works, researchers can make effort on an effective combination of autonomous driving and IoV communications issues, where the communication resource control and vehicle control are jointly optimized, while the reward can be determined by the ultimate objective - driving performance of the controlled vehicle.\par
Similarly in most existing works on smart grid, the system state and observation information are considered to be readily available to the agents for optimal decision. However in reality, as the state information needs to be obtained by sensors distributed throughout the smart grid, and transmitted to the EMS possibly from remote areas over long distance via the IoT communications network \cite{Bedi2018}, the communication layer can be included in the DRL model as well to reflect the real-world problem more accurately. Although the problem of smart grid state estimation over IoT networks has been addressed in some recent research works \cite{Rana2017,Rana2017a}, there have not been studies on incorporating the delay and inaccuracy in state estimation in the DRL models for optimal control of smart grid to the best of our knowledge.\par

\section{Challenges, Open Issues, and Future Research Directions}
Although DRL is a powerful theoretical tool that is well-suited to the task of introducing artificial intelligence to AIoT systems, there are still a lot of challenges and open issues to be overcome and addressed. The following lists some of the future research directions in this area.\par

\subsection{Incomplete Perception Problem}
In AIoT systems, it might not be possible for the agent to have perfect and complete perception of the state of the environment. This could be due to
\begin{itemize}
	\item limited sensing capabilities of sensors in the perception layer;
	\item information loss due to limited transmission capability in the network layer;
\end{itemize} 

An important challenge in applying DRL to AIoT system is to learn with incomplete perception or partially observable states. The MDP model is no longer valid, as the state information is no longer sufficient to support the decision on optimal action. The action can be improved if more information is available to the agent in addition to the state information. Although the DRL algorithms and methods introduced in Section II.E can be applied, there are still some open issues with the POMDP-based DRL algorithms. Firstly, an agent in POMDP needs to select an action based on the observation history space which grows exponentially. Approaches proposed for this problem require large memory and can only work well for small discrete observation spaces \cite{igl2018deep}. Secondly, when introducing belief state to POMDP problems, the belief space will not grow exponentially but the knowledge of the model becomes essential for the agent, which is not suitable for many complicated scenarios. Finally, nearly all these algorithms in POMDP problems need to face a challenge referred to as information gathering and exploitation dilemma. In a POMDP, the agent does not know what the current state is exactly. It needs to decide whether to gather more information about the true state first or to exploit its current knowledge first. Obviously, in order to find the optimal policy, an agent in POMDP needs to have more interactions with the environment. Apart from the above challenges associated with POMDP-based DRL problems, the DRL model formulation and parameter optimization for various AIoT systems are different case by case. Moreover, more efficient algorithms could be designed according to the specific characteristics of AIoT systems. \par

\subsection{Delayed Control Problem}
In DRL problems, we normally consider that an action is exerted as soon as it is selected by the agent, and a corresponding reward is immediately available at the agent. However, a challenge in applying DRL to real-world AIoT system is to learn despite the existence of control delay, i.e., the delay between measuring a system's state and acting upon it. Control delay is always present in real systems due to transporting measurement data to the learning agent, computing the next action, and changing the state of the actuator. Therefore, it is important to design RL/DRL algorithms which take the control delay into account.\par

Most of the existing RL algorithms don't consider the control delay. At each time step $t$, the state $s_{t}$ of the environment is observed, and an action $a_{t}$ is immediately determined by the agent. However in practice, the actual action $\hat{a}_{t}$ executed at time step $t$ might be the action generated $\tau$ time steps before, i.e., $\hat{a}_{t}=a_{t-\tau}$. In this case, the next state $s_{t+1}$ depends on the current state and a previously determined action, i.e.,  $(s_{t},a_{t-\tau})$, instead of the current state and currently determined action pair $(s_{t},a_{t})$, which makes the state transition violating the Markov property. Therefore, the MDP model based on which RL/DRL algorithms are developed are no longer valid and a POMDP model is more appropriate.\par

In order to deal with the delayed control problem, existing works in RL developed several methods \cite{Katsikopoulos2003,Walsh2009,Schuitema2010}. The first method \cite{Katsikopoulos2003} incorporates the past actions taken during the length of the delay into the current state in formulating an MDP model, so that the classical RL methods such as TD-learning and Q-Learning can be applied. However, this method results in larger state space with the state dimensionality depending on the number of time steps for the delay. The second method \cite{Walsh2009} learns a state transition model so that it can predict the state at which the currently selected action is actually going to be executed. Then, a model-based RL algorithm can be applied. However, the learning process of the underlying model is usually time-consuming and will incur additional delay itself. Finally in the third method \cite{Schuitema2010},  the classical model-free RL algorithms such as TD-learning and Q-Learning are applied, except that at each time step $t$, the Q-function $Q(s_{t},\hat{a}_{t})$ with respect to current state $s_{t}$ and actually executed action $\hat{a}_{t}$ is updated, instead of the normal $Q(s_{t},a_{t})$ with respect to current state $s_{t}$ and currently generated action $a_{t}$.\par

The above methods mostly focus on the constant delay problem. However, the actual delay in an AIoT system is likely to be stochastic. Moreover, the delay can depend on the communication and computation resource control actions in the IoT communications networks and edge/fog/cloud servers. Therefore, developing RL algorithms to consider stochastic control delay or control delay that depends on other parameters is an open issue. Another important challenge is how to extend the above algorithms from RL to DRL leveraging the powerful NNs while dealing with the intrinsic complexities.\par 

\subsection{Multi-Agent Control Problem}
The agent in RL is a virtual concept that learns the optimal policy by interacting with the environment. In AIoT system, agents can be implemented in IoT devices, edge/fog servers, and cloud servers as discussed previously. For a single RL task, there are some typical scenarios for the implementation of agents:
\begin{itemize}
	\item centralized architecture: a single agent in a cloud server, edge/fog node, or an IoT device;
	\item distributed architecture: multiple agents with each agent implemented in an IoT device or edge/fog server;
	\item semi-distributed architecture: one centralized agent in a cloud server or edge/fog server and multiple distributive agents in edge/fog servers or IoT devices. 
\end{itemize}

	\par
	For distributed and semi-distributed architecture, it is an important challenge to enable efficient collaboration and fair competition among multiple agents in a single RL task. The tasks of each agent in a MA system may be different, and they are coupled to each other. Therefore, the design of a reasonable joint reward function becomes a challenge, which may directly affect the performance of the learning policy. 
	Compared to the stable environment in the single-agent RL problem, the environment in the MA RL is complex and dynamic, which brings challenges to design of MA DRL approaches.
	
	In most existing MA DRL methods, the agents are assumed to have same capability. For examples, the robots in a multi-robot system have the same manipulation ability, or the multiple vehicles in a cooperative driving scenario have the same kinematic performance. Thus, the application of DRL in heterogeneous MA systems remains to be further studied. The heterogeneity makes cooperative decision more complex, since each agent needs to model other agents when their capabilities are unknown.
	Although the MA DRL algorithms and methods introduced in Section II.F can be applied to solve the problem of space explosion and guarantee the convergence of the algorithm, the DRL model formulation, parameter optimization, as well as algorithm adaptation and improvement remain to be open issues. Moreover, significant progress in the field of MA RL can be achieved by a more intensive cross-domain research between the fields of ML, game theory, and control theory.
	\par 

\subsection{Joint Resource and Actuator Control Problem}
In AIoT systems, there are two levels of control, i.e., resource control and actuator control as discussed previously. Although the ultimate objective is to optimize the long-term reward of the physical system by selecting appropriate actuator control actions, the computation and network resource control actions will impact the physical system performance through their effects on the network and computation system performances. For example, an efficient network resource control policy can result in larger data transmission rates for the sensory data, and thus allow more information to be available at the cloud server for the agent to derive an improved policy. Currently, most existing research works either optimize the computation and/or network performances for IoT systems, or optimize the physical system performance considering an ideal communication and computation environment. Therefore, how to jointly optimize the two levels of control actions to achieve an optimized physical system performance is an important open issue for applying DRL in AIoT system. \par 

When the RL/DRL environment includes more than one layer in AIoT architecture, the corresponding RL/DRL model will be more complex as discussed in Section III. For example, instead of optimizing normal network performance such as transmission delay, transmission power, and packet loss rate in the network layer, the communication resource control actions need to be selected to optimize the control performance of a physical autonomous system, which may be a function of the network performance. In order to optimize the control performance, the best trade-off between several network performance metrics may need to be considered. For example, larger amount of sensory data may be transmitted at the cost of larger transmission delay, which relieves the incomplete perception problem but deteriorates the delayed control problem as discussed above. \par
There are many challenges to model and solve such complex RL/DRL problems. Firstly, feature selection is an crucial task. An appropriate feature selection can lead to better generalization which is helpful for the bias-overfitting tradeoff. When too many features are taken into consideration, it is hard for the agent to determine which features are more indispensable. Although some features may play a key role in reconstruction of the observation, they may be discarded because they are not related to the current task directly. Secondly, the selection of algorithm and function approximator is also a tough task. The function approximator used for value function or policy converts the features into abstraction in higher level. Sometimes the approximator is too simple to avoid the bias, while sometimes the approximator is too complex to obtain a good generalization result from the limited dataset, i.e., overfitting. Errors resulted from this bias/overfitting problem need to be overcome, so an appropriate approximator needs to be used according to the current task. Thirdly, in such complex RL/DRL problems, the objective function needs to be modified. Typical approaches include reward shaping and discount factor tuning. Reward shaping adds an additional function $F(s_t,a_t)$ to the original reward function $r(s_t,a_t)$. It is mainly used for DRL problems with sparse and delayed rewards \cite{lample2017playing}. Discount factor tuning helps to adjust the impact of temporally distant rewards. When the discount factor is high, the training process tends to be instable in convergence and when the discount factor is low, some potential rewards will be discarded \cite{franccois2018introduction}. Hence, modifying the objective function can help to tackle the above problems to some extent.\par

\section{Conclusion}
This paper has presented the model, applications and challenges of DRL in AIoT systems. Firstly, a summary of the existing RL/DRL methods has been provided. Then, the general model of AIoT system has been proposed, including the DRL framework for AIoT based on the three-layer structure of IoT. The applications of DRL in AIoT have been classified into several categories, and the applied methods and the typical state/action/reward in the models have been summarized.
Finally, the challenges and open issues for future research have been identified.

\appendix

%\section{{\color{red}Building Blocks of DRL}}
\subsection{Building Blocks of DRL - Reinforcement Learning}

\begin{figure}
	\centering
	%% label for second subfigure
	\includegraphics[width=0.45\textwidth]{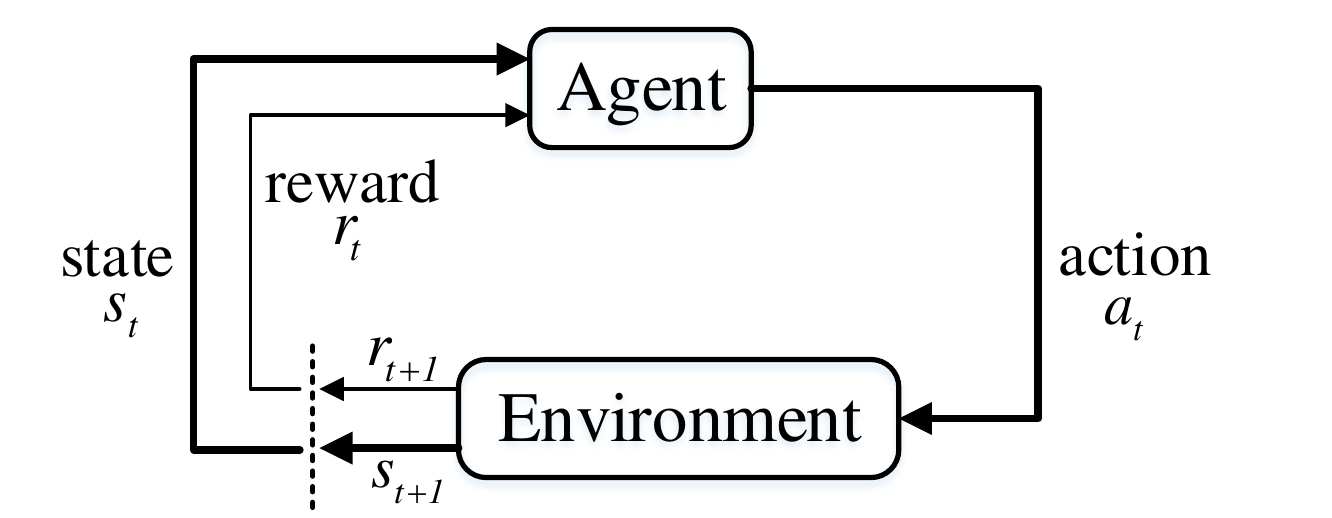}
	\caption{Reinforcement learning process.}
	\label{RL}
	%\label{fig:subfig} %% label for entire figure
\end{figure}

Generally, reinforcement learning (RL) is a type of algorithms in machine learning (ML) that can achieve optimal control of a Markov Decision Process (MDP) \cite{sutton2018reinforcement}. As discussed in Section I, there are generally two entities in RL as shown in Fig. \ref{RL} - an agent and an environment. The environment evolves over time in a stochastic manner and may be in one of the states within a state space at any point in time. The agent performs as the action executor and interacts with the environment. When it performs an action under a certain state, the environment will generate a reward function as signals for positive or negative behaviour. Moreover, the action will also impact on the next state that the environment will transit to. The stochastic evolution of the state-action pair over time forms an MDP, which consists of the following elements

\begin{itemize}
	\item state $s$, which is used to represent a specific status of environment in a possible state space $\mathcal{S}$. In MDP, the state comprises all the necessary information of the environment for the agent to choose the optimal action from the action space.
	\item action $a$, which is chosen by the agent from an action space $\mathcal{A}$ in a specific state $s$. An RL agent interacts with the environment and learn how to behave in different states by observing the consequences of its actions.
	\item reward $r(s,a)$, which is generated when the agent takes a certain action $a$ in a state $s$. Reward indicts the intrinsic desirability of an action in a certain state. 
	\item transition probability $P\left(s'|s,a\right)$, which is the conditional probability that the next state of system will be $s'\in\mathcal{S}$ given the current state $s$ and action $a$. In model-based RL, this transition probability is considered to be known by the agent, while agent does not require this information in model-free RL.
\end{itemize}

%policy
A policy determines how an agent selects actions in different states, which can be categorized into either a stochastic policy or a deterministic policy \cite{franccois2018introduction}. In stochastic case, the policy is described by $\pi\left(s, a\right):=P(a|s)$, which denotes the probability that an action $a$ may be chosen in state $s$. In deterministic case, the policy is described by $\pi\left(s\right):=a$, which denotes the action $a$ that must be chosen in state $s$. \par

For simplicity of introduction, we focus on the discrete time model, where the agent interacts with the environment at each of a sequence of discrete time steps $t=0,1,2,3,\cdots$. The goal of agent is to learn how to map states to actions, i.e., to find a policy $\pi$ to optimize the value function $V^{\pi}(s_{0})$ for any state $s_{0}\in\mathcal{S}$. The value function $V^{\pi}(s_{0})$ is the expected reward when a policy $\pi$ is taken with an initial state $s_{0}$, i.e., 
\begin{equation}
V^{\pi}(s_{0})=\mathrm{E}_{\tau_{s_{0}}\sim \pi}[G(\tau_{s_{0}})],
\label{equ1}
\end{equation} 
where $\mathrm{E}$ stands for expectation, $\tau_{s_{0}}$ is a trajectory or sequence of triplets $(s_{t},a_{t},r_{t+1})$, $t\in\{0,1,2,\cdots\}$ with $r_{t+1}=r(s_{t},a_{t})$, $a_{t}\sim \pi\left(s_{t}, a_{t}\right)$ or $a_{t}=\pi\left(s_{t}\right)$ and $s_{t+1}\sim P\left(s_{t+1}|s_{t},a_{t}\right)$. $G(\tau_{s_{0}})=\sum_{t=1}^{T}f(r_{t})$ can be the total reward, discounted total reward, or average reward of trajectory $\tau_{s_{0}}$, where $T$ is the terminal time step that can be $\infty$. \par 

Apart from value function, another important function is Q-function $Q^{\pi}(s_{0},a_{0})$, which is the expected reward for taking action $a_{0}$ in state $s_{0}$ and thereafter following a policy $\pi$. When policy $\pi$ is the optimal policy $\pi^{*}$, value function and Q-function are denoted by $V^{*}(s)$ and $Q^{*}(s, a)$, respectively. Note that $V^{*}(s)=\max_{a}Q^{*}(s, a)$. If the Q-functions $Q^{*}(s, a)$, $a\in\mathcal{A}$ are given, the optimal policy can be easily found by $\pi^{*}=\arg\max_{a}Q^{*}(s, a)$. \par

In order to learn the value functions or Q-functions, the Bellman optimality equations are usually used. Taking the discounted MDP with a discount factor of $\gamma$ for example, the Bellman optimality equations for value function and Q-function are
\begin{equation}
\begin{split}
V^{*}(s_{t})=\max_{a_{t}}[r_{t+1}+\gamma\sum_{s_{t+1}}P(s_{t+1}|s_{t},a_{t})V^{*}\left(s_{t+1}\right) ],
\label{equ2}
\end{split}
\end{equation}
\noindent and
\begin{equation}
\begin{split}
&Q^{*}(s_{t}, a_{t})=  \\ 
&r_{t+1}+\gamma\sum_{s_{t+1}}P(s_{t+1}|s_{t},a_{t}) \max _{a_{t+1}}Q(s_{t+1},a_{t+1}),
\label{equ3}
\end{split}
\end{equation}
\noindent respectively. \par

Bellman equations represent the relation between the value/Q-functions of the current state and next state. For example, it can be inferred from \eqref{equ3} that the expected reward equals to the sum of the immediate reward and the maximum expected reward thereafter. When the future expected reward is obtained, the expected reward since current state can be calculated. Bellman equations are the basis of an important class of RL algorithms using the ``bootstrap" method, such as Q-Learning and Temporal-Difference (TD)-learning. During the learning process, the agent first initializes the value/Q-functions to some random values. Then, it iteratively repeats the policy prediction and policy evaluation phases until the convergence of the value/Q-functions. In the policy prediction phase, the agent chooses an action according to the current value/Q-functions, which results in an immediate reward and a new state. In the policy evaluation phase, it updates the value/Q-functions according to the Bellman equations \eqref{equ2} or \eqref{equ3} given the immediate reward and the new state.  \par

In the policy prediction phase, instead of always selecting the greedy action that maximizes the current value/Q-functions, a ``soft" policy such as $\epsilon$-greedy, $\epsilon$-soft, and softmax is usually used to explore the environment seeking the potential to learn a better policy. Moreover, according to the different methods adopted in policy evaluation phase, RL algorithms can be either \textit{on-policy} or \textit{off-policy}, depending on whether the value/Q-functions of the predicted policy or an hypothetical (e.g., greedy) policy is estimated.  \par

%Usually, a large amount of memory is required to store the value functions and Q-functions. In some cases when only small, finite state sets are involved, it is possible to store these in the form of tables or arrays. This method is called the \textit{tabular} method.\par
%
%However, in most of the real-world problems, the state sets are large, sometimes infinite, which makes it impossible to store the value functions or Q functions in the form of tables. Therefore, the trial-and-error interaction with the environment is hard to be learned due to the formidable computation complexity and storage capacity requirements. This is where DL comes into the picture - some functions of RL such as Q functions or policy functions are approximated with a smaller set of parameters by the application of DL. The combination of RL and DL results in the more powerful DRL.\par

\begin{figure}[!t]
	\centering
	%\label{fig:subfig:b} %% label for second subfigure
	\includegraphics[width=0.45\textwidth]{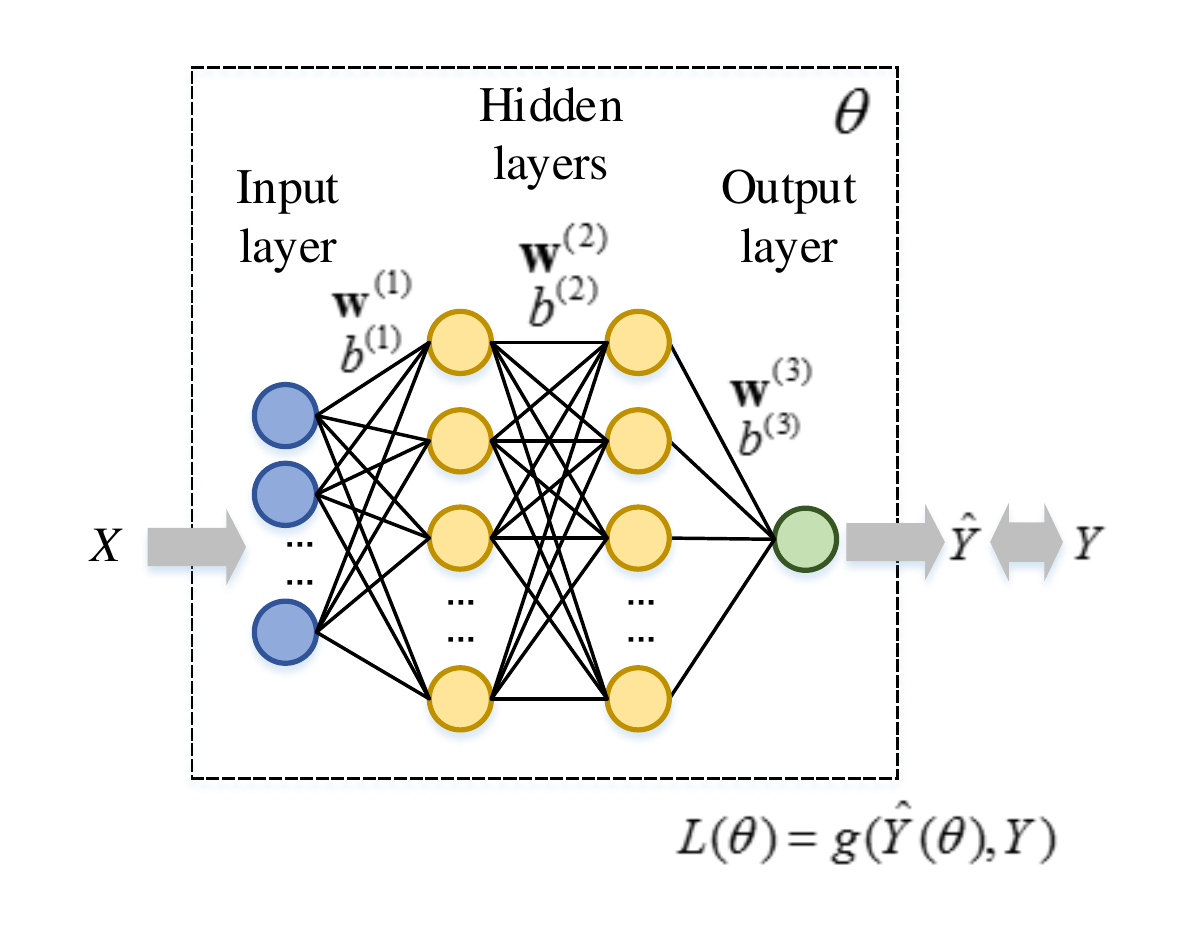}
	\caption{Feedforward neural network.}
	\label{dl_feedforward} %% label for entire figure
\end{figure}

\subsection{Building Blocks of DRL - Deep Learning}
Deep learning (DL) refers to a subset of ML algorithms and techniques that leverage artificial neural networks (ANN) to learn from large amount of data in an autonomous way. It is able to perform well in tasks like regression and classification. Regression task deals with predicting a continuous value, while classification task predicts the output from a set of finite categorical values. Given input data $X$ and output data $Y$, NN models can be viewed as mathematical models defining a function $f:X\rightarrow Y$ or a distribution over $X$ or both $X$ and $Y$. The learning rule of NN modifies its parameters in order for a given input $X$, the network can produce a favored output $\hat{Y}$ that best approximates the target output data $Y$. \par

A general feedforward NN, as shown in Fig. \ref{dl_feedforward}, is constructed by an input layer, one or more hidden layers and an output layer. Each layer consists of one or multiple neurons which represent different non-linear nodes in the model. As illustrated in Fig. \ref{dl_feedforward}, neuron $i$ in layer $j$ has a vector of weights $\mathbf{w}^{(j)}_i$ for the connections from layer $j-1$ to itself and a bias value $b^{(j)}_i$. It also has an activation function $h$ such as Sigmoid, Logistic, Tanh and ReLU. The output of neuron $i$ in layer $j$ equals to $a^{(j)}_i=h({\mathbf{w}^{(j)}_i}\mathbf{a}_{j-1}^{\mathrm{T}}+b^{(j)}_i)$, where $\mathbf{a}_{j-1}$ is the vector of outputs from neurons in layer $j-1$. The typical parameters of NN are the weights and bias of every node. Any NN with two or more hidden layers can be called Deep Neural Network (DNN).\par

\begin{figure}[!t]
	\centering
	%\label{fig:subfig:b} %% label for second subfigure
	\includegraphics[width=0.45\textwidth]{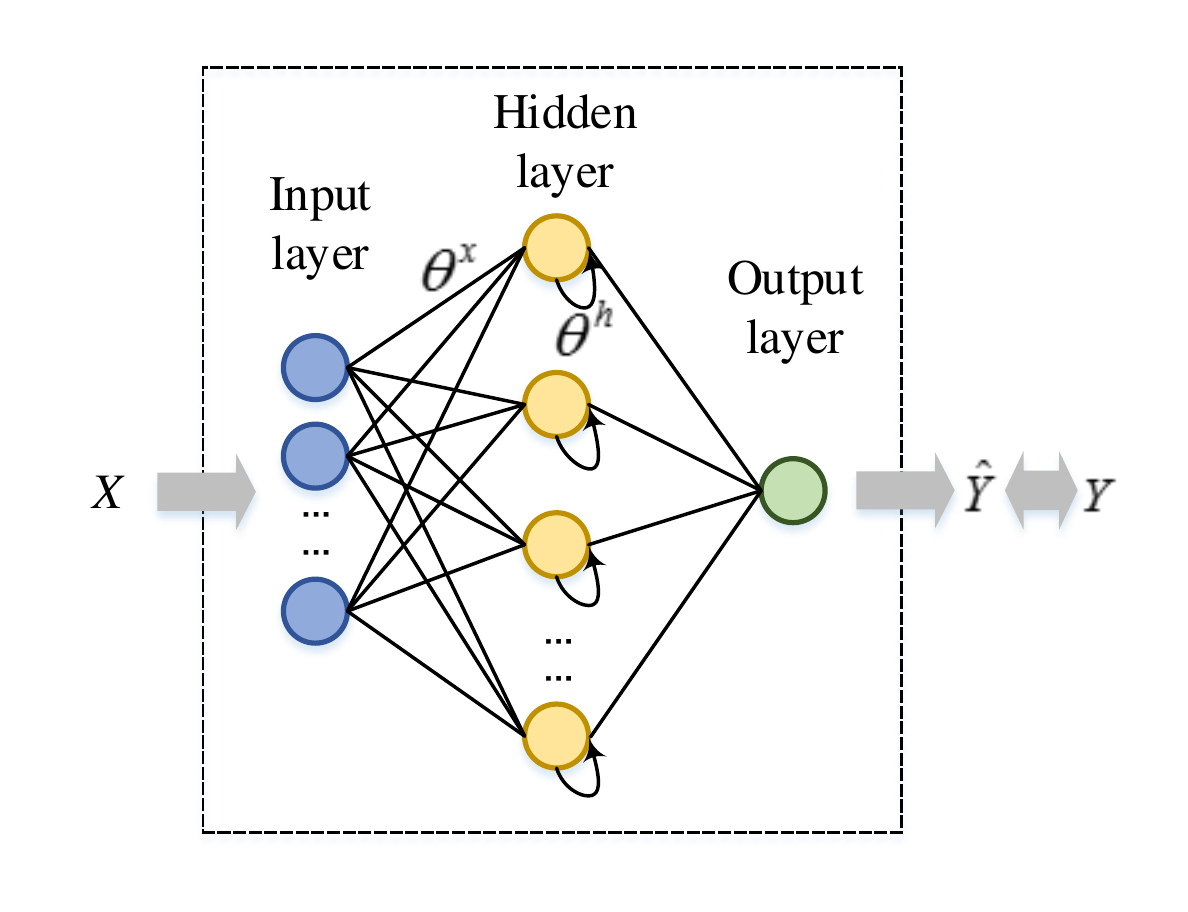}
	\caption{Recurrent neural network.}
	\label{dl_rnn} %% label for entire figure
\end{figure}

A feedforward network has no notion of order in time, and the only input it considers is the current input data it has been exposed to. Recurrent neural network (RNN) refers to a special type of NNs which can process sequences of inputs by using internal memory. In RNN, the prior output of neurons in hidden state can be used as input along with the current input data, which enables the network to learn from history. The basic architecture of RNN is illustrated in Fig. \ref{dl_rnn}. At each time step, the input of RNN is propagated in the feedforward NN. First, it is modified by the weight matrix $\theta^x$. Meanwhile, the hidden state of the previous time step is multiplied by another weight matrix $\theta^h$. Then, those two parts are added together and activated by the neuron. A special RNN architecture called Long Short-Term Memory (LSTM) is widely used\cite{hochreiter1997long}. LSTM is able to solve shortcomings in RNN, i.e. vanishing gradient, exploding gradient and long term dependencies \cite{hochreiter2001gradient}.\par

Usually, a loss function $L(\theta)=g(\hat{Y}(\theta),Y)$ is used in DL, which is a function of the output $\hat{Y}(\theta)$ from NN and the target output $Y$. The loss function evaluates how well a specific NN along with current learned parameter values $\theta$ models the given data $Y=f(X)$. \par

%Loss functions can be classified into two major categories depending on the type of learning tasks - Regression losses and Classification losses. The common regression loss functions include Mean Square Error (MSE), Mean Absolute Error (MAE), Mean Bias Error (MBE). And common classification loss functions include Support Vector Machine (SVM) loss and Cross entropy loss.\par

The objective of the NN is to minimize the loss function, i.e., $\min_{\theta} L(\theta)$. For this purpose, the parameters $\theta$ in NNs are updated by a method called gradient descent. Given a function $L(\theta)$, the simple gradient $\nabla_{\theta}L(\theta)=\frac{\partial L(\theta)}{\partial \theta}$ is usually used to update the parameters. The gradient descent method starts from an initial point $\theta_0$. As a mini-batch of input data is fed to NN, the average loss function over all input data in the mini-batch is derived, and used to find the minimum of $L(\theta)$ by taking a step along the descent direction, i.e.,
\begin{equation}
\label{equ4}
\theta \leftarrow \theta -\alpha\nabla_{\theta}L(\theta),
\end{equation}
\noindent where $\alpha$ is a hyper-parameter named step size. It is set to determine how fast the parameter values move towards the optimal direction. The above process is repeated iteratively as more mini-batches of input data are fed to NN until convergence. \par

The simple gradient $\nabla_{\theta} L(\theta)$ is easy to derive, but simple gradient descend is often not the most efficient method to optimize the loss function. During training, an appropriate value of step size $\alpha$ should be set because if the value is too big, it may not be able to reach the local minimum and if the value is too small, it may take too much time to reach the local optimal point. On the other hand, natural gradients $\nabla_{\theta}^{\mathrm{N}}L(\theta)$ do not follow the usual steepest direction in the parameter space, but along the steepest descent direction with respect to the Fisher metric in the space of distributions. Specifically, the Fisher information metric $F$ is usually used to determine the step size, so that $\nabla_{\theta}^{\mathrm{N}}L(\theta)=\nabla_{\theta}L(\theta)F^{-1}$. Then, \eqref{equ4} can be used to update the parameters by replacing $\alpha\nabla_{\theta}L(\theta)$ with $\nabla_{\theta}^{\mathrm{N}}L(\theta)$.

% References
\bibliography{DRLIoT}{}

\begin{thebibliography}{100}
\providecommand{\url}[1]{#1}
\csname url@samestyle\endcsname
\providecommand{\newblock}{\relax}
\providecommand{\bibinfo}[2]{#2}
\providecommand{\BIBentrySTDinterwordspacing}{\spaceskip=0pt\relax}
\providecommand{\BIBentryALTinterwordstretchfactor}{4}
\providecommand{\BIBentryALTinterwordspacing}{\spaceskip=\fontdimen2\font plus
\BIBentryALTinterwordstretchfactor\fontdimen3\font minus
  \fontdimen4\font\relax}
\providecommand{\BIBforeignlanguage}[2]{{%
\expandafter\ifx\csname l@#1\endcsname\relax
\typeout{** WARNING: IEEEtran.bst: No hyphenation pattern has been}%
\typeout{** loaded for the language `#1'. Using the pattern for}%
\typeout{** the default language instead.}%
\else
\language=\csname l@#1\endcsname
\fi
#2}}
\providecommand{\BIBdecl}{\relax}
\BIBdecl

\bibitem{antsaklis1991}
P.~J. Antsaklis, K.~M. Passino, and S.~Wang, ``An introduction to autonomous
  control systems,'' \emph{IEEE Control Syst. Mag.}, vol.~11, no.~4, pp. 5--13,
  1991.

\bibitem{aIoT}
\BIBentryALTinterwordspacing
``Smarter {Things}: The autonomous {IoT},'' 2018. [Online]. Available:
  \url{http://gdruk.com/smarter-things-autonomous-iot/}
\BIBentrySTDinterwordspacing

\bibitem{Mohammadi2018}
M.~{Mohammadi}, A.~{Al-Fuqaha}, S.~{Sorour}, and M.~{Guizani}, ``Deep learning
  for {IoT} big data and streaming analytics: A survey,'' \emph{IEEE Commun.
  Surveys Tuts.}, vol.~20, no.~4, pp. 2923--2960, 2018.

\bibitem{sutton2018reinforcement}
R.~S. Sutton and A.~G. Barto, \emph{Reinforcement learning: An
  introduction}.\hskip 1em plus 0.5em minus 0.4em\relax MIT press, 2018.

\bibitem{mnih2015human}
V.~Mnih, K.~Kavukcuoglu, D.~Silver, A.~A. Rusu, J.~Veness, M.~G. Bellemare,
  A.~Graves, M.~Riedmiller, A.~K. Fidjeland, G.~Ostrovski \emph{et~al.},
  ``Human-level control through deep reinforcement learning,'' \emph{Nature},
  vol. 518, no. 7540, p. 529, 2015.

\bibitem{Sezer2017}
O.~B. Sezer, E.~Dogdu, and A.~M. Ozbayoglu, ``Context-aware computing,
  learning, and big data in internet of things: a survey,'' \emph{IEEE Internet
  of Things J.}, vol.~5, no.~1, pp. 1--27, 2017.

\bibitem{Samie2019}
F.~Samie, L.~Bauer, and J.~Henkel, ``From cloud down to things: An overview of
  machine learning in {Internet} of {Things},'' \emph{IEEE Internet of Things
  J.}, 2019.

\bibitem{Mahdavinejad2018}
M.~S. Mahdavinejad, M.~Rezvan, M.~Barekatain, P.~Adibi, P.~Barnaghi, and A.~P.
  Sheth, ``Machine learning for {Internet} of {Things} data analysis: A
  survey,'' \emph{Digital Communications and Networks}, vol.~4, no.~3, pp.
  161--175, 2018.

\bibitem{Cui2018}
L.~Cui, S.~Yang, F.~Chen, Z.~Ming, N.~Lu, and J.~Qin, ``A survey on application
  of machine learning for {Internet} of {Things},'' \emph{International Journal
  of Machine Learning and Cybernetics}, vol.~9, no.~8, pp. 1399--1417, 2018.

\bibitem{zantalis2019review}
F.~Zantalis, G.~Koulouras, S.~Karabetsos, and D.~Kandris, ``A review of machine
  learning and {IoT} in smart transportation,'' \emph{Future Internet},
  vol.~11, no.~4, p.~94, 2019.

\bibitem{Chen2019}
Q.~Chen, W.~Wang, F.~Wu, S.~De, R.~Wang, B.~Zhang, and X.~Huang, ``A survey on
  an emerging area: Deep learning for smart city data,'' \emph{IEEE Trans.
  Emerg. Topics Comput. Intell.}, 2019.

\bibitem{Qolomany2019}
B.~Qolomany, A.~Al-Fuqaha, A.~Gupta, D.~Benhaddou, S.~Alwajidi, J.~Qadir, and
  A.~C. Fong, ``Leveraging machine learning and big data for smart buildings: A
  comprehensive survey,'' \emph{IEEE Access}, vol.~7, pp. 90\,316--90\,356,
  2019.

\bibitem{Hossain2019}
E.~Hossain, I.~Khan, F.~Un-Noor, S.~S. Sikander, and M.~S.~H. Sunny,
  ``Application of big data and machine learning in smart grid, and associated
  security concerns: A review,'' \emph{IEEE Access}, vol.~7, pp.
  13\,960--13\,988, 2019.

\bibitem{zhang2018review}
D.~Zhang, X.~Han, and C.~Deng, ``Review on the research and practice of deep
  learning and reinforcement learning in smart grids,'' \emph{CSEE J. of Power
  and Energy Systems}, vol.~4, no.~3, pp. 362--370, 2018.

\bibitem{Sharma2019}
S.~K. Sharma and X.~Wang, ``Towards massive machine type communications in
  ultra-dense cellular {IoT} networks: Current issues and machine
  learning-assisted solutions,'' \emph{IEEE Commun. Surveys Tuts.}, 2019.

\bibitem{Alsheikh2014}
M.~A. Alsheikh, S.~Lin, D.~Niyato, and H.-P. Tan, ``Machine learning in
  wireless sensor networks: Algorithms, strategies, and applications,''
  \emph{IEEE Commun. Surveys Tuts.}, vol.~16, no.~4, pp. 1996--2018, 2014.

\bibitem{Chen2019b}
M.~{Chen}, U.~{Challita}, W.~{Saad}, C.~{Yin}, and M.~{Debbah}, ``Artificial
  neural networks-based machine learning for wireless networks: A tutorial,''
  \emph{IEEE Commun. Surveys Tuts.}, p.~1, 2019.

\bibitem{Zappone2019}
A.~Zappone, M.~Di~Renzo, and M.~Debbah, ``Wireless networks design in the era
  of deep learning: Model-based, ai-based, or both,'' \emph{arXiv preprint
  arXiv:1902.02647}, 2019.

\bibitem{Mao2018}
Q.~Mao, F.~Hu, and Q.~Hao, ``Deep learning for intelligent wireless networks: A
  comprehensive survey,'' \emph{IEEE Commun. Surveys Tuts.}, vol.~20, no.~4,
  pp. 2595--2621, 2018.

\bibitem{Luong2019}
N.~C. Luong, D.~T. Hoang, S.~Gong, D.~Niyato, P.~Wang, Y.-C. Liang, and D.~I.
  Kim, ``Applications of deep reinforcement learning in communications and
  networking: A survey,'' \emph{IEEE Commun. Surveys Tuts.}, 2019.

\bibitem{Abdulkareem2019}
K.~H. {Abdulkareem}, M.~A. {Mohammed}, S.~S. {Gunasekaran}, M.~N. {Al-Mhiqani},
  A.~A. {Mutlag}, S.~A. {Mostafa}, N.~S. {Ali}, and D.~A. {Ibrahim}, ``A review
  of fog computing and machine learning: Concepts, applications, challenges,
  and open issues,'' \emph{IEEE Access}, p.~1, 2019.

\bibitem{Rodrigues2019}
T.~K. {Rodrigues}, K.~{Suto}, H.~{Nishiyama}, J.~{Liu}, and N.~{Kato},
  ``Machine learning meets computation and communication control in evolving
  edge and cloud: Challenges and future perspective,'' \emph{IEEE Commun.
  Surveys Tuts.}, p.~1, 2019.

\bibitem{Chen2019a}
J.~{Chen} and X.~{Ran}, ``Deep learning with edge computing: A review,''
  \emph{Proceedings of the IEEE}, vol. 107, no.~8, pp. 1655--1674, Aug. 2019.

\bibitem{Zhu2018}
H.~{Zhu}, Y.~{Cao}, W.~{Wang}, T.~{Jiang}, and S.~{Jin}, ``Deep reinforcement
  learning for mobile edge caching: Review, new features, and open issues,''
  \emph{IEEE Network}, vol.~32, no.~6, pp. 50--57, Nov. 2018.

\bibitem{Arulkumaran2017}
K.~{Arulkumaran}, M.~P. {Deisenroth}, M.~{Brundage}, and A.~A. {Bharath},
  ``Deep reinforcement learning: A brief survey,'' \emph{IEEE Signal Processing
  Magazine}, vol.~34, no.~6, pp. 26--38, Nov. 2017.

\bibitem{Francois-Lavet2018}
V.~Fran{\c{c}}ois-Lavet, P.~Henderson, R.~Islam, M.~G. Bellemare, J.~Pineau
  \emph{et~al.}, ``An introduction to deep reinforcement learning,''
  \emph{Foundations and Trends{\textregistered} in Machine Learning}, vol.~11,
  no. 3-4, pp. 219--354, 2018.

\bibitem{van2016deep}
H.~Van~Hasselt, A.~Guez, and D.~Silver, ``Deep reinforcement learning with
  double {Q}-learning,'' in \emph{Thirtieth AAAI Conference on Artificial
  Intelligence}, 2016.

\bibitem{franccois2018introduction}
V.~Fran{\c{c}}ois-Lavet, P.~Henderson, R.~Islam, M.~G. Bellemare, J.~Pineau
  \emph{et~al.}, ``An introduction to deep reinforcement learning,''
  \emph{Foundations and Trends{\textregistered} in Machine Learning}, vol.~11,
  no. 3-4, pp. 219--354, 2018.

\bibitem{schaul2015prioritized}
T.~Schaul, J.~Quan, I.~Antonoglou, and D.~Silver, ``Prioritized experience
  replay,'' \emph{arXiv preprint arXiv:1511.05952}, 2015.

\bibitem{wang2015dueling}
Z.~Wang, T.~Schaul, M.~Hessel, H.~Van~Hasselt, M.~Lanctot, and N.~De~Freitas,
  ``Dueling network architectures for deep reinforcement learning,''
  \emph{arXiv preprint arXiv:1511.06581}, 2015.

\bibitem{amari1998natural}
S.-I. Amari, ``Natural gradient works efficiently in learning,'' \emph{Neural
  computation}, vol.~10, no.~2, pp. 251--276, 1998.

\bibitem{kakade2002natural}
S.~M. Kakade, ``A natural policy gradient,'' in \emph{Advances in neural
  information processing systems}, 2002, pp. 1531--1538.

\bibitem{schulman2015trust}
J.~Schulman, S.~Levine, P.~Abbeel, M.~Jordan, and P.~Moritz, ``Trust region
  policy optimization,'' in \emph{International Conference on Machine
  Learning}, 2015, pp. 1889--1897.

\bibitem{silver2014deterministic}
D.~Silver, G.~Lever, N.~Heess, T.~Degris, D.~Wierstra, and M.~Riedmiller,
  ``Deterministic policy gradient algorithms,'' in \emph{ICML}, 2014.

\bibitem{williams1992simple}
R.~J. Williams, ``Simple statistical gradient-following algorithms for
  connectionist reinforcement learning,'' \emph{Machine learning}, vol.~8, no.
  3-4, pp. 229--256, 1992.

\bibitem{konda2000actor}
V.~R. Konda and J.~N. Tsitsiklis, ``Actor-critic algorithms,'' in
  \emph{Advances in neural information processing systems}, 2000, pp.
  1008--1014.

\bibitem{wang2016sample}
Z.~Wang, V.~Bapst, N.~Heess, V.~Mnih, R.~Munos, K.~Kavukcuoglu, and
  N.~de~Freitas, ``Sample efficient actor-critic with experience replay,''
  \emph{arXiv preprint arXiv:1611.01224}, 2016.

\bibitem{mnih2016asynchronous}
V.~Mnih, A.~P. Badia, M.~Mirza, A.~Graves, T.~Lillicrap, T.~Harley, D.~Silver,
  and K.~Kavukcuoglu, ``Asynchronous methods for deep reinforcement learning,''
  in \emph{International conference on machine learning}, 2016, pp. 1928--1937.

\bibitem{haarnoja2018soft}
T.~Haarnoja, A.~Zhou, P.~Abbeel, and S.~Levine, ``Soft actor-critic: Off-policy
  maximum entropy deep reinforcement learning with a stochastic actor,''
  \emph{arXiv preprint arXiv:1801.01290}, 2018.

\bibitem{lillicrap2015continuous}
T.~P. Lillicrap, J.~J. Hunt, A.~Pritzel, N.~Heess, T.~Erez, Y.~Tassa,
  D.~Silver, and D.~Wierstra, ``Continuous control with deep reinforcement
  learning,'' \emph{arXiv preprint arXiv:1509.02971}, 2015.

\bibitem{barth2018distributed}
G.~Barth-Maron, M.~W. Hoffman, D.~Budden, W.~Dabney, D.~Horgan, A.~Muldal,
  N.~Heess, and T.~Lillicrap, ``Distributed distributional deterministic policy
  gradients,'' \emph{arXiv preprint arXiv:1804.08617}, 2018.

\bibitem{fujimoto2018addressing}
S.~Fujimoto, H.~van Hoof, and D.~Meger, ``Addressing function approximation
  error in actor-critic methods,'' \emph{arXiv preprint arXiv:1802.09477},
  2018.

\bibitem{lowe2017multi}
R.~Lowe, Y.~Wu, A.~Tamar, J.~Harb, O.~P. Abbeel, and I.~Mordatch, ``Multi-agent
  actor-critic for mixed cooperative-competitive environments,'' in
  \emph{Advances in Neural Information Processing Systems}, 2017, pp.
  6379--6390.

\bibitem{heess2015memory}
N.~Heess, J.~J. Hunt, T.~P. Lillicrap, and D.~Silver, ``Memory-based control
  with recurrent neural networks,'' \emph{arXiv preprint arXiv:1512.04455},
  2015.

\bibitem{gu2016q}
S.~Gu, T.~Lillicrap, Z.~Ghahramani, R.~E. Turner, and S.~Levine, ``Q-prop:
  Sample-efficient policy gradient with an off-policy critic,'' \emph{arXiv
  preprint arXiv:1611.02247}, 2016.

\bibitem{schulman2017proximal}
J.~Schulman, F.~Wolski, P.~Dhariwal, A.~Radford, and O.~Klimov, ``Proximal
  policy optimization algorithms,'' \emph{arXiv preprint arXiv:1707.06347},
  2017.

\bibitem{wu2017scalable}
Y.~Wu, E.~Mansimov, R.~B. Grosse, S.~Liao, and J.~Ba, ``Scalable trust-region
  method for deep reinforcement learning using kronecker-factored
  approximation,'' in \emph{Advances in neural information processing systems},
  2017, pp. 5279--5288.

\bibitem{shani2013survey}
G.~Shani, J.~Pineau, and R.~Kaplow, ``A survey of point-based {POMDP}
  solvers,'' \emph{Autonomous Agents and Multi-Agent Systems}, vol.~27, no.~1,
  pp. 1--51, 2013.

\bibitem{dai2013pomdp}
P.~Dai, C.~H. Lin, D.~S. Weld \emph{et~al.}, ``{POMDP}-based control of
  workflows for crowdsourcing,'' \emph{Artificial Intelligence}, vol. 202, pp.
  52--85, 2013.

\bibitem{wierstra2007solving}
D.~Wierstra, A.~Foerster, J.~Peters, and J.~Schmidhuber, ``Solving deep memory
  {POMDPs} with recurrent policy gradients,'' in \emph{International Conference
  on Artificial Neural Networks}.\hskip 1em plus 0.5em minus 0.4em\relax
  Springer, 2007, pp. 697--706.

\bibitem{hausknecht2015deep}
M.~Hausknecht and P.~Stone, ``Deep recurrent {Q}-learning for partially
  observable {MDPs},'' in \emph{2015 AAAI Fall Symposium Series}, 2015.

\bibitem{wayne2018unsupervised}
G.~Wayne, C.-C. Hung, D.~Amos, M.~Mirza, A.~Ahuja, A.~Grabska-Barwinska,
  J.~Rae, P.~Mirowski, J.~Z. Leibo, A.~Santoro \emph{et~al.}, ``Unsupervised
  predictive memory in a goal-directed agent,'' \emph{arXiv preprint
  arXiv:1803.10760}, 2018.

\bibitem{egorov2015deep}
M.~Egorov, ``Deep reinforcement learning with {POMDPs},'' 2015.

\bibitem{zhu2018improving}
P.~Zhu, X.~Li, P.~Poupart, and G.~Miao, ``On improving deep reinforcement
  learning for {POMDPs},'' \emph{arXiv preprint arXiv:1804.06309}, 2018.

\bibitem{foerster2016learning}
J.~N. Foerster, Y.~M. Assael, N.~de~Freitas, and S.~Whiteson, ``Learning to
  communicate to solve riddles with deep distributed recurrent {Q}-networks,''
  \emph{arXiv preprint arXiv:1602.02672}, 2016.

\bibitem{bu2008comprehensive}
L.~Bu, R.~Babu, B.~De~Schutter \emph{et~al.}, ``A comprehensive survey of
  multiagent reinforcement learning,'' \emph{IEEE Trans. Syst., Man, Cybern. C,
  Appl., Rev.}, vol.~38, no.~2, pp. 156--172, 2008.

\bibitem{Foerster2017}
J.~Foerster, N.~Nardelli, G.~Farquhar, T.~Afouras, P.~H. Torr, P.~Kohli, and
  S.~Whiteson, ``Stabilising experience replay for deep multi-agent
  reinforcement learning,'' in \emph{Proceedings of the 34th International
  Conference on Machine Learning-Volume 70}.\hskip 1em plus 0.5em minus
  0.4em\relax JMLR. org, 2017, pp. 1146--1155.

\bibitem{van2016coordinated}
E.~Van~der Pol and F.~A. Oliehoek, ``Coordinated deep reinforcement learners
  for traffic light control,'' \emph{Proceedings of Learning, Inference and of
  Multi-Agent Systems (at NIPS 2016)}, 2016.

\bibitem{foerster2018counterfactual}
J.~N. Foerster, G.~Farquhar, T.~Afouras, N.~Nardelli, and S.~Whiteson,
  ``Counterfactual multi-agent policy gradients,'' in \emph{Thirty-Second AAAI
  Conference on Artificial Intelligence}, 2018.

\bibitem{park2016learning}
T.~Park, N.~Abuzainab, and W.~Saad, ``Learning how to communicate in the
  {Internet} of {Things}: Finite resources and heterogeneity,'' \emph{IEEE
  Access}, vol.~4, pp. 7063--7073, 2016.

\bibitem{kwon2019intelligent}
M.~Kwon, J.~Lee, and H.~Park, ``Intelligent {I}o{T} connectivity: Deep
  reinforcement learning approach,'' \emph{IEEE Sensors J.}, 2019.

\bibitem{renaud2006coordinated}
J.-C. Renaud and C.-K. Tham, ``Coordinated sensing coverage in sensor networks
  using distributed reinforcement learning,'' in \emph{2006 14th IEEE
  International Conference on Networks}, vol.~1.\hskip 1em plus 0.5em minus
  0.4em\relax IEEE, 2006, pp. 1--6.

\bibitem{chen2019deep}
J.~Chen, T.~Shu, T.~Li, and C.~W. de~Silva, ``Deep reinforced learning tree for
  spatiotemporal monitoring with mobile robotic wireless sensor networks,''
  \emph{IEEE Trans. Systems, Man, and Cybernetics: Systems}, 2019.

\bibitem{su2019cooperative}
Y.~Su, X.~Lu, Y.~Zhao, L.~Huang, and X.~Du, ``Cooperative communications with
  relay selection based on deep reinforcement learning in wireless sensor
  networks,'' \emph{IEEE Sens. J.}, vol.~19, no.~20, pp. 9561--9569, 2019.

\bibitem{zhu2017new}
J.~Zhu, Y.~Song, D.~Jiang, and H.~Song, ``A new deep-{Q}-learning-based
  transmission scheduling mechanism for the cognitive {I}nternet of things,''
  \emph{IEEE Internet Things J.}, vol.~5, no.~4, pp. 2375--2385, 2017.

\bibitem{oda2017design}
T.~Oda, R.~Obukata, M.~Ikeda, L.~Barolli, and M.~Takizawa, ``Design and
  implementation of a simulation system based on deep {Q}-network for mobile
  actor node control in wireless sensor and actor networks,'' in \emph{2017
  31st International Conference on Advanced Information Networking and
  Applications Workshops (WAINA)}.\hskip 1em plus 0.5em minus 0.4em\relax IEEE,
  2017, pp. 195--200.

\bibitem{kunzel2018weight}
G.~K{\"u}nzel, G.~P. Cainelli, I.~M{\"u}ller, and C.~E. Pereira, ``Weight
  adjustments in a routing algorithm for wireless sensor and actuator networks
  using {Q}-learning,'' \emph{IFAC-PapersOnLine}, vol.~51, no.~10, pp. 58--63,
  2018.

\bibitem{leong2018deep}
A.~S. Leong, A.~Ramaswamy, D.~E. Quevedo, H.~Karl, and L.~Shi, ``Deep
  reinforcement learning for wireless sensor scheduling in cyber-physical
  systems,'' \emph{arXiv preprint arXiv:1809.05149}, 2018.

\bibitem{jiang2019cooperative}
N.~Jiang, Y.~Deng, O.~Simeone, and A.~Nallanathan, ``Cooperative deep
  reinforcement learning for multiple-group {NB-IoT} networks optimization,''
  in \emph{ICASSP 2019-2019 IEEE International Conference on Acoustics, Speech
  and Signal Processing (ICASSP)}.\hskip 1em plus 0.5em minus 0.4em\relax IEEE,
  2019, pp. 8424--8428.

\bibitem{chafii2018enhancing}
M.~Chafii, F.~Bader, and J.~Palicot, ``Enhancing coverage in narrow band-{IoT}
  using machine learning,'' in \emph{2018 IEEE Wireless Communications and
  Networking Conference (WCNC)}.\hskip 1em plus 0.5em minus 0.4em\relax IEEE,
  2018, pp. 1--6.

\bibitem{lei2016iwsn}
L.~Lei, Y.~Kuang, X.~S. Shen, K.~Yang, J.~Qiao, and Z.~Zhong, ``Optimal
  reliability in energy harvesting industrial wireless sensor networks,''
  \emph{IEEE Trans. Wirel. Commun.}, vol.~15, no.~8, pp. 5399--5413, 2016.

\bibitem{chu2018reinforcement}
M.~Chu, H.~Li, X.~Liao, and S.~Cui, ``Reinforcement learning based multi-access
  control and battery prediction with energy harvesting in {IoT} systems,''
  \emph{IEEE Internet Things J.}, 2018.

\bibitem{li2019partially}
D.~Li, S.~Xu, and J.~Zhao, ``Partially observable double {DQN} based {I}o{T}
  scheduling for energy harvesting,'' in \emph{2019 IEEE International
  Conference on Communications Workshops (ICC Workshops)}.\hskip 1em plus 0.5em
  minus 0.4em\relax IEEE, 2019, pp. 1--6.

\bibitem{qiu2019deep}
C.~Qiu, Y.~Hu, Y.~Chen, and B.~Zeng, ``Deep deterministic policy gradient
  ({DDPG}) based energy harvesting wireless communications,'' \emph{IEEE
  Internet Things J.}, 2019.

\bibitem{He2018}
H.~{He}, H.~{Shan}, A.~{Huang}, Q.~{Ye}, and W.~{Zhuang}, ``Reinforcement
  learning-based computing and transmission scheduling for {LTE}-u-enabled
  {IoT},'' in \emph{Proc. IEEE Global Communications Conf. (GLOBECOM)}, Dec.
  2018, pp. 1--6.

\bibitem{Liu2019}
X.~{Liu}, Z.~{Qin}, and Y.~{Gao}, ``Resource allocation for edge computing in
  {IoT} networks via reinforcement learning,'' in \emph{Proc. ICC 2019 - 2019
  IEEE Int. Conf. Communications (ICC)}, May 2019, pp. 1--6.

\bibitem{huang2019deep}
L.~Huang, X.~Feng, C.~Zhang, L.~Qian, and Y.~Wu, ``Deep reinforcement
  learning-based joint task offloading and bandwidth allocation for multi-user
  mobile edge computing,'' \emph{Digital Communications and Networks}, vol.~5,
  no.~1, pp. 10--17, 2019.

\bibitem{Huang2019}
L.~{Huang}, S.~{Bi}, and Y.~J. {Zhang}, ``Deep reinforcement learning for
  online computation offloading in wireless powered mobile-edge computing
  networks,'' \emph{IEEE Trans. Mobile Comput.}, p.~1, 2019.

\bibitem{Lei2019}
L.~{Lei}, H.~{Xu}, X.~{Xiong}, K.~{Zheng}, W.~{Xiang}, and X.~{Wang},
  ``Multi-user resource control with deep reinforcement learning in {IoT} edge
  computing,'' \emph{IEEE Internet of Things J.}, p.~1, 2019.

\bibitem{Qiu2019}
X.~{Qiu}, L.~{Liu}, W.~{Chen}, Z.~{Hong}, and Z.~{Zheng}, ``Online deep
  reinforcement learning for computation offloading in blockchain-empowered
  mobile edge computing,'' \emph{IEEE Trans. Veh. Technol.}, vol.~68, no.~8,
  pp. 8050--8062, Aug. 2019.

\bibitem{Chen2019c}
X.~{Chen}, H.~{Zhang}, C.~{Wu}, S.~{Mao}, Y.~{Ji}, and M.~{Bennis}, ``Optimized
  computation offloading performance in virtual edge computing systems via deep
  reinforcement learning,'' \emph{IEEE Internet of Things J.}, vol.~6, no.~3,
  pp. 4005--4018, Jun. 2019.

\bibitem{Min2019}
M.~{Min}, L.~{Xiao}, Y.~{Chen}, P.~{Cheng}, D.~{Wu}, and W.~{Zhuang},
  ``Learning-based computation offloading for {IoT} devices with energy
  harvesting,'' \emph{IEEE Trans. Veh. Technol.}, vol.~68, no.~2, pp.
  1930--1941, Feb. 2019.

\bibitem{Chen2019d}
J.~{Chen}, S.~{Chen}, Q.~{Wang}, B.~{Cao}, G.~{Feng}, and J.~{Hu}, ``iraf: A
  deep reinforcement learning approach for collaborative mobile edge computing
  {IoT} networks,'' \emph{IEEE Internet Things J.}, vol.~6, no.~4, pp.
  7011--7024, Aug. 2019.

\bibitem{wang2018smart}
J.~Wang, L.~Zhao, J.~Liu, and N.~Kato, ``Smart resource allocation for mobile
  edge computing: A deep reinforcement learning approach,'' \emph{IEEE Trans.
  Emerg. Top. Comput.}, 2019.

\bibitem{Ren2019}
J.~{Ren}, H.~{Wang}, T.~{Hou}, S.~{Zheng}, and C.~{Tang}, ``Federated
  learning-based computation offloading optimization in edge
  computing-supported {Internet of Things},'' \emph{IEEE Access}, vol.~7, pp.
  69\,194--69\,201, 2019.

\bibitem{cheng2019space}
N.~Cheng, F.~Lyu, W.~Quan, C.~Zhou, H.~He, W.~Shi, and X.~Shen,
  ``Space/aerial-assisted computing offloading for {{I}o{T}} applications: A
  learning-based approach,'' \emph{IEEE J. Sel. Areas Commun.}, vol.~37, no.~5,
  pp. 1117--1129, 2019.

\bibitem{zhu2018caching}
H.~Zhu, Y.~Cao, X.~Wei, W.~Wang, T.~Jiang, and S.~Jin, ``Caching transient data
  for {Internet} of {Things}: A deep reinforcement learning approach,''
  \emph{IEEE Internet Things J.}, 2018.

\bibitem{wei2018joint}
Y.~Wei, F.~R. Yu, M.~Song, and Z.~Han, ``Joint optimization of caching,
  computing, and radio resources for fog-enabled {{I}o{T}} using natural
  actor-critic deep reinforcement learning,'' \emph{IEEE Internet Things J.},
  2018.

\bibitem{sasaki2017study}
H.~Sasaki, T.~Horiuchi, and S.~Kato, ``A study on vision-based mobile robot
  learning by deep {Q}-network,'' in \emph{2017 56th Annual Conference of the
  Society of Instrument and Control Engineers of Japan (SICE)}.\hskip 1em plus
  0.5em minus 0.4em\relax IEEE, 2017, pp. 799--804.

\bibitem{xin2017application}
J.~Xin, H.~Zhao, D.~Liu, and M.~Li, ``Application of deep reinforcement
  learning in mobile robot path planning,'' in \emph{2017 Chinese Automation
  Congress (CAC)}.\hskip 1em plus 0.5em minus 0.4em\relax IEEE, 2017, pp.
  7112--7116.

\bibitem{mIoT}
M.~Saravanan, P.~Kumar, and A.~Sharma, ``{I}o{T} enabled indoor autonomous
  mobile robot using cnn and q-learning,'' \emph{2019 IEEE International
  Conference on Industry 4.0, Artificial Intelligence, and Communications
  Technology (IAICT)}, pp. 7--13, 2019.

\bibitem{yan2018path}
T.~Yan, Y.~Zhang, and B.~Wang, ``Path planning for mobile robot's continuous
  action space based on deep reinforcement learning,'' in \emph{2018
  International Conference on Big Data and Artificial Intelligence
  (BDAI)}.\hskip 1em plus 0.5em minus 0.4em\relax IEEE, 2018, pp. 42--46.

\bibitem{tongloy2017asynchronous}
T.~Tongloy, S.~Chuwongin, K.~Jaksukam, C.~Chousangsuntorn, and S.~Boonsang,
  ``Asynchronous deep reinforcement learning for the mobile robot navigation
  with supervised auxiliary tasks,'' in \emph{2017 2nd International Conference
  on Robotics and Automation Engineering (ICRAE)}.\hskip 1em plus 0.5em minus
  0.4em\relax IEEE, 2017, pp. 68--72.

\bibitem{yang2018hierarchical}
Z.~Yang, K.~Merrick, L.~Jin, and H.~A. Abbass, ``Hierarchical deep
  reinforcement learning for continuous action control,'' \emph{IEEE Trans.
  Neural Netw. Learn. Syst.}, no.~99, pp. 1--11, 2018.

\bibitem{gu2017deep}
S.~Gu, E.~Holly, T.~Lillicrap, and S.~Levine, ``Deep reinforcement learning for
  robotic manipulation with asynchronous off-policy updates,'' in \emph{2017
  IEEE International Conference on Robotics and Automation (ICRA)}.\hskip 1em
  plus 0.5em minus 0.4em\relax IEEE, 2017, pp. 3389--3396.

\bibitem{kalashnikov2018scalable}
D.~Kalashnikov, A.~Irpan, P.~Pastor, J.~Ibarz, A.~Herzog, E.~Jang, D.~Quillen,
  E.~Holly, M.~Kalakrishnan, V.~Vanhoucke \emph{et~al.}, ``Scalable deep
  reinforcement learning for vision-based robotic manipulation,'' in
  \emph{Conference on Robot Learning}, 2018, pp. 651--673.

\bibitem{tsurumine2019deep}
Y.~Tsurumine, Y.~Cui, E.~Uchibe, and T.~Matsubara, ``Deep reinforcement
  learning with smooth policy update: Application to robotic cloth
  manipulation,'' \emph{Robotics and Autonomous Systems}, vol. 112, pp. 72--83,
  2019.

\bibitem{yasuda2018collective}
T.~Yasuda and K.~Ohkura, ``Collective behavior acquisition of real robotic
  swarms using deep reinforcement learning,'' in \emph{2018 Second IEEE
  International Conference on Robotic Computing (IRC)}.\hskip 1em plus 0.5em
  minus 0.4em\relax IEEE, 2018, pp. 179--180.

\bibitem{sun2009cooperative}
X.~Sun, T.~Mao, J.~D. Kralik, and L.~E. Ray, ``Cooperative multi-robot
  reinforcement learning: A framework in hybrid state space,'' in \emph{2009
  IEEE/RSJ International Conference on Intelligent Robots and Systems}.\hskip
  1em plus 0.5em minus 0.4em\relax IEEE, 2009, pp. 1190--1196.

\bibitem{sartoretti2019distributed}
G.~Sartoretti, Y.~Wu, W.~Paivine, T.~S. Kumar, S.~Koenig, and H.~Choset,
  ``Distributed reinforcement learning for multi-robot decentralized collective
  construction,'' in \emph{Distributed Autonomous Robotic Systems}.\hskip 1em
  plus 0.5em minus 0.4em\relax Springer, 2019, pp. 35--49.

\bibitem{long2018towards}
P.~Long, T.~Fanl, X.~Liao, W.~Liu, H.~Zhang, and J.~Pan, ``Towards optimally
  decentralized multi-robot collision avoidance via deep reinforcement
  learning,'' in \emph{2018 IEEE International Conference on Robotics and
  Automation (ICRA)}.\hskip 1em plus 0.5em minus 0.4em\relax IEEE, 2018, pp.
  6252--6259.

\bibitem{mataric1997reinforcement}
M.~J. Matari{\'c}, ``Reinforcement learning in the multi-robot domain,'' in
  \emph{Robot colonies}.\hskip 1em plus 0.5em minus 0.4em\relax Springer, 1997,
  pp. 73--83.

\bibitem{liu2019lifelong}
B.~Liu, L.~Wang, M.~Liu, and C.~Xu, ``Lifelong federated reinforcement
  learning: A learning architecture for navigation in cloud robotic systems,''
  \emph{arXiv preprint arXiv:1901.06455}, 2019.

\bibitem{liu2018reinforcement}
H.~Liu, S.~Liu, and K.~Zheng, ``A reinforcement learning-based resource
  allocation scheme for cloud robotics,'' \emph{IEEE Access}, vol.~6, pp.
  17\,215--17\,222, 2018.

\bibitem{yang2005survey}
E.~Yang and D.~Gu, ``A survey on multiagent reinforcement learning towards
  multi-robot systems.'' in \emph{CIG}, 2005.

\bibitem{saha2018comprehensive}
O.~Saha and P.~Dasgupta, ``A comprehensive survey of recent trends in cloud
  robotics architectures and applications,'' \emph{Robotics}, vol.~7, no.~3,
  p.~47, 2018.

\bibitem{yu2016deep}
A.~Yu, R.~Palefsky-Smith, and R.~Bedi, ``Deep reinforcement learning for
  simulated autonomous vehicle control,'' \emph{Course Project Reports:
  Winter}, pp. 1--7, 2016.

\bibitem{vitelli2016carma}
M.~Vitelli and A.~Nayebi, ``{CARMA}: A deep reinforcement learning approach to
  autonomous driving,'' Tech. rep. Stanford University, Tech. Rep., 2016.

\bibitem{mirchevska2018high}
B.~Mirchevska, C.~Pek, M.~Werling, M.~Althoff, and J.~Boedecker, ``High-level
  decision making for safe and reasonable autonomous lane changing using
  reinforcement learning,'' in \emph{2018 21st International Conference on
  Intelligent Transportation Systems (ITSC)}.\hskip 1em plus 0.5em minus
  0.4em\relax IEEE, 2018, pp. 2156--2162.

\bibitem{wu2017flow}
C.~Wu, A.~Kreidieh, K.~Parvate, E.~Vinitsky, and A.~M. Bayen, ``Flow:
  Architecture and benchmarking for reinforcement learning in traffic
  control,'' \emph{arXiv preprint arXiv:1710.05465}, 2017.

\bibitem{gamage2017reinforcement}
H.~D. Gamage and J.~B. Lee, ``Reinforcement learning based driving speed
  control for two vehicle scenario,'' in \emph{Australasian Transport Research
  Forum (ATRF), 39th, 2017, Auckland, New Zealand}, 2017.

\bibitem{you2018highway}
C.~You, J.~Lu, D.~Filev, and P.~Tsiotras, ``Highway traffic modeling and
  decision making for autonomous vehicle using reinforcement learning,'' in
  \emph{2018 IEEE Intelligent Vehicles Symposium (IV)}.\hskip 1em plus 0.5em
  minus 0.4em\relax IEEE, 2018, pp. 1227--1232.

\bibitem{pal2018reinforcement}
M.~K. Pal, R.~Bhati, A.~Sharma, S.~K. Kaul, S.~Anand, and P.~Sujit, ``A
  reinforcement learning approach to jointly adapt vehicular communications and
  planning for optimized driving,'' in \emph{2018 21st International Conference
  on Intelligent Transportation Systems (ITSC)}.\hskip 1em plus 0.5em minus
  0.4em\relax IEEE, 2018, pp. 3287--3293.

\bibitem{wang2017formulation}
P.~Wang and C.-Y. Chan, ``Formulation of deep reinforcement learning
  architecture toward autonomous driving for on-ramp merge,'' in \emph{2017
  IEEE 20th International Conference on Intelligent Transportation Systems
  (ITSC)}.\hskip 1em plus 0.5em minus 0.4em\relax IEEE, 2017, pp. 1--6.

\bibitem{wang2013cooperative}
Q.~Wang and C.~Phillips, ``Cooperative collision avoidance for multi-vehicle
  systems using reinforcement learning,'' in \emph{2013 18th International
  Conference on Methods \& Models in Automation \& Robotics (MMAR)}.\hskip 1em
  plus 0.5em minus 0.4em\relax IEEE, 2013, pp. 98--102.

\bibitem{khamis2014adaptive}
M.~A. Khamis and W.~Gomaa, ``Adaptive multi-objective reinforcement learning
  with hybrid exploration for traffic signal control based on cooperative
  multi-agent framework,'' \emph{Engineering Applications of Artificial
  Intelligence}, vol.~29, pp. 134--151, 2014.

\bibitem{ye2018deep}
H.~Ye and G.~Y. Li, ``Deep reinforcement learning based distributed resource
  allocation for {V2V} broadcasting,'' in \emph{2018 14th International
  Wireless Communications \& Mobile Computing Conference (IWCMC)}.\hskip 1em
  plus 0.5em minus 0.4em\relax IEEE, 2018, pp. 440--445.

\bibitem{challita2019interference}
U.~Challita, W.~Saad, and C.~Bettstetter, ``Interference management for
  cellular-connected {UAVs}: A deep reinforcement learning approach,''
  \emph{IEEE Trans. Wirel. Commun.}, 2019.

\bibitem{he2018integrated}
Y.~He, N.~Zhao, and H.~Yin, ``Integrated networking, caching, and computing for
  connected vehicles: A deep reinforcement learning approach,'' \emph{IEEE
  Trans. Veh. Technol.}, vol.~67, no.~1, pp. 44--55, 2018.

\bibitem{atallah2018scheduling}
R.~F. Atallah, C.~M. Assi, and M.~J. Khabbaz, ``Scheduling the operation of a
  connected vehicular network using deep reinforcement learning,'' \emph{IEEE
  Trans. Intell. Transp. Syst.}, no.~99, pp. 1--14, 2018.

\bibitem{qi2019knowledge}
Q.~Qi, J.~Wang, Z.~Ma, H.~Sun, Y.~Cao, L.~Zhang, and J.~Liao,
  ``Knowledge-driven service offloading decision for vehicular edge computing:
  A deep reinforcement learning approach,'' \emph{IEEE Trans. Veh. Technol.},
  2019.

\bibitem{qi2018vehicular}
Q.~Qi and Z.~Ma, ``Vehicular edge computing via deep reinforcement learning,''
  \emph{arXiv preprint arXiv:1901.04290}, 2018.

\bibitem{zheng2015smdp}
K.~Zheng, H.~Meng, P.~Chatzimisios, L.~Lei, and X.~Shen, ``An {SMDP}-based
  resource allocation in vehicular cloud computing systems,'' \emph{IEEE Trans.
  Ind. Electron.}, vol.~62, no.~12, pp. 7920--7928, 2015.

\bibitem{ydeep}
Y.~Liu, H.~Yu, S.~Xie, and Y.~Zhang, ``Deep reinforcement learning for
  offloading and resource allocation in vehicle edge computing and networks,''
  \emph{IEEE Trans. Veh. Technol.}, vol.~68, no.~11, pp. 11\,158 --11\,168,
  2019.

\bibitem{hou2018q}
L.~Hou, L.~Lei, K.~Zheng, and X.~Wang, ``A {Q}-learning based proactive caching
  strategy for non-safety related services in vehicular networks,'' \emph{IEEE
  Internet Things J.}, 2018.

\bibitem{talpaert2019exploring}
V.~Talpaert, I.~Sobh, B.~R. Kiran, P.~Mannion, S.~Yogamani, A.~El-Sallab, and
  P.~Perez, ``Exploring applications of deep reinforcement learning for
  real-world autonomous driving systems,'' \emph{arXiv preprint
  arXiv:1901.01536}, 2019.

\bibitem{kendall2018learning}
A.~Kendall, J.~Hawke, D.~Janz, P.~Mazur, D.~Reda, J.-M. Allen, V.-D. Lam,
  A.~Bewley, and A.~Shah, ``Learning to drive in a day,'' \emph{arXiv preprint
  arXiv:1807.00412}, 2018.

\bibitem{xu2017end}
H.~Xu, Y.~Gao, F.~Yu, and T.~Darrell, ``End-to-end learning of driving models
  from large-scale video datasets,'' in \emph{Proceedings of the IEEE
  conference on computer vision and pattern recognition}, 2017, pp. 2174--2182.

\bibitem{zheng2016soft}
K.~Zheng, L.~Hou, H.~Meng, Q.~Zheng, N.~Lu, and L.~Lei, ``Soft-defined
  heterogeneous vehicular network: architecture and challenges,'' \emph{IEEE
  Netw.}, vol.~30, no.~4, pp. 72--80, 2016.

\bibitem{liang2019toward}
L.~Liang, H.~Ye, and G.~Y. Li, ``Toward intelligent vehicular networks: A
  machine learning framework,'' \emph{IEEE Internet of Things J.}, vol.~6,
  no.~1, pp. 124--135, 2019.

\bibitem{ye2017machine}
H.~Ye, L.~Liang, G.~Y. Li, J.~Kim, L.~Lu, and M.~Wu, ``Machine learning for
  vehicular networks,'' \emph{arXiv preprint arXiv:1712.07143}, 2017.

\bibitem{inbook}
A.~Mehmood, S.~H. Ahmed, and M.~Sarkar, ``Cyber-physical systems in vehicular
  communications,'' in \emph{Handbook of Research on Advanced Trends in
  Microwave and Communication Engineering}.\hskip 1em plus 0.5em minus
  0.4em\relax IGI Global, 2017, pp. 477--497.

\bibitem{xiao2017vehicular}
Y.~Xiao and C.~Zhu, ``Vehicular fog computing: Vision and challenges,'' in
  \emph{2017 IEEE International Conference on Pervasive Computing and
  Communications Workshops (PerCom Workshops)}.\hskip 1em plus 0.5em minus
  0.4em\relax IEEE, 2017, pp. 6--9.

\bibitem{nobre2019vehicular}
J.~C. Nobre, A.~M. de~Souza, D.~Rosario, C.~Both, L.~A. Villas, E.~Cerqueira,
  T.~Braun, and M.~Gerla, ``Vehicular software-defined networking and fog
  computing: integration and design principles,'' \emph{Ad Hoc Networks},
  vol.~82, pp. 172--181, 2019.

\bibitem{ning2019vehicular}
Z.~Ning, J.~Huang, and X.~Wang, ``Vehicular fog computing: Enabling real-time
  traffic management for smart cities,'' \emph{IEEE Wirel. Commun.}, vol.~26,
  no.~1, pp. 87--93, 2019.

\bibitem{franccois2016deep}
V.~Fran{\c{c}}ois-Lavet, D.~Taralla, D.~Ernst, and R.~Fonteneau, ``Deep
  reinforcement learning solutions for energy microgrids management,'' in
  \emph{European Workshop on Reinforcement Learning (EWRL 2016)}, 2016.

\bibitem{qiu2016heterogeneous}
X.~Qiu, T.~A. Nguyen, and M.~L. Crow, ``Heterogeneous energy storage
  optimization for microgrids,'' \emph{IEEE Trans. Smart Grid}, vol.~7, no.~3,
  pp. 1453--1461, 2016.

\bibitem{mbuwir2017reinforcement}
B.~Mbuwir, F.~Ruelens, F.~Spiessens, and G.~Deconinck, ``Reinforcement
  learning-based battery energy management in a solar microgrid,''
  \emph{Energy-Open}, vol.~2, no.~4, p.~36, 2017.

\bibitem{zeng2018dynamic}
P.~Zeng, H.~Li, H.~He, and S.~Li, ``Dynamic energy management of a microgrid
  using approximate dynamic programming and deep recurrent neural network
  learning,'' \emph{IEEE Trans. Smart Grid}, 2018.

\bibitem{venayagamoorthy2016dynamic}
G.~K. Venayagamoorthy, R.~K. Sharma, P.~K. Gautam, and A.~Ahmadi, ``Dynamic
  energy management system for a smart microgrid,'' \emph{IEEE Trans. Neural
  Netw. Learn. Syst.}, vol.~27, no.~8, pp. 1643--1656, 2016.

\bibitem{ji2019real}
Y.~Ji, J.~Wang, J.~Xu, X.~Fang, and H.~Zhang, ``Real-time energy management of
  a microgrid using deep reinforcement learning,'' \emph{Energies}, vol.~12,
  no.~12, p. 2291, 2019.

\bibitem{mocanu2018line}
E.~Mocanu, D.~C. Mocanu, P.~H. Nguyen, A.~Liotta, M.~E. Webber, M.~Gibescu, and
  J.~G. Slootweg, ``On-line building energy optimization using deep
  reinforcement learning,'' \emph{IEEE Trans. Smart Grid}, 2018.

\bibitem{wen2015optimal}
Z.~Wen, D.~O’Neill, and H.~Maei, ``Optimal demand response using device-based
  reinforcement learning,'' \emph{IEEE Trans. Smart Grid}, vol.~6, no.~5, pp.
  2312--2324, 2015.

\bibitem{o2010residential}
D.~O'Neill, M.~Levorato, A.~Goldsmith, and U.~Mitra, ``Residential demand
  response using reinforcement learning,'' in \emph{2010 First IEEE
  International Conference on Smart Grid Communications}.\hskip 1em plus 0.5em
  minus 0.4em\relax IEEE, 2010, pp. 409--414.

\bibitem{claessens2016convolutional}
B.~J. Claessens, P.~Vrancx, and F.~Ruelens, ``Convolutional neural networks for
  automatic state-time feature extraction in reinforcement learning applied to
  residential load control,'' \emph{arXiv preprint arXiv:1604.08382}, 2016.

\bibitem{ruelens2014demand}
F.~Ruelens, B.~J. Claessens, S.~Vandael, S.~Iacovella, P.~Vingerhoets, and
  R.~Belmans, ``Demand response of a heterogeneous cluster of electric water
  heaters using batch reinforcement learning,'' in \emph{2014 Power Systems
  Computation Conference}.\hskip 1em plus 0.5em minus 0.4em\relax IEEE, 2014,
  pp. 1--7.

\bibitem{ruelens2016reinforcement}
F.~Ruelens, B.~J. Claessens, S.~Quaiyum, B.~De~Schutter, R.~Babu{\v{s}}ka, and
  R.~Belmans, ``Reinforcement learning applied to an electric water heater:
  from theory to practice,'' \emph{IEEE Trans. Smart Grid}, vol.~9, no.~4, pp.
  3792--3800, 2016.

\bibitem{lu2018dynamic}
R.~Lu, S.~H. Hong, and X.~Zhang, ``A dynamic pricing demand response algorithm
  for smart grid: reinforcement learning approach,'' \emph{Applied Energy},
  vol. 220, pp. 220--230, 2018.

\bibitem{lu2019incentive}
R.~Lu and S.~H. Hong, ``Incentive-based demand response for smart grid with
  reinforcement learning and deep neural network,'' \emph{Applied energy}, vol.
  236, pp. 937--949, 2019.

\bibitem{lim2014strategic}
Y.~Lim and H.-M. Kim, ``Strategic bidding using reinforcement learning for load
  shedding in microgrids,'' \emph{Computers \& Electrical Engineering},
  vol.~40, no.~5, pp. 1439--1446, 2014.

\bibitem{xiao2017energy}
X.~Xiao, C.~Dai, Y.~Li, C.~Zhou, and L.~Xiao, ``Energy trading game for
  microgrids using reinforcement learning,'' in \emph{International Conference
  on Game Theory for Networks}.\hskip 1em plus 0.5em minus 0.4em\relax
  Springer, 2017, pp. 131--140.

\bibitem{xiao2018reinforcement}
L.~Xiao, X.~Xiao, C.~Dai, M.~Pengy, L.~Wang, and H.~V. Poor, ``Reinforcement
  learning-based energy trading for microgrids,'' \emph{arXiv preprint
  arXiv:1801.06285}, 2018.

\bibitem{kim2014dynamic}
B.-G. Kim, Y.~Zhang, M.~Van Der~Schaar, and J.-W. Lee, ``Dynamic pricing for
  smart grid with reinforcement learning,'' in \emph{2014 IEEE Conference on
  Computer Communications Workshops (INFOCOM WKSHPS)}.\hskip 1em plus 0.5em
  minus 0.4em\relax IEEE, 2014, pp. 640--645.

\bibitem{kim2016dynamic}
B.~G. Kim, Y.~Zhang, M.~Van Der~Schaar, and J.-W. Lee, ``Dynamic pricing and
  energy consumption scheduling with reinforcement learning,'' \emph{IEEE
  Trans. Smart Grid}, vol.~7, no.~5, pp. 2187--2198, 2016.

\bibitem{reddy2011strategy}
P.~P. Reddy and M.~M. Veloso, ``Strategy learning for autonomous agents in
  smart grid markets,'' in \emph{Twenty-second international joint conference
  on artificial intelligence}, 2011.

\bibitem{foruzan2018reinforcement}
E.~Foruzan, L.-K. Soh, and S.~Asgarpoor, ``Reinforcement learning approach for
  optimal distributed energy management in a microgrid,'' \emph{IEEE Trans.
  Power Syst.}, vol.~33, no.~5, pp. 5749--5758, 2018.

\bibitem{Bedi2018}
G.~{Bedi}, G.~K. {Venayagamoorthy}, R.~{Singh}, R.~R. {Brooks}, and K.~{Wang},
  ``Review of {Internet} of {Things} {(IoT)} in electric power and energy
  systems,'' \emph{IEEE Internet Things J.}, vol.~5, no.~2, pp. 847--870, Apr.
  2018.

\bibitem{Rana2017}
M.~{Rana}, L.~{Li}, and S.~W. {Su}, ``Distributed state estimation over
  unreliable communication networks with an application to smart grids,''
  \emph{IEEE Trans. on Green Communications and Networking}, vol.~1, no.~1, pp.
  89--96, Mar. 2017.

\bibitem{Rana2017a}
------, ``Distributed state estimation over unreliable communication networks
  with an application to smart grids,'' \emph{IEEE Trans. on Green
  Communications and Networking}, vol.~1, no.~1, pp. 89--96, Mar. 2017.

\bibitem{igl2018deep}
M.~Igl, L.~Zintgraf, T.~A. Le, F.~Wood, and S.~Whiteson, ``Deep variational
  reinforcement learning for {POMDPs},'' \emph{arXiv preprint
  arXiv:1806.02426}, 2018.

\bibitem{Katsikopoulos2003}
K.~V. Katsikopoulos and S.~E. Engelbrecht, ``Markov decision processes with
  delays and asynchronous cost collection,'' \emph{IEEE Trans. Automat.
  Contr.}, vol.~48, no.~4, pp. 568--574, 2003.

\bibitem{Walsh2009}
T.~J. Walsh, A.~Nouri, L.~Li, and M.~L. Littman, ``Learning and planning in
  environments with delayed feedback,'' \emph{AUTON. AGENT MULTI-AG.}, vol.~18,
  no.~1, p.~83, 2009.

\bibitem{Schuitema2010}
E.~Schuitema, L.~Bu{\c{s}}oniu, R.~Babu{\v{s}}ka, and P.~Jonker, ``Control
  delay in reinforcement learning for real-time dynamic systems: a memoryless
  approach,'' in \emph{2010 IEEE/RSJ International Conference on Intelligent
  Robots and Systems}.\hskip 1em plus 0.5em minus 0.4em\relax IEEE, 2010, pp.
  3226--3231.

\bibitem{lample2017playing}
G.~Lample and D.~S. Chaplot, ``Playing {FPS} games with deep reinforcement
  learning,'' in \emph{Thirty-First AAAI Conference on Artificial
  Intelligence}, 2017.

\bibitem{hochreiter1997long}
S.~Hochreiter and J.~Schmidhuber, ``Long short-term memory,'' \emph{Neural
  computation}, vol.~9, no.~8, pp. 1735--1780, 1997.

\bibitem{hochreiter2001gradient}
S.~Hochreiter, Y.~Bengio, P.~Frasconi, J.~Schmidhuber \emph{et~al.}, ``Gradient
  flow in recurrent nets: the difficulty of learning long-term dependencies,''
  2001.

\end{thebibliography}
\bibliographystyle{IEEEtran}

%\begin{IEEEbiography}[{\includegraphics[width=1in,height=1.25in,clip,keepaspectratio]{lei}}]{Lei Lei}
%(SM'16) received the B.S. and Ph.D. degrees in telecommunications engineering from the Beijing University of Posts and Telecommunications, Beijing, China, in 2001 and 2006, respectively. She is currently an associate professor in the College of Engineering and Physical Sciences at the University of Guelph, Canada. Her research interests mainly lie in Machine Learning/Deep Reinforcement Learning, Internet of Things/Internet of Vehicles, Mobile Edge Computing, and Smart Grid. 
%\end{IEEEbiography}
%
%\begin{IEEEbiography}[{\includegraphics[width=1in,height=1.25in,clip,keepaspectratio]{kan}}]{Kan Zheng}
%	(SM'16) received the B.S. and Ph.D. degrees in telecommunications engineering from the Beijing University of Posts and Telecommunications, Beijing, China, in 2001 and 2006, respectively. She is currently an associate professor in the College of Engineering and Physical Sciences at the University of Guelph, Canada. Her research interests mainly lie in Machine Learning/Deep Reinforcement Learning, Internet of Things/Internet of Vehicles, Mobile Edge Computing, and Smart Grid. 
%\end{IEEEbiography}

\end{document}